\documentclass{article}

\PassOptionsToPackage{numbers, compress}{natbib}

\usepackage[preprint]{neurips_2026}

\usepackage[utf8]{inputenc} 
\usepackage[T1]{fontenc}    
\usepackage{hyperref}       
\usepackage{url}            
\usepackage{booktabs}       
\usepackage{amsfonts}       
\usepackage{nicefrac}       
\usepackage{microtype}      
\usepackage{xcolor}         

\usepackage{tabularx}
\newcolumntype{C}{>{\centering\arraybackslash}X}
\usepackage{amsthm}
\newtheorem{lemma}{Lemma}
\newtheorem{proposition}{Proposition}
\usepackage{amsmath}        
\usepackage{amssymb}
\usepackage{cleveref}       
\usepackage{graphicx}
\usepackage{doi}
\usepackage{float}          
\usepackage{placeins}   
\usepackage{subcaption} 
\usepackage{wrapfig}    
\usepackage{afterpage}  

\hypersetup{
    pdftitle={Mean Mode Screaming: Mean-Variance Split Residuals for 1000-Layer Diffusion Transformers},
    pdfsubject={Computer Science},
    pdfauthor={Pengqi Lu},
    pdfkeywords={Diffusion Transformers, Scaling Laws, Deep Learning Dynamics, Mean Mode Screaming},
}

\title{Mean Mode Screaming: Mean--Variance Split Residuals for 1000-Layer Diffusion Transformers}

\author{%
  Pengqi Lu \\
  Beijing, China \\
  \texttt{luer5old@gmail.com} \\
}

\begin{document}

\maketitle

\begin{figure}[H]
    \centering
    \includegraphics[width=\linewidth]{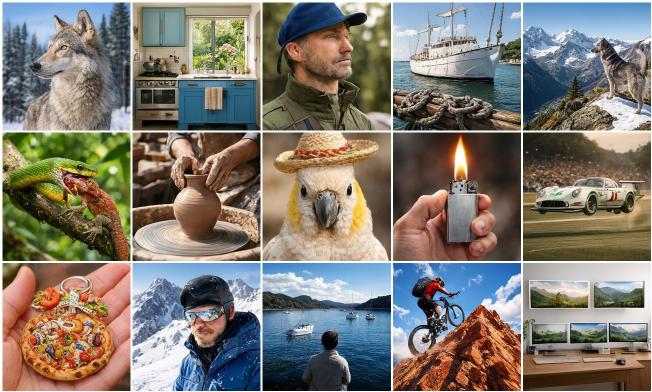}
    \caption{\textbf{Text-to-image generation samples from our 1000-layer MV-Split DiT.} More samples are provided in Appendix~\ref{sec:appendix_more_samples}. Code: \url{https://github.com/erwold/mv-split}.\ Model weights: \url{https://huggingface.co/StableKirito/mvsplit-dit-1000l}.}
    \label{fig:teaser}
\end{figure}

\begin{abstract}
Scaling Diffusion Transformers (DiTs) to hundreds of layers introduces a structural vulnerability: networks can enter a silent, mean-dominated collapse state that homogenizes token representations and suppresses centered variation. Through mechanistic auditing, we isolate the trigger event of this collapse as \textbf{Mean Mode Screaming (MMS)}. MMS can occur even when training appears stable, with a mean-coherent backward shock on residual writers that opens deep residual branches and drives the network into a mean-dominated state. We show this behavior is driven by an exact decomposition of these gradients into mean-coherent and centered components, compounded by the structural suppression of attention-logit gradients through the null space of the Softmax Jacobian once values homogenize.

To address this, we propose \textbf{Mean--Variance Split (MV-Split) Residuals}, which combine a separately gained centered residual update with a leaky trunk-mean replacement. On a 400-layer single-stream DiT, MV-Split prevents the divergent collapse that crashes the un-stabilized baseline; it tracks close to the baseline's pre-crash trajectory while remaining substantially better than token-isotropic gating methods such as LayerScale across the full schedule. Finally, we present a 1000-layer DiT as a scale-validation run at boundary scales, establishing that the architecture remains stably trainable at extreme depth.
\end{abstract}


\section{Introduction}
Scaling laws for generative modeling~\cite{scalinglaws2024} indicate that depth is an important dimension of capacity and model performance. Training ultra-deep Diffusion Transformers (DiTs)~\cite{ho2020denoising,song2021score,peebles2023dit}, however, introduces structural reliability issues that are not well described by standard exploding or vanishing gradient heuristics. In some runs, optimization remains stable for thousands of steps and then diverges within a few updates, with the loss returning near its initialization level and not recovering. These events can occur without NaNs or obvious forward saturation.

In this work, we study a \textbf{mean-dominated collapse state} in ultra-deep DiTs, in which token representations homogenize and centered token variation is suppressed. We reserve the term \textbf{Mean Mode Screaming (MMS)} for the abrupt entry event into this state: a spike in the mean-coherent gradient component, rapid residual branch opening, and subsequent Q/K gradient suppression.

Mechanistically, this failure exploits a geometric asymmetry between the token-mean and centered subspaces. Row-stochastic attention strictly preserves pure-mean states, while the centered component is propagated by a separate mixing operator and can become contractive in deep layers. On the backward pass, gradients admit an exact decomposition into mean-coherent and centered components; as token alignment increases, the mean-coherent component accumulates coherently with sequence length and can dominate the residual branch update. Once values homogenize, attention-logit gradients are suppressed through the null space of the Softmax Jacobian, suppressing Q/K learning and locking the network into the collapsed state.

Existing depth stabilizers suppress the entire residual branch isotropically in token space: ReZero~\cite{bachlechner2021rezero} and LayerScale~\cite{touvron2021layerscale} apply scalar and per-channel learnable gates respectively, shrinking the mean and centered components together. This stabilizes training but slows convergence by also damping the centered signal responsible for spatially varying feature learning.

These observations motivate \textbf{MV-Split Residuals}, which combine a separately gained centered residual update with a leaky trunk-mean replacement. By damping the mean path without shrinking the centered path by the same factor, MV-Split stabilizes training without the convergence cost of isotropic residual gating.

Our contributions are:
\begin{enumerate}
    \item \textbf{Characterization.} We characterize a mean-dominated collapse state and distinguish it from MMS, the abrupt entry event into this state. A standard-initialization control reaches the same collapse state more progressively across depth.
    \item \textbf{Mechanism.} We show that row-stochastic attention preserves pure-mean states, that gradients split exactly into mean-coherent and centered components, with the mean-coherent component entering an $\mathcal{O}(T)$ coherent regime when tokens align, and that value homogenization suppresses attention-logit gradients through the null space of the Softmax Jacobian.
    \item \textbf{Method and result.} We propose \textbf{MV-Split Residuals}, which combine a separately gained centered residual update with a leaky trunk-mean replacement. In matched 400-layer quantitative evaluation, MV-Split removes collapse events and converges faster than LayerScale; in a separate 1000-layer run, the same design remains stably trainable and serves as a scale-validation run at boundary scales.
\end{enumerate}

\section{Preliminaries}
\label{sec:preliminary}
We first describe the backbone, initialization, and training objective used in
the main training runs.

\begin{figure}[t]
    \centering
    \begin{minipage}[c]{0.44\linewidth}
        \centering
        \includegraphics[width=\linewidth]{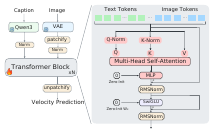}
    \end{minipage}
    \hfill
    \begin{minipage}[c]{0.36\linewidth}
        \centering
        \includegraphics[width=\linewidth]{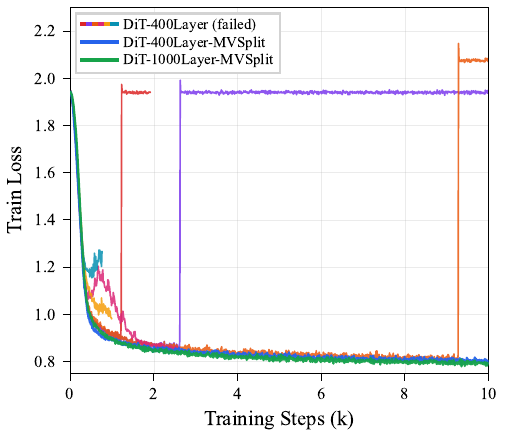}
    \end{minipage}
    \hfill
    \begin{minipage}[c]{0.19\linewidth}
        \centering
        \includegraphics[width=\linewidth]{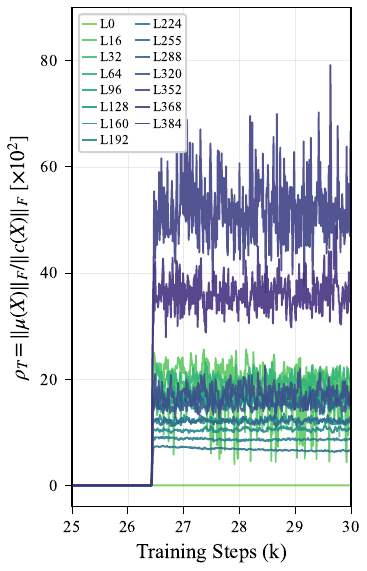}
    \end{minipage}
    
    \caption{\textbf{Baseline DiT and representative training diagnostics.}
    \textbf{(Left)} Single-stream DiT backbone.
    \textbf{(Middle)} Training loss over the first 10k steps for the
    un-stabilized 400-layer baseline and the MV-Split 400-/1000-layer runs.
    \textbf{(Right)} Per-layer energy ratio
    $\rho_T=\|\mu(X)\|_F/\|c(X)\|_F$
    (Appendix~\ref{sec:appendix_metrics}) across L0--L384 in a baseline run.}
    \label{fig:arch_and_dynamics}
\end{figure}

\subsection{Minimal Single-Stream Multi-Modal Diffusion Transformer}
\label{sec:backbone}
We use a deliberately stripped-down single-stream DiT~\cite{peebles2023dit} so that deep residual propagation, rather than external modulation or skip pathways, remains the dominant carrier of both signal and gradients. Concretely, we employ a \textbf{Post-Norm} residual chain~\cite{vaswani2017attention,xiong2020layernorm} ($X_{l+1} = \text{RMSNorm}(X_l + f_l(X_l))$~\cite{zhang2019root}) without AdaLN~\cite{peebles2023dit} or other per-layer modulation mechanisms, to avoid introducing alternative depthwise control channels that would complicate attribution of the collapse dynamics. Instead of cross-attention, we concatenate VAE-encoded~\cite{kingma2014vae,rombach2022high} image tokens $X_{img}$ and text embedding tokens $X_{txt}$ into a unified sequence $X_{in} = [X_{img}; X_{txt}]$~\cite{bao2023uvit,esser2024sd3,qin2025luminaimage,zimage2025}, forcing self-attention~\cite{vaswani2017attention} to handle all multimodal interaction. For positional encoding, we apply a 2D extension of RoPE~\cite{su2024roformer} to image tokens following recent vision/diffusion Transformer practice~\cite{chu2024visionllama,lu2024fit}, and leave text tokens without rotary positional encoding.
The left panel of Figure~\ref{fig:arch_and_dynamics} gives the corresponding
backbone schematic.

\subsection{Residual Writer Zero Initialization}
\label{sec:zero_init_setting}
For the main training runs used in the main text, except the LayerScale
control, we zero-initialize the \textbf{residual writers} ($W_O$ and $W_2$),
following the broader practice of identity-initialized residual branches and
zero-initialized output pathways in residual and diffusion
architectures~\cite{goyal2017largebatch,zhang2019fixup,peebles2023dit,zhang2023controlnet,zhu2025unveilingadalnzero}.
Here $W_O$ is the attention output projection. For the FFN branch, we write the
SwiGLU~\cite{shazeer2020glu} feed-forward transformation as
\begin{equation}
    [g_l, v_l] = W_{13} x_l, \qquad \mathrm{FFN}(x_l) = W_2 \bigl( \mathrm{SiLU}(g_l) \odot v_l \bigr),
\end{equation}
so $W_2$ is the residual writer of the FFN block. In these zero-writer
training runs, the internal branch parameters (e.g., $Q,K,V$ and $W_{13}$)
remain at their standard initialization.
Appendix~\ref{sec:appendix_standard_init} shows that standard initialization
does not avoid the mean-dominated regime; the same collapse appears from the start
as a depth-progressive front, rather than through the delayed writer-opening
spike that defines MMS in the zero-writer training runs.

\subsection{Rectified Flow Matching}
We train the model using a \textbf{Rectified Flow}~\cite{liu2022rectified,lipman2023flowmatching} objective. Given a data distribution $x_0$ (VAE latents) and a Gaussian noise distribution $x_1 \sim \mathcal{N}(0, I)$, we define a linear interpolation path $z_t = (1-t)x_0 + t x_1$ for $t \in [0, 1]$.
The model $v_\theta$ is trained to predict the vector field pointing from noise toward data:
\begin{equation}
    \mathcal{L} = \mathbb{E}_{t, x_0, x_1} \left[ \| v_\theta(z_t, X_{txt}) - (x_0 - x_1) \|^2 \right]
\end{equation}

\section{Failure Dynamics: Mean-Dominated Collapse}
\label{sec:diagnosis}

To understand the failure mode limiting depth scaling, we analyze a representative abrupt-failure run from the main diagnostic regime. We first introduce a token-space decomposition that separates sequence-mean and centered variation. We then use this decomposition to trace the observed divergence sequence: a mean-coherent gradient shock, residual branch opening, mean-dominated forward collapse, and Q/K gradient suppression. Section~\ref{sec:mechanism} explains why this sequence occurs.

\subsection{Geometric Preliminaries: Token-Space Asymmetry}
\label{sec:geometry}

The failure dynamics are fundamentally tied to how information is distributed across tokens. Let $\mathbf{1}\in\mathbb{R}^T$ denote the all-ones vector, and define $J \triangleq \frac{1}{T}\mathbf{1}\mathbf{1}^\top$ and $P \triangleq I - J$. For any token sequence $X \in \mathbb{R}^{T \times D}$, we write
\begin{equation}
    X = JX + PX \;\equiv\; \mu(X) + c(X),
\end{equation}
where $\mu(X)\triangleq JX$ is the sequence-mean component and $c(X)\triangleq PX$ is the centered variation component. Row-stochastic attention acts asymmetrically on these two subspaces.

\begin{proposition}[Pure-mean component is preserved]
\label{prop:mean-fixed}
For any row-stochastic attention matrix $A$ satisfying $A\mathbf{1}=\mathbf{1}$, $A\,\mu(X) = \mu(X)$.
\end{proposition}

Note that Proposition~\ref{prop:mean-fixed} governs only the pure-mean component of the input. For a general input $X = JX + PX$, the output mean satisfies $\mu(AX) = JAX = JX + JAPX$; centered variation can therefore contribute to the output mean through the leakage term $JAPX$.

\begin{proposition}[Centered component is governed by $PAP$]
\label{prop:centered-pap}
For any row-stochastic attention matrix $A$ satisfying $A\mathbf{1}=\mathbf{1}$,
\begin{equation}
    c(AX) = PAX = PAPX,
    \quad\text{and therefore}\quad
    \|c(AX)\|_F \le \|PAP\|_2\,\|c(X)\|_F.
\end{equation}
\end{proposition}

We denote $\mu_{\text{eff}}(A) \triangleq \|PAP\|_2$. When $\mu_{\text{eff}}(A)<1$, the layer is strictly contractive on the centered subspace. This geometric asymmetry imposes a structural vulnerability: row-stochastic attention leaves pure-mean states invariant, while its action on token-specific variation is governed by $\mu_{\text{eff}}$ and can become contractive. Consequently, the network must rely on residual branches to continuously replenish the centered subspace. If the residual updates become dominated by the mean component, the representation is driven toward a pure-mean state.

\subsection{Tracing the Divergence Event: From Trigger to Lock-in}
\label{sec:empirical_evidence}

\begin{figure}[t!]
    \centering
    \includegraphics[width=\linewidth]{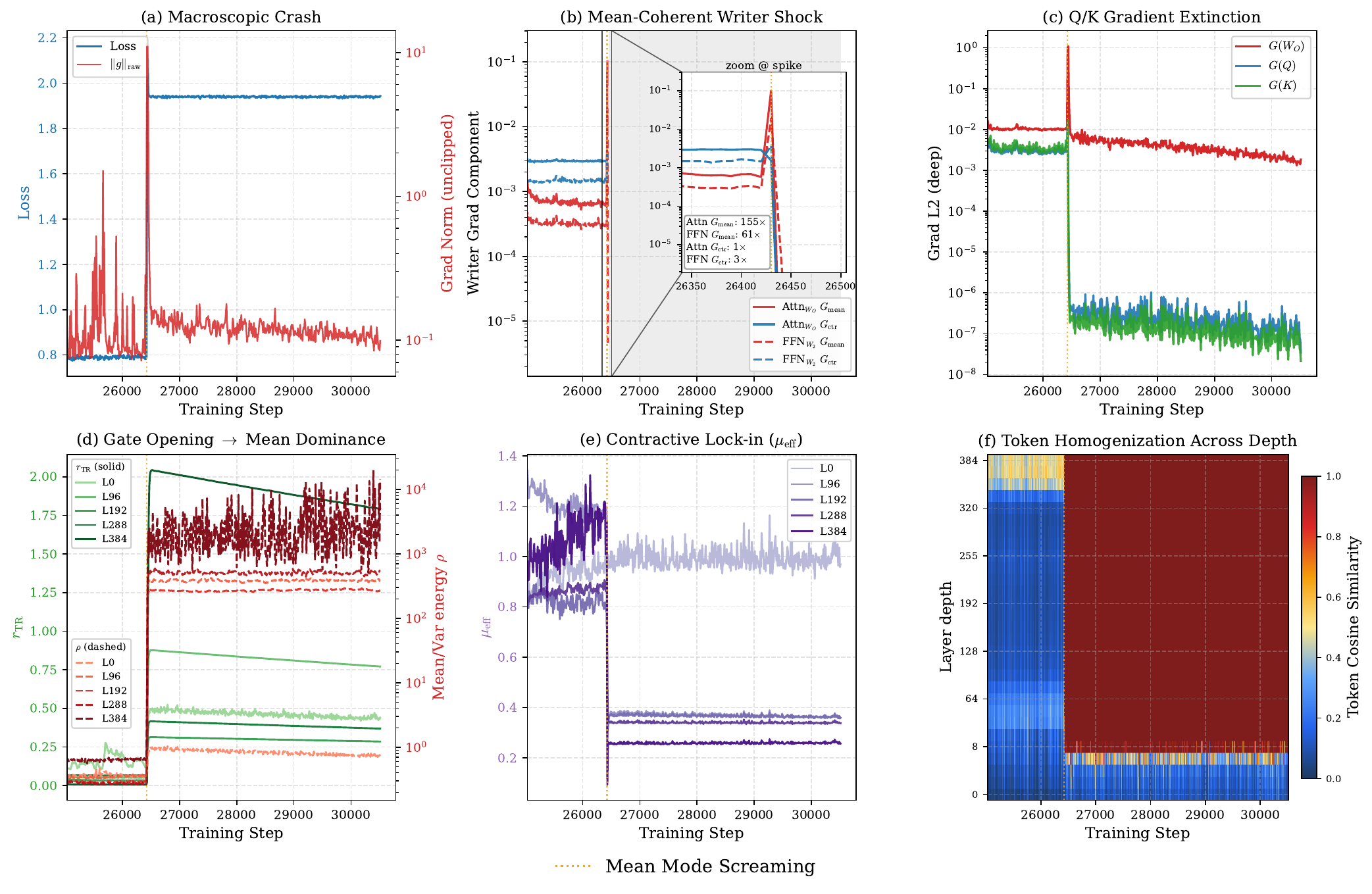}
    \caption{\textbf{Empirical trajectory of a representative divergence event
(400-layer).} The vertical dashed line marks the
divergence step. \textbf{(a--c) Backward trigger:} The global gradient norm
spikes (a). The spike is concentrated in the mean-coherent gradient component
$G_{\text{mean}}$, while the centered component $G_{\text{ctr}}$ shows no
comparable amplification (b). After the spike, Q/K gradients drop by roughly
four orders of magnitude while $W_O$ gradients remain nonzero (c).
\textbf{(d--f) Forward lock-in:} The residual branch opens and the
mean/centered energy ratio $\rho_T$ rises sharply (d). Deep attention remains
contractive on the centered subspace, with limited variance replenishment (e).
Token representations homogenize across depth, with cosine similarity
approaching one in deep layers (f).}
    \label{fig:dynamics}
\end{figure}

Figure~\ref{fig:dynamics} traces the divergence in a 400-layer baseline through a tight chronological sequence. The backward pass exhibits a mode-selective shock: the gradient spike is concentrated primarily in the mean-coherent component while Q/K gradients collapse in lockstep, leaving residual writers as the dominant active learning channel. This shock then locks in across the forward pass --- branches open into a mean-dominated regime ($\rho_T$ explodes), and with attention contractive on the centered subspace and no branch-side variance replenishment, tokens homogenize across depth into a trivial mean-prediction baseline.

This empirical sequence isolates two questions for the mechanistic analysis in Section~\ref{sec:mechanism}: (1) why the gradient amplifies specifically in the mean-coherent direction, and (2) why token homogenization structurally suppresses Q/K gradients.

\section{Mechanism}
\label{sec:mechanism}

\subsection{Gradient Decomposition and Backward Alignment Amplification Law}
\label{sec:mechanism_trigger}

Consider a token-wise linear map $W$ (e.g., residual writers $W_O, W_2$) whose gradient takes the form $\nabla_W \mathcal{L} = \sum_{t=1}^T \delta_t y_t^\top$. Decomposing the forward inputs $y_t$ and backward gradients $\delta_t$ into their sequence means ($\bar y,\bar\delta$) and centered residuals ($\tilde y_t,\tilde\delta_t$), the cross-terms vanish identically under summation (proof in Appendix~\ref{sec:appendix_gmd_proof}), yielding an exact additive decomposition:
\begin{equation}
    \nabla_W \mathcal{L}
    \;=\;
    \underbrace{T\,\bar{\delta}\,\bar{y}^\top}_{\Delta W_{\mu}\ \text{(mean-coherent, $\mathcal{O}(T)$ when aligned)}}
    \;+\;
    \underbrace{\textstyle\sum_{t=1}^T \tilde{\delta}_t \tilde{y}_t^\top}_{\Delta W_{c}\ \text{(centered, diffusive)}}.
    \label{eq:writer-gradient-decomposition}
\end{equation}
We denote $G_{\text{mean}}\triangleq\|\Delta W_\mu\|_F$ and $G_{\text{ctr}}\triangleq\|\Delta W_c\|_F$. This decomposition exposes a scaling transition. The mean component has norm $\|\Delta W_\mu\|_F=T\,\|\bar{\delta}\|\,\|\bar{y}\|$, so it remains small when sequence means cancel; under weak centered alignment, $\Delta W_c$ sums diffusively. As representations and adjoints homogenize, however, the sequence means stop canceling, $\|\bar y\|$ and $\|\bar\delta\|$ become order-one, and the rank-1 mean mode enters its coherent $\mathcal{O}(T)$ regime. Operationally, Mean Mode Screaming acts as a sharp transition from diffusive cancellation to coherent accumulation.

To quantify this transition, we define the dimensionless \emph{alignment amplification} $\mathcal{A}$ as the ratio of the true gradient energy to the independent-token baseline. As derived in Appendix~\ref{sec:appendix_alignment_law}, expanding this ratio yields an identity linking the cross-token coherent amplification of gradients to microscopic token alignment. Under an equal-magnitude proxy, it takes the compact form:
\begin{equation}
  \underbrace{\frac{\|\nabla_W\mathcal{L}\|_F^2}{\sum_t\|\delta_t\|^2\|y_t\|^2}}_{\text{Amplification } \mathcal{A}} \;-\; 1
  \;=\;
  \frac{\sum_{s\ne t}(\delta_s^{\!\top}\delta_t)(y_s^{\!\top}y_t)}{\sum_t\|\delta_t\|^2\|y_t\|^2}
  \;\approx\;
  (T-1)\,\underbrace{\mathbb{E}_{s\ne t}\bigl[\cos(y_s,y_t)\cos(\delta_s,\delta_t)\bigr]}_{\text{Pairwise alignment } \kappa}.
  \label{eq:alignment_amp_exact}
\end{equation}

Equation~\ref{eq:alignment_amp_exact} identifies when token-wise gradients stop canceling and enter a coherent accumulation regime. When tokens are heterogeneous, signed off-diagonal terms cancel ($\kappa \approx 0$) and $\mathcal{A} \approx 1$. As both representations and adjoints become aligned in deep layers, the signed off-diagonal terms stop canceling; in the limiting case $\cos(y_s,y_t)\to 1$ and $\cos(\delta_s,\delta_t)\to 1$, giving $\kappa\to 1$ and the gradient enters its $\mathcal{O}(T)$ coherent-amplification regime. We empirically audit this transition in Section~\ref{sec:exp_scaling_law} using the absolute-coherence upper-envelope proxy $\hat\kappa\triangleq\mathbb{E}_{s\ne t}[|\cos(y_s,y_t)|\,|\cos(\delta_s,\delta_t)|]$.

\subsection{Q/K Gradient Extinction via the Softmax Null Space}
\label{sec:jacobian-vanishing}

A gradient spike alone would not lock in the failure if the attention
path could restore token variation. However, once the residual stream
becomes mean-dominated, the value vectors homogenize. Consequently,
the Softmax Jacobian zeroes out the constant component of the
attention-weight gradient.

\begin{lemma}[Softmax null space under value collapse]
\label{lem:softmax-nullspace}
For one attention row $i$, if $V_j=\bar v$ for all $j$, then
$\partial\mathcal{L}/\partial S_i=\mathbf{0}$, where $S_i$ is the
vector of pre-softmax logits.
\end{lemma}

By the chain rule,
$\partial\mathcal{L}/\partial a_{ij}=\langle\partial\mathcal{L}/
\partial Y_i,V_j\rangle$ is independent of $j$ when $V_j=\bar v$,
yielding $\partial\mathcal{L}/\partial a_i\propto\mathbf{1}$. Because
$J_{\text{sm}}(a_i)\mathbf{1}=\mathbf{0}$, the logit gradient strictly
vanishes. Under approximate homogeneity, this null space still removes
the constant component, strongly suppressing Q/K learning while the
residual-writer gradient (Eq.~\ref{eq:writer-gradient-decomposition})
is not zeroed by this Softmax null space (proof in Appendix~\ref{sec:appendix_qk_extinction}).

\section{Method: MV-Split Residuals}
\label{sec:method}

Section~\ref{sec:mechanism} isolates a single unstable mode: the
rank-one mean-coherent gradient update $\Delta W_\mu$. We therefore
decouple its residual gain from the centered update. Let
$X_l\in\mathbb{R}^{T\times D}$ be the trunk and
$F_l\triangleq f_l(X_l)$ the branch output. Using the orthogonal
projectors $J$ and $P=I-J$ from Section~\ref{sec:geometry}, we
replace the standard Post-Norm merge $X_{l+1}=\text{RMSNorm}
(X_l+F_l)$ with a subspace-routed merge:
\begin{align}
  Z_l &\triangleq X_l
       \;+\;
       \underbrace{\beta\odot(PF_l)}_{\text{centered path}}
       \;+\;
       \underbrace{\alpha\odot J(F_l - X_l)}_{\text{mean path}},
  \label{eq:mv_merge}\\
  X_{l+1} &= \text{RMSNorm}(Z_l),
\end{align}
where $\alpha,\beta\in\mathbb{R}^D$ are per-block learnable vectors
broadcast across tokens. Our multimodal transformer implementation applies the residual projectors segment-wise ($J_{\mathrm{seg}}, P_{\mathrm{seg}}$) to avoid directly mixing image and text means in the residual control path (Appendix~\ref{sec:appendix_mv_split_seg}).

\textbf{Forward dynamics.} Prior to token-dependent RMS normalization,
projecting Eq.~\ref{eq:mv_merge} exactly decouples the pre-normalization merge:
\begin{equation}
  PZ_l = PX_l + \beta\odot(PF_l), \qquad JZ_l = (1-\alpha)\odot(JX_l) + \alpha\odot(JF_l).
  \label{eq:mv_dynamics}
\end{equation}
The centered subspace follows a standard residual update with gain
$\beta$, while the mean subspace becomes a per-feature leaky integrator (when $0<\alpha_d\le 1$):
each layer contracts the trunk mean by $1-\alpha_d$ before adding a
fresh correction.

\textbf{Backward dynamics.} Let $G_l\triangleq\partial\mathcal{L}/\partial Z_l$. Because $J,P$ are
self-adjoint and orthogonal, the gradient flowing back into the branch
factors along the same split:
\begin{equation}
  \frac{\partial\mathcal{L}}{\partial F_l}
  \;=\;\beta\odot(PG_l)+\alpha\odot(JG_l).
  \label{eq:mv_backward}
\end{equation}
Centered and mean-coherent gradients receive independent gains.
Together with (\ref{eq:mv_dynamics}), a small $\alpha$ both damps
mean-coherent forward accumulation and shrinks the $\Delta W_\mu$
component of the gradient
(Eq.~\ref{eq:writer-gradient-decomposition}) by the same factor,
without tying the local centered branch-gradient to the small mean gain $\alpha$.

\paragraph{Comparison to other residual-gain methods.}
LayerScale~\cite{touvron2021layerscale} and
ReZero~\cite{bachlechner2021rezero} apply a single residual gain
(per-channel and scalar, respectively) that does not distinguish
the mean and centered subspaces, so $\Delta W_\mu$ and $\Delta W_c$
are suppressed jointly. We elaborate on the structural distinctions between MV-Split and
these residual-gain methods in Appendix~\ref{sec:appendix_layerscale}.

\section{Experiments}
\label{sec:experiments}

The 400-layer comparison is matched in backbone, optimizer, data, batch size, and non-residual primitives on ImageNet-2012~\cite{russakovsky2015imagenet} latents
encoded with a frozen FLUX.2 VAE~\cite{rombach2022high,blackforestlabs2025flux2}
and conditioned on a frozen Qwen3-0.6B text
encoder~\cite{qwen3technicalreport}; each stabilizer (un-stabilized Post-Norm baseline, LayerScale controls, MV-Split) uses its standard residual-initialization protocol (Appendix~\ref{sec:appendix_training_config}). A separate
1000-layer run uses the same residual design and is
reported as a 1000-layer scale-validation run (Figure~\ref{fig:teaser} and
Appendix~\ref{sec:appendix_more_samples}), trained from ImageNet pre-training
through post-training on
a separate $\sim$50k curated image set. Detailed training configuration is provided in Appendix~\ref{sec:appendix_training_config}. Additional details on how we
ruled out alternative explanations for the loss spike and localized the
failure to MMS are reported in Appendix~\ref{sec:appendix_diagnosis}.

\subsection{Testing the Alignment-Amplification Law}
\label{sec:exp_scaling_law}

\begin{wrapfigure}[17]{r}{0.7\linewidth}
    \vspace{-1.0em}
    \centering
    \includegraphics[width=\linewidth]{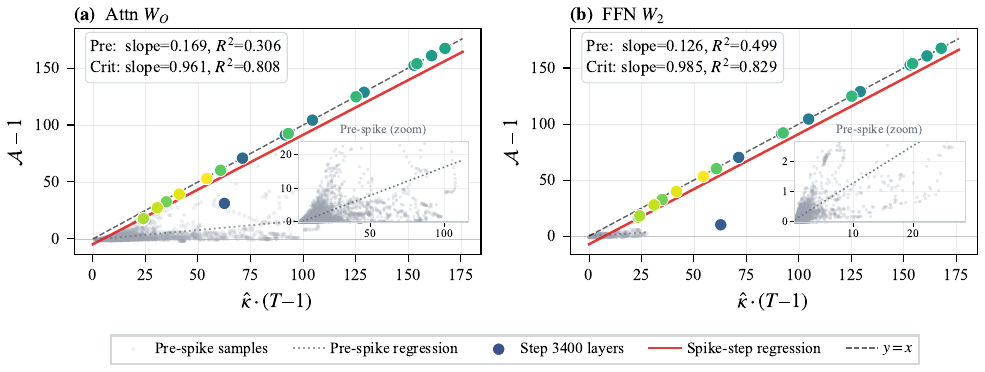}
    \caption{\small\textbf{Writer amplification at the gradient spike
    (400-layer Base $\eta$ run, $t^\star{=}3400$, measured on the $T{=}256$ image-token segment).}
    Each point plots $\mathcal{A}-1$ against the equal-magnitude
    absolute-coherence upper-envelope proxy $(T-1)\hat\kappa$ for
    \textbf{(a)} \texttt{Attn\_WO} and \textbf{(b)} \texttt{FFN\_W2}.
    Gray points are pre-spike layer-step samples; colored points are
    active layers at $t^\star$ ($\mathcal{A}-1>0.5$). The dashed line is
    the absolute-coherence saturation envelope; fitted slopes and $R^2$ values are shown in each panel.}
    \label{fig:scaling_law}
    \vspace{0.6em}
\end{wrapfigure}
Figure~\ref{fig:scaling_law} tests Eq.~\ref{eq:alignment_amp_exact} in a
representative unstable 400-layer run whose writer-gradient norm spikes
at step $t^\star=3400$. Before the spike, both writers lie well below
the saturation envelope. Absolute cross-token coherence is present, but the
signed off-diagonal terms in Eq.~\ref{eq:alignment_amp_exact} still
cancel. The small pre-spike slopes therefore measure how loose the
envelope is in this run, not new constants.

At step $t^\star$, the active layers lie close to the saturation envelope for
both \texttt{Attn\_WO} and \texttt{FFN\_W2}. The main observation is that
the spike occurs when signed cancellation at the residual writer largely
disappears. The same near-saturation appears in the attention and FFN
writers, supporting a writer-interface explanation rather than an
attention-specific one.

The largest active-layer values reach $\mathcal{A}-1\approx 167$, corresponding
to a $\sim 13\times$ writer-gradient norm amplification relative to the
independent-token baseline. The
shallowest active layer remains below the saturation envelope, consistent with
a boundary region where absolute coherence is already high but sign
cancellation has not fully disappeared. These measurements support the
mechanism in Section~\ref{sec:mechanism_trigger}: MMS occurs when
residual writers lose signed cancellation across tokens, allowing the
mean-coherent update $\Delta W_\mu$ to approach its coherent
$\mathcal{O}(T)$ scaling regime.

\subsection{MV-Split Shifts the Stability-Constrained Quality Frontier}
\label{sec:exp_frontier}

\begin{wrapfigure}[27]{r}{0.7\linewidth}
    \vspace{-1.0em}
    \centering
    \includegraphics[width=\linewidth]{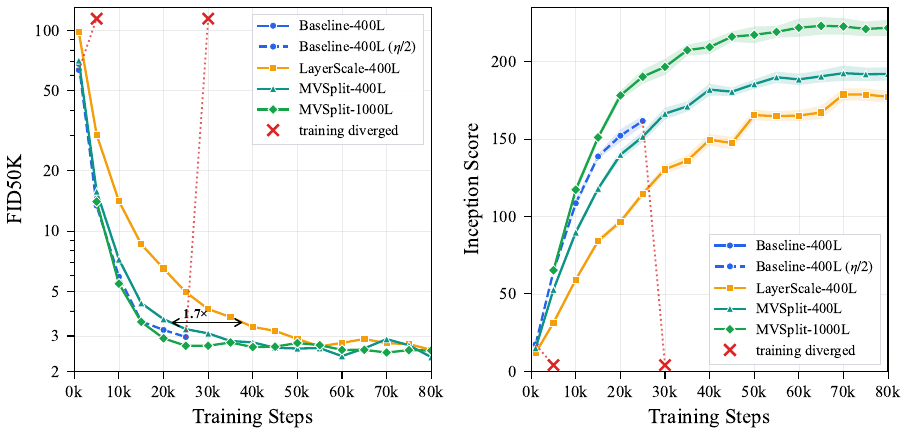}\\[-0.4em]
    \includegraphics[width=\linewidth]{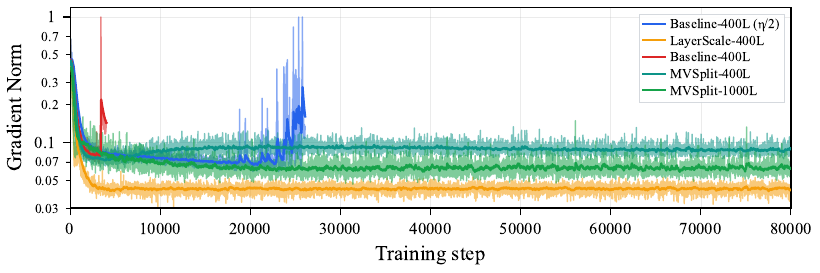}
    \caption{\small\textbf{Quality and optimizer stability over 80k steps
    (ImageNet $256{\times}256$).}
    \textbf{(Top)} FID-50K and Inception Score. \textbf{(Bottom)} Post-clipping
    global gradient norm. The 400-layer curves define the controlled
    comparison: among the non-divergent 400-layer runs, MV-Split preserves a
    higher bounded gradient band than LayerScale while avoiding the spikes of
    the un-stabilized baselines. The 1000-layer MV-Split trace is included as
    scale validation.}
    \label{fig:main_results}
    \vspace{0.6em}
\end{wrapfigure}

We next evaluate whether MV-Split changes the usable quality frontier under an
explicit stability constraint: a run is treated as usable only if it remains
non-divergent over the measured training horizon.

Figure~\ref{fig:main_results} and Table~\ref{tab:trajectory_summary} show the
resulting stability-constrained quality frontier. The un-stabilized baselines
are useful references for early learning speed, but they do not define stable
frontier points: both enter the mean-dominated failure state. Reducing the
learning rate delays this failure rather than removing it. LayerScale remains
stable over the measured horizon, but its token-isotropic per-channel gain also
reduces the centered residual updates needed for token-varying feature learning.

Under this stability constraint, MV-Split shifts the controlled 400-layer
frontier. It does not uniformly dominate the unstable baselines at early
checkpoints, but those trajectories leave the stable set; MV-Split preserves
much of their early convergence speed while avoiding their collapse. Among the
non-divergent 400-layer runs, MV-Split is already substantially ahead of
LayerScale by 20k--30k steps, and the added 40k/50k checkpoints show that this
advantage persists rather than reflecting a short early transient. The
gradient-norm trace also separates MV-Split from simple global shrinkage: it
operates in a higher bounded gradient band than LayerScale, while avoiding the
spikes seen in the un-stabilized runs.

\begin{wraptable}[20]{r}{0.64\linewidth}
    \vspace{-0.8em}
    \centering
    \caption{\small\textbf{Stability and convergence across 400-/1000-layer DiT runs.} The 400-layer rows define the matched stability-constrained comparison. The 1000-layer row is a separate scale-validation point and is not part of the matched 400-layer frontier comparison. FID-50K and IS are computed with Euler sampling, 25 NFE, and CFG scale $w=2.0$ for all rows. ``\textemdash'' denotes divergence before the checkpoint or failure to produce a valid evaluation. Bold highlights the best non-divergent result within the matched 400-layer comparison. The default-LR baseline diverges before the first checkpoint. $^\dagger$ The lower-LR baseline diverges later in training and is shown only as a pre-crash speed reference; it is not counted as a stable frontier point. 400L LayerScale reports the best stable member of the $\lambda_{\mathrm{init}}$ sweep.}
    \label{tab:trajectory_summary}
    \scriptsize
    \renewcommand{\arraystretch}{1.08}
    \setlength{\tabcolsep}{3pt}
    \resizebox{\linewidth}{!}{%
    \begin{tabular}{@{}lccccc@{}}
        \toprule
        & \multicolumn{5}{c}{\textbf{FID$\downarrow$\,/\,IS$\uparrow$}} \\
        \cmidrule(l){2-6}
        \textbf{Method} & @10k & @20k & @30k & @40k & @50k \\
        \midrule
        400L Base ($\eta$)              & ---                   & ---                   & ---                   & ---                   & --- \\
        400L Base ($\eta/2$)$^\dagger$  & 5.92 / 108.6          & 3.22 / 152.2          & ---                   & ---                   & --- \\
        400L LayerScale                 & 14.08 / 59.2          & 6.50 / 96.6           & 4.09 / 130.5          & 3.33 / 149.6          & 2.90 / 165.5 \\
        400L MV-Split                   & \textbf{7.23 / 89.8}  & \textbf{3.64 / 139.9} & \textbf{3.09 / 166.5} & \textbf{2.79 / 182.0} & \textbf{2.60 / 185.5} \\
        \midrule
        1000L MV-Split                  & 5.47 / 117.3          & 2.92 / 178.2          & 2.68 / 196.6          & 2.64 / 209.4          & 2.77 / 217.3 \\
        \bottomrule
    \end{tabular}%
    }
    \vspace{0.6em}
\end{wraptable}

The 1000-layer run extends this observation to boundary depth. The same
residual design remains stable over the measured training horizon and reaches
strong fixed-checkpoint FID/IS values at the reported boundary depth. Because
this run uses a separate training and post-training pipeline, we do not use it
as a matched frontier point against the 400-layer controls. Instead, it serves
as scale validation: the residual mechanism that shifts the controlled
400-layer frontier remains usable at 1000 layers.
Additional GenEval and DPG-Bench measurements for the post-trained checkpoint
are reported in Appendix~\ref{sec:appendix_t2i_eval} as calibration rather
than as state-of-the-art comparison.

\FloatBarrier
\subsection{Writer-Gradient Mode Decomposition}
\label{sec:exp_gmd_control}

\begin{wrapfigure}[20]{r}{0.7\linewidth}
    \vspace{-0.5em}
    \centering
    \includegraphics[width=\linewidth]{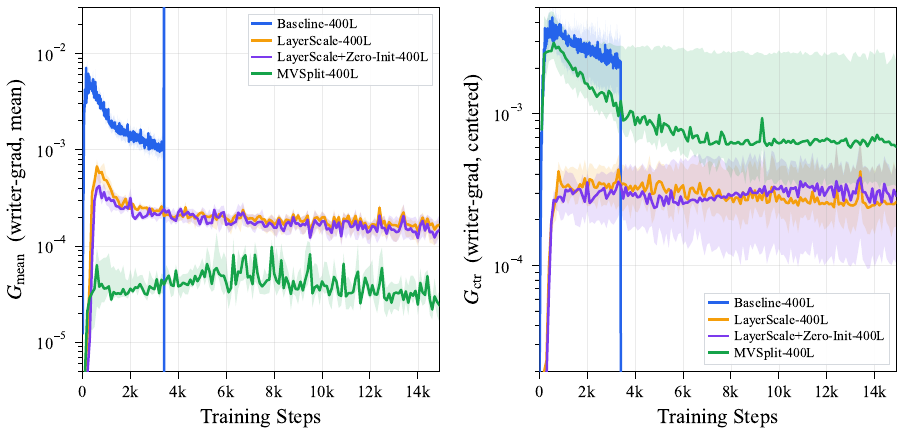}
    \caption{\small\textbf{Residual-writer gradient mode decomposition.}
    Per-step median across depth of the mean-coherent ($G_{\text{mean}}$, left)
    and centered ($G_{\text{ctr}}$, right) writer-gradient magnitudes; shaded
    regions denote the interquartile range (IQR; 25--75\% across depth). Token-isotropic per-channel gating compresses both
    modes; MV-Split bounds the mean-coherent component while preserving a
    higher, stable centered band.}
    \label{fig:gmd_control}
    \vspace{-0.5em}
\end{wrapfigure}
The convergence curves alone do not distinguish mode-selective control from a
smaller effective learning rate. We therefore measure the two writer-gradient
components from Eq.~\ref{eq:writer-gradient-decomposition}: the mean-coherent
component $G_{\text{mean}}$ and the centered component $G_{\text{ctr}}$.

Figure~\ref{fig:gmd_control} shows that LayerScale bounds the mean-coherent
writer component, but does so by shrinking the centered component as well.
This is expected from a token-isotropic residual gain: the same per-channel
multiplier is applied before any token-space split, so the method provides no
explicit mechanism to preserve centered variation while damping the token-mean component. The resulting low centered-gradient band
is consistent with the slower convergence observed in
Figure~\ref{fig:main_results}.

MV-Split changes this pattern. The mean-coherent component remains bounded,
while the centered component stays in a higher stable band. This supports the
intended mechanism of Eq.~\ref{eq:mv_backward}: the mean and centered
components receive separate gains at the residual merge. Thus the improved
stability in Section~\ref{sec:exp_frontier} is not explained by uniformly
smaller gradients, but by damping the writer-gradient mode associated with the
collapse.

\textbf{Deferred analyses.} Beyond stability, a linear probe confirms
\textbf{the token mean acts as an implicit global timestep carrier}
(near-perfect $R^2$
predicting $t$ across depth), justifying our design to gain-limit rather than
strictly project out the mean subspace
(Appendix~\ref{sec:appendix_timestep_probe}). Infrastructure-level optimizations for ultra-deep training are deferred to Appendix~\ref{sec:infra}.

\section{Related Work}
\label{sec:related-work}

\subsection{Deep Diffusion Transformers and Residual Stability}

Diffusion Transformers replace U-Net backbones~\cite{dhariwal2021diffusion}
with Transformer blocks over latent or image patches. DiT~\cite{peebles2023dit}
showed that increasing Transformer compute through depth, width, or token count
improves generative quality, while U-ViT~\cite{bao2023uvit} and
MMDiT/Stable Diffusion~3~\cite{esser2024sd3} demonstrate that token-based
diffusion backbones can support long skips, multimodal token mixing, and
rectified-flow text-to-image generation. Unlike standard DiT conditioning
stacks that inject the noise or timestep level through AdaLN or related
modulation paths, recent work suggests that explicit noise/timestep
conditioning is not always required for denoising generative
models~\cite{sun2025noiseconditioning,sahraee2026geometrynoise}. Our focus is
complementary to this objective-level question: we use a noise-agnostic
backbone to study a depthwise residual-stream failure mode in ultra-deep DiTs
and a residual merge that stabilizes this signal path.
Appendix~\ref{sec:appendix_timestep_probe} further shows that our trained
network implicitly carries the continuous timestep in the token-mean subspace.

Training instability in deep Transformers is often addressed by changing
normalization placement, residual scaling, or residual connectivity. Post-LN
Transformers can require warmup because large gradients appear near the output
layers at initialization, whereas Pre-LN changes this gradient
geometry~\cite{xiong2020layernorm}. Admin~\cite{liu2020admin} attributes
instability to residual-branch dependence that amplifies update perturbations.
ReZero~\cite{bachlechner2021rezero}, LayerScale~\cite{touvron2021layerscale},
DeepNorm~\cite{wang2022deepnet}, and Keel~\cite{chen2026postlayernorm}
stabilize depth by gating, rescaling, or altering the residual/carry path, with
DeepNorm and Keel reporting 1{,}000-layer-scale Transformer training. These
methods control the residual branch or carry path as a whole. MV-Split targets
a different axis: it combines a separately gained centered residual update with a leaky trunk-mean replacement, damping the mean-coherent channel without applying the same shrinkage to centered token variation.

MMS is also related to work on training spikes and attention/representation
collapse, but its diagnostic object is different. Loss-spike and proxy studies
connect large-scale instabilities to sub-layer Jacobian bounds, attention-logit
growth, output-logit divergence, and predictive activation/gradient-norm
trends~\cite{takase2024spikenomore,wortsman2024proxies}. Other work shows that
self-attention can drive token representations toward rank-one uniformity with
depth~\cite{dong2021attention}, that rank collapse can vanish Q/K gradients
via signal-propagation arguments~\cite{noci2022signalprop}, and that low
attention entropy is associated with unstable or divergent Transformer
training~\cite{zhai2023sigmareparam}. MMS, in contrast, is diagnosed through an
exact gradient decomposition into mean-coherent and centered
components, followed by a mean-dominated forward state in which centered token
variation is suppressed and Q/K learning is reduced through the Softmax
Jacobian null space. We therefore do not claim that MMS explains all
deep-Transformer spikes; it identifies a specific residual-subspace failure
pathway in ultra-deep Post-Norm DiTs, and MV-Split acts at the residual
interface rather than as a global residual shrinkage or attention-operator
correction. Appendix~\ref{sec:appendix_negative_results} reports negative
controls that test several superficially related alternatives.

\section{Conclusion}
We study a \textbf{mean-dominated collapse state} that limits depth scaling in very deep Diffusion Transformers, and we use \textbf{Mean Mode Screaming (MMS)} for the abrupt writer-gradient event that accompanies entry into this state in zero-writer training runs. The main mechanism is an imbalance between a mean-coherent writer update that can grow as $\mathcal{O}(T)$ and a centered path that is not sufficiently replenished once deep layers become contractive. MV-Split addresses this by combining a separately gained centered residual update with a leaky trunk-mean replacement. Under matched-backbone stabilizer protocols at 400 layers, MV-Split removes collapse events and gives the best stable frontier among the methods we evaluate; a separate 1000-layer run shows that the same design remains trainable at that depth.

{
\small
\bibliographystyle{unsrt}
\bibliography{references}
}

\clearpage
\appendix

\section{Diagnostic Metrics and Definitions}
\label{sec:appendix_metrics}

Table~\ref{tab:diagnostics_full} provides the mathematical definitions for all diagnostic metrics referenced in our analysis. Spatially coherent metrics are estimated robustly on sampled token subsets during the live training pass.

\begin{table}[h]
    \centering
    \caption{\textbf{Diagnostic Metrics Glossary.} Definitions for all diagnostic metrics referenced in our analysis. $X$ denotes forward representations; $A$ denotes attention matrices; $\Delta$ denotes backward gradients.}
    \footnotesize
    \renewcommand{\arraystretch}{1.4}
    \begin{tabularx}{\linewidth}{l l >{\raggedright\arraybackslash}X}
        \toprule
        \textbf{Metric} & \textbf{Formal Definition} & \textbf{Description} \\
        \midrule
        \textbf{Writer GMD} ($G_{\text{mean}}, G_{\text{ctr}}$)
        & $\|\Delta W_{\mu}\|_F$, $\|\Delta W_{c}\|_F$
        & Frobenius norms of the matrix components $\Delta W_{\mu} = T\bar{\delta}\bar{y}^\top$ and $\Delta W_{c} = \sum_t \tilde{\delta}_t \tilde{y}_t^\top$, decoupling the mean-coherent and centered components of the writer-weight update. \\

        \textbf{Q/K Grad Norm} ($G(Q), G(K)$)
        & $\text{RMS}(\nabla_{W_{Q, K}} \mathcal{L})$
        & Root-mean-square of the gradients with respect to the query and key projection weights. \\

        \textbf{Energy Ratio} ($\rho_T$)
        & $\frac{\|\mu(X)\|_F}{\|c(X)\|_F + \epsilon}$
        & Ratio between the energy of the mean component $\mu(X)$ and the centered component $c(X)$ of the token representation. \\

        \textbf{TR Ratio} ($r_{\text{TR}}$)
        & $\frac{\|U(X)\|_F}{\|X\|_F + \epsilon}$
        & Ratio between the residual-branch update $U(X)$ and the residual-stream state $X$. \\

        \textbf{Variance Gain} (VarGain)
        & $\frac{\|c(U(X))\|_F}{\|c(X)\|_F + \epsilon}$
        & Ratio between the centered energy of the branch update $c(U(X))$ and the centered energy of the input $c(X)$. \\

        \textbf{Attn Contraction} ($\mu_{\text{eff}}$)
        & $\|PAP\|_2$ \scriptsize{(centered power iter.)}
        & Spectral norm of the attention operator restricted to the centered subspace, where $P=I-J$ projects out the token mean. \\

        \textbf{Row Diversity} ($\text{RowDiv}$)
        & $\frac{\|A - JA\|_F}{\|A\|_F}$
        & Relative deviation of the attention rows from their column-mean profile. \\

        \textbf{Centered Retention} ($\text{Ret}(c \leftarrow c)$)
        & $\frac{\|c(A \cdot c(X))\|_F}{\|c(X)\|_F + \epsilon}$
        & Fraction of centered input energy that remains in the centered subspace after one attention operation. \\
        \textbf{Mean Leakage} ($\text{Leakage}(\mu \leftarrow c)$)
        & $\frac{\|\mu(A \cdot c(X))\|_F}{\|c(X)\|_F + \epsilon}$
        & Fraction of centered input energy mapped into the mean subspace by the attention operator via $JAP$. \\
        \textbf{Token Cosine Similarity} (TCS)
        & $\mathbb{E}_{i\ne j}[\cos(X_i, X_j)]$
        & Average pairwise cosine similarity between token representations, estimated on sampled token pairs when needed. High values indicate token homogenization. \\
        \bottomrule
    \end{tabularx}
    \label{tab:diagnostics_full}
\end{table}

\FloatBarrier

\clearpage
\section{Standard Initialization Enters the Same Mean-Dominated State}
\label{sec:appendix_standard_init}

The main diagnostic runs use zero-initialized residual writers, a standard
identity-start choice for deep residual networks. This choice keeps the early
trajectory well behaved and makes the delayed MMS event easy to isolate. We now
ask how the same backbone behaves when residual writers are initialized open
from the first step. To test this, we train a 128-layer single-stream DiT with
no residual gating and Gaussian initialization,
$\mathcal{W}\sim\mathcal{N}(0,0.02^2)$~\cite{shoeybi2019megatron}, for the
residual writers and final projection. This control is not a matched convergence
comparison; it tests whether the mean-dominated state is specific to the
identity-start writer schedule.

\begin{figure}[htbp]
    \centering
    \includegraphics[width=\linewidth]{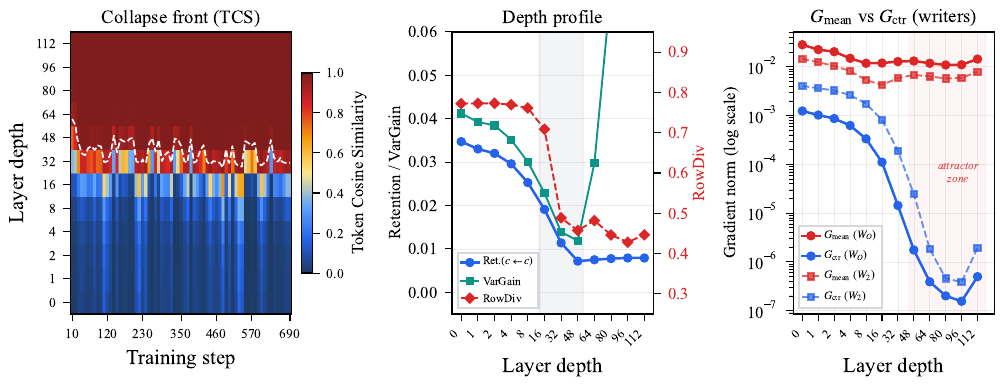}
    \caption{\textbf{Standard-initialization control for a 128-layer DiT.}
    \textbf{(a)} Token cosine similarity (TCS) over training steps and layer
    depth. The dashed white contour marks $\mathrm{TCS}=0.9$.
    \textbf{(b)} Depth profiles of centered retention
    $\mathrm{Ret}(c{\leftarrow}c)$, centered branch replenishment VarGain, and
    attention row diversity RowDiv; curves report the median over diagnostic
    checkpoints from steps 10--690.
    \textbf{(c)} Median writer-gradient decomposition for the attention output
    projection $W_O$ and FFN output projection $W_2$, reporting
    $G_{\mathrm{mean}}$ and $G_{\mathrm{ctr}}$ across depth.}
    \label{fig:collapse_front}
\end{figure}

Figure~\ref{fig:collapse_front} shows that standard initialization enters the
same mean-dominated regime. The temporal pattern differs from the zero-writer
runs: instead of a delayed writer-gradient spike after an initially stable
period, deep layers have high token similarity from the beginning of training,
and the high-similarity region forms a depth-wise collapse front. In this run,
the loss quickly reaches a high plateau, consistent with a residual stream that
has lost most of its token-varying information in deep layers.

The forward diagnostics connect this behavior to the same subspace mechanism
studied in the main text. In deep layers, centered retention and branch-side
centered replenishment are both small, so centered variation is not maintained
through depth. RowDiv remains nonzero, ruling out the simpler explanation that
attention has literally collapsed to identical rows. The failure is instead a
subspace imbalance: row-stochastic attention preserves pure-mean states, while the centered component is weakly retained and weakly
replenished.

The writer-gradient decomposition shows the same imbalance on the backward
path. In the deep collapsed layers, the mean-coherent writer component
$G_{\mathrm{mean}}$ dominates the centered component $G_{\mathrm{ctr}}$ by
several orders of magnitude for both $W_O$ and $W_2$. Thus the standard-init
run reaches the same endpoint as the zero-writer MMS runs: token variation is
suppressed and residual writer updates become mean-dominated.

Writer initialization therefore changes the temporal presentation of the
failure, not the subspace-level failure mode. Identity-start runs expose the
failure as a delayed MMS event, while standard initialization exposes it as an
early depth-wise collapse front. In both cases, the common failure channel is
the imbalance between the invariant pure-mean direction and insufficient
centered-subspace maintenance. MV-Split therefore targets the residual writer
geometry rather than a peculiarity of the zero-writer schedule.

\clearpage
\section{Derivations of Writer Gradient Scaling}
\label{sec:appendix_scaling_derivation}

\subsection{Proof of the Gradient Mode Decomposition}
\label{sec:appendix_gmd_proof}

In Section~\ref{sec:mechanism_trigger}, we state that the gradient admits an exact additive decomposition into a rank-1 mean-coherent component $\Delta W_\mu$ and a centered variation component $\Delta W_c$, with cross-terms identically vanishing. We provide the brief algebraic proof here.

Let the forward input $y_t \in \mathbb{R}^n$ and the backward gradient $\delta_t \in \mathbb{R}^m$ for token $t$ be decomposed into their sequence means and zero-mean centered components:
\begin{align}
    y_t &= \bar{y} + \tilde{y}_t, \quad \text{where} \quad \bar{y} = \frac{1}{T}\sum_{t=1}^T y_t, \quad \sum_{t=1}^T \tilde{y}_t = \mathbf{0}, \\
    \delta_t &= \bar{\delta} + \tilde{\delta}_t, \quad \text{where} \quad \bar{\delta} = \frac{1}{T}\sum_{t=1}^T \delta_t, \quad \sum_{t=1}^T \tilde{\delta}_t = \mathbf{0}.
\end{align}

The parameter gradient for a token-wise linear map $W$ is the sum of outer products over the sequence length. Substituting the decomposed terms yields:
\begin{equation}
    \nabla_W \mathcal{L}
    \;=\; \sum_{t=1}^T \delta_t y_t^\top
    \;=\; \sum_{t=1}^T \bigl(\bar{\delta} + \tilde{\delta}_t\bigr) \bigl(\bar{y} + \tilde{y}_t\bigr)^\top.
\end{equation}

Expanding the outer product gives four summation terms:
\begin{equation}
    \nabla_W \mathcal{L}
    \;=\; \sum_{t=1}^T \bar{\delta}\bar{y}^\top
    \;+\; \sum_{t=1}^T \bar{\delta}\tilde{y}_t^\top
    \;+\; \sum_{t=1}^T \tilde{\delta}_t\bar{y}^\top
    \;+\; \sum_{t=1}^T \tilde{\delta}_t\tilde{y}_t^\top.
\end{equation}

Because the sequence means $\bar{y}$ and $\bar{\delta}$ are constant across tokens, they can be factored out of the summations for the cross-terms:
\begin{align}
    \sum_{t=1}^T \bar{\delta}\tilde{y}_t^\top &= \bar{\delta} \underbrace{\biggl( \sum_{t=1}^T \tilde{y}_t^\top \biggr)}_{=\,\mathbf{0}^\top} = \mathbf{0}, \\
    \sum_{t=1}^T \tilde{\delta}_t\bar{y}^\top &= \underbrace{\biggl( \sum_{t=1}^T \tilde{\delta}_t \biggr)}_{=\,\mathbf{0}} \bar{y}^\top = \mathbf{0}.
\end{align}

The cross-terms evaluate identically to zero matrices, and the first term sums to $T\,\bar{\delta}\,\bar{y}^\top$. The gradient therefore admits an exact additive decomposition into a mean-coherent rank-1 component ($\Delta W_\mu$) and a centered component ($\Delta W_c$):
\begin{equation}
    \nabla_W \mathcal{L} \;=\; \underbrace{T\,\bar{\delta}\,\bar{y}^\top}_{\Delta W_{\mu}} \;+\; \underbrace{\sum_{t=1}^T \tilde{\delta}_t \tilde{y}_t^\top}_{\Delta W_{c}}.
\end{equation}
This recovers Equation~\ref{eq:writer-gradient-decomposition} as an algebraic identity rather than an approximation.

\subsection{Derivation of the Alignment-Amplification Law}
\label{sec:appendix_alignment_law}

We derive Eq.~(\ref{eq:alignment_amp_exact}) and its equal-magnitude specialization.

\paragraph{Exact expansion and diagonal--off-diagonal split.}
Let $W$ be a token-wise linear map with gradient
\begin{equation}
  \nabla_W\mathcal{L}
  \;=\;
  \sum_{t=1}^{T}\delta_t y_t^{\top}
  \;\in\;\mathbb{R}^{m\times n},
\end{equation}
where $y_t\in\mathbb{R}^{n}$ is the forward input and
$\delta_t\in\mathbb{R}^{m}$ is the corresponding backward gradient for
token $t$. Using the Frobenius inner-product identity for rank-1 matrices,
\[
\langle ab^\top,\;cd^\top\rangle_F
=
\langle a,c\rangle\,\langle b,d\rangle,
\]
we obtain
\begin{equation}
  \|\nabla_W\mathcal{L}\|_F^{2}
  \;=\;
  \Big\langle \sum_{t=1}^{T}\delta_t y_t^\top,\;
              \sum_{s=1}^{T}\delta_s y_s^\top \Big\rangle_F
  \;=\;
  \sum_{t=1}^{T}\sum_{s=1}^{T}
  \langle y_t,y_s\rangle\,\langle \delta_t,\delta_s\rangle.
  \label{eq:apx-frob-expansion}
\end{equation}
Separating diagonal and off-diagonal terms gives
\begin{equation}
  \|\nabla_W\mathcal{L}\|_F^{2}
  \;=\;
  \underbrace{\sum_{t=1}^{T}\|y_t\|^{2}\|\delta_t\|^{2}}_{S}
  \;+\;
  \underbrace{\sum_{s\ne t}\langle y_t,y_s\rangle\,
  \langle \delta_t,\delta_s\rangle}_{C}.
  \label{eq:apx-diag-off}
\end{equation}
Here $S$ is the diagonal, no-interaction baseline, while $C$ collects the
cross-token interference terms.

\paragraph{Rayleigh-quotient form.}
Define the per-token magnitude
\begin{equation}
  w_t \triangleq \|\delta_t\|\,\|y_t\|,
  \qquad
  w \in \mathbb{R}_{\ge 0}^{T},
\end{equation}
and the pairwise alignment matrix
\begin{equation}
  M_{ts}
  \triangleq
  \cos(y_t,y_s)\,\cos(\delta_t,\delta_s).
\end{equation}
Then Eq.~(\ref{eq:apx-frob-expansion}) becomes
\begin{equation}
  \|\nabla_W\mathcal{L}\|_F^{2}
  \;=\;
  w^\top M w,
\end{equation}
while the diagonal baseline is
\begin{equation}
  S = \sum_{t=1}^{T} w_t^2 = \|w\|_2^2,
\end{equation}
since $M_{tt}=1$. Therefore the alignment amplification (cf.\ main text Eq.~\ref{eq:alignment_amp_exact}) is
\begin{equation}
  \mathcal{A}
  \;\triangleq\;
  \frac{\|\nabla_W\mathcal{L}\|_F^{2}}{S}
  \;=\;
  \frac{w^\top M w}{\|w\|_2^2},
\end{equation}
and subtracting the diagonal baseline yields
\begin{equation}
  \mathcal{A}-1
  \;=\;
  \frac{w^\top(M-I)w}{\|w\|_2^2}
  \;=\;
  \frac{\sum_{s\ne t} w_t w_s M_{ts}}
       {\sum_{t=1}^{T} w_t^2}.
  \label{eq:apx-exact}
\end{equation}
This is exactly Eq.~(\ref{eq:alignment_amp_exact}) in the main text.

Moreover, $M$ is positive semidefinite. Indeed, it is the Hadamard product
of the cosine Gram matrix of $\{y_t\}$ and that of $\{\delta_t\}$; both are
positive semidefinite, and the Schur Product Theorem preserves positive
semidefiniteness. Thus $\mathcal{A}$ is a Rayleigh quotient of a PSD matrix.

\paragraph{Equal-magnitude specialization.}
Suppose the per-token magnitude is approximately constant,
\begin{equation}
  w_t \equiv w_0.
\end{equation}
Then Eq.~(\ref{eq:apx-exact}) reduces to
\begin{equation}
  \mathcal{A}-1
  \;=\;
  \frac{1}{T}\sum_{s\ne t} M_{ts}.
\end{equation}
Define
\begin{equation}
  \kappa
  \triangleq
  \mathbb{E}_{s\ne t}[M_{ts}]
  \;=\;
  \mathbb{E}_{s\ne t}\bigl[\cos(y_t,y_s)\cos(\delta_t,\delta_s)\bigr],
\end{equation}
where $\mathbb{E}_{s\ne t}$ denotes the uniform average over ordered pairs
$(s,t)$ with $s\ne t$. Then
\begin{equation}
  \mathcal{A}-1
  \;=\;
  (T-1)\,\kappa.
  \label{eq:scaling_law}
\end{equation}
When $\kappa \approx 0$, the cross-terms cancel on average and
\[
\|\nabla_W\mathcal{L}\|_F^{2} = \mathcal{O}(T),
\qquad
\|\nabla_W\mathcal{L}\|_F = \mathcal{O}(\sqrt{T}).
\]
When $\kappa \to 1$, the accumulation becomes coherent and
\[
\|\nabla_W\mathcal{L}\|_F^{2} = \mathcal{O}(T^2),
\qquad
\|\nabla_W\mathcal{L}\|_F = \mathcal{O}(T).
\]

\paragraph{Absolute-coherence upper bound.}
From Eq.~(\ref{eq:apx-exact}),
\begin{equation}
  |\mathcal{A}-1|
  \;\le\;
  \frac{\sum_{s\ne t} w_t w_s |M_{ts}|}
       {\sum_{t=1}^{T} w_t^2}.
\end{equation}
Under the equal-magnitude approximation, this becomes
\begin{equation}
  |\mathcal{A}-1|
  \;\le\;
  (T-1)\,\hat\kappa,
  \qquad
  \hat\kappa
  \triangleq
  \mathbb{E}_{s\ne t}
  \bigl[
    |\cos(y_t,y_s)|\,|\cos(\delta_t,\delta_s)|
  \bigr].
\end{equation}
In the main-text experiment, $\hat\kappa$ is used as an absolute-coherence
proxy. The gap between $(T-1)\hat\kappa$ and the signed quantity
$\mathcal{A}-1$ reflects signed cancellation, together with any looseness
introduced by the absolute-value relaxation.

\subsection{Proof of Lemma~\ref{lem:softmax-nullspace}}
\label{sec:appendix_qk_extinction}

Let $a_i\in\mathbb{R}^T$ denote the $i$-th attention row written as a
column vector, so that
\begin{equation}
  Y_i \;=\; \sum_{j=1}^{T} a_{ij}V_j,
  \qquad
  a_i \;=\; \operatorname{softmax}(S_i).
\end{equation}
By the chain rule,
\begin{equation}
  \frac{\partial \mathcal{L}}{\partial a_{ij}}
  \;=\;
  \Bigl\langle \frac{\partial \mathcal{L}}{\partial Y_i},\,V_j \Bigr\rangle.
\end{equation}
If $V_j=\bar v$ for all $j$, then the right-hand side is independent of
$j$. Hence
\begin{equation}
  \frac{\partial \mathcal{L}}{\partial a_i}
  \;=\;
  \gamma_i \mathbf{1},
  \qquad
  \gamma_i \triangleq
  \Bigl\langle \frac{\partial \mathcal{L}}{\partial Y_i},\,\bar v \Bigr\rangle.
\end{equation}

The softmax Jacobian at $a_i$ is
\begin{equation}
  J_{\mathrm{sm}}(a_i)
  \;=\;
  \operatorname{diag}(a_i) - a_i a_i^\top,
\end{equation}
and satisfies
\begin{equation}
  J_{\mathrm{sm}}(a_i)\mathbf{1}
  \;=\;
  a_i - a_i(\mathbf{1}^\top a_i)
  \;=\;
  \mathbf{0},
\end{equation}
since $\mathbf{1}^\top a_i = 1$. Exploiting the symmetry of the Softmax Jacobian,
\begin{equation}
  \frac{\partial \mathcal{L}}{\partial S_i}
  \;=\;
  J_{\mathrm{sm}}(a_i)^\top\frac{\partial \mathcal{L}}{\partial a_i}
  \;=\;
  J_{\mathrm{sm}}(a_i)\frac{\partial \mathcal{L}}{\partial a_i}
  \;=\;
  \gamma_i\,J_{\mathrm{sm}}(a_i)\mathbf{1}
  \;=\;
  \mathbf{0}.
\end{equation}
Since this holds for every row $i$, the logit gradient vanishes
identically, which proves Lemma~\ref{lem:softmax-nullspace}.

\paragraph{Residual-writer gradients bypass the Softmax null space.}
The null-space argument above concerns only the gradient through the attention
logits $S_i$, and therefore the Q/K pathway. It does not zero the attention
output projection. If $H_i=\sum_j a_{ij}V_j$ denotes the pre-$W_O$ attention
output and $g_i$ is the upstream adjoint at the output projection, then
\begin{equation}
  \nabla_{W_O}\mathcal L=\sum_i g_i H_i^\top,
\end{equation}
which bypasses the Softmax Jacobian. Under value homogenization,
$H_i\to\bar H$ becomes approximately token-constant, so the writer gradient
becomes mean-coherent rather than zero. Gradients to the value pathway can
also remain nonzero: the strict null-space extinction applies to the logit,
and hence to the Q/K pathway, only.

\clearpage
\section{Detailed Comparison: MV-Split vs. LayerScale and ReZero}
\label{sec:appendix_layerscale}

LayerScale~\cite{touvron2021layerscale} parameterizes the merge as
$Z_l^{\text{LS}}=X_l+\lambda_l\odot F_l$ with
$\lambda_l\in\mathbb{R}^D$ a per-channel learnable vector;
ReZero~\cite{bachlechner2021rezero} is the single-scalar special
case $\lambda_l\in\mathbb{R}$ initialized at zero. We show that
both differ from MV-Split in three structural respects, each of
which corresponds to a specific failure mode in
Sec.~\ref{sec:mechanism}.

\paragraph{1. Open-loop vs.\ leaky-integrator mean dynamics.}
Projecting the two merges into the mean subspace via $J$:
\begin{equation}
  J Z_l^{\text{LS}} = J X_l + (\lambda_l\odot J F_l),
  \qquad
  J Z_l^{\text{MV}} = (1-\alpha)\odot J X_l + \alpha\odot J F_l.
\end{equation}
LayerScale leaves the trunk's mean component
\emph{untouched at every layer} (the coefficient of $JX_l$ is
identically $1$): it does not damp the carried trunk mean and only scales newly injected branch updates.
MV-Split contracts the trunk's mean by $1-\alpha$ before each
injection, which is a leaky integrator whenever $\alpha_d\in(0,1)$.
The two cannot be made equivalent by any choice of $\lambda_l$:
taking $F_l\equiv 0$ gives $JZ_l^{\text{LS}}=JX_l$ while
$JZ_l^{\text{MV}}=(1-\alpha)\odot JX_l$, so the dynamics differ
whenever $\alpha\ne 0$, irrespective of $\lambda_l$.

\paragraph{2. Isotropic vs.\ anisotropic gain on the residual branch.}
By Eq.~\ref{eq:writer-gradient-decomposition} the
gradient decomposes as $\nabla_W\mathcal{L}=\Delta W_\mu+\Delta W_c$
with $\|\Delta W_\mu\|_F\sim T\hat\kappa$ in the coherent regime and
$\|\Delta W_c\|_F$ scaling diffusively under weak centered alignment. In the scalar-gain simplification, both modes are scaled by the same gain:
\begin{equation}
  \frac{\|\Delta W_\mu^{\text{LS}}\|_F}{\|\Delta W_c^{\text{LS}}\|_F}
  \;\propto\;\sqrt{T}\hat\kappa.
\end{equation}
For scalar gates like ReZero, the ratio is exactly invariant. LayerScale uses a per-channel diagonal operator $\lambda_l\in\mathbb{R}^D$; while feature-wise anisotropy can incidentally alter the mean/centered ratio, it provides no structural token-subspace filter.
MV-Split scales the two modes by $\alpha$ and $\beta$ independently. In the scalar-gain simplification:
\begin{equation}
  \frac{\|\Delta W_\mu^{\text{MV}}\|_F}{\|\Delta W_c^{\text{MV}}\|_F}
  \;\propto\;\frac{\alpha}{\beta}\sqrt{T}\hat\kappa,
\end{equation}
allowing the unstable-to-stable ratio $\alpha/\beta$ to be reduced
without coupling to the absolute centered gain $\beta$. If the leaky mean
replacement term were removed (i.e.\ replacing $\alpha\odot J(F_l-X_l)$ in
Eq.~\ref{eq:mv_merge} by $\alpha\odot JF_l$), the special case $\alpha=\beta$
would reduce to a LayerScale-like token-space isotropic branch gain. With the
leaky term in Eq.~\ref{eq:mv_merge}, however, MV-Split remains dynamically
distinct from LayerScale even when $\alpha=\beta$, because $JZ_l^{\text{MV}}=
(1-\alpha)\odot JX_l+\alpha\odot JF_l$ contracts the trunk's mean component at
every layer (cf.\ \S 1 above), whereas LayerScale leaves it untouched.

\paragraph{3. Independent gain on the centered path.}
Whatever absolute gain on the centered branch is needed for
stability at a given depth, MV-Split treats it as a free parameter
set independently of $\alpha$ (Eq.~\ref{eq:mv_dynamics}). LayerScale ties the two
paths to the same token-independent per-channel gain, so any reduction in the
mean-coherent contribution unavoidably reduces centered
replenishment by the same factor.

Sec.~\ref{sec:mechanism} describes a self-reinforcing failure: a gradient
spike injects mean-mode content into the trunk, the trunk's mean direction
aligns with the residual branch's coherent direction, and this alignment amplifies the
next mean-coherent update. The two MV-Split gates intervene at two different
points: $\alpha$ contracts the trunk's mean component at every layer
(Eq.~\ref{eq:mv_dynamics}), bounding mean accumulation; the $\alpha/\beta$ gap
damps the mean-coherent update relative to the centered update
(Eq.~\ref{eq:mv_backward}), interrupting the alignment-amplification step. A scalar gate scales both modes equally, while LayerScale applies no explicit token-subspace filter; neither contracts the carried trunk mean.

\clearpage
\section{Segment-wise Projectors for Multimodal Sequences}
\label{sec:appendix_mv_split_seg}

For a sequence partitioned into image tokens $\mathcal{I}$ and text
tokens $\mathcal{T}$, we define group-mean projectors that average
within each segment only:
\begin{equation}
  J_{\text{seg}} = \mathrm{blkdiag}(J_{\mathcal{I}}, J_{\mathcal{T}}),
  \qquad
  P_{\text{seg}} = I - J_{\text{seg}},
\end{equation}
where $J_{\mathcal{I}}=\frac{1}{|\mathcal{I}|}\mathbf{1}_{\mathcal{I}}
\mathbf{1}_{\mathcal{I}}^\top$ and similarly for $J_{\mathcal{T}}$.
The projector does not average across modalities; it prevents the residual merge from directly mixing image and text means through the mean operator, preserving modality-specific mean scales in the residual control path. The diagnostic and mechanistic derivations in the main text use the global sequence-mean projector; the segment-wise projector is used only in the multimodal MV-Split residual merge.

\paragraph{Segment-wise control still acts on the global mean mode.}
Let $J_g=\frac{1}{T}\mathbf{1}\mathbf{1}^\top$ denote the global projector. The
global mean subspace is contained in the segment-wise mean subspace, so
\begin{equation}
  J_g\,J_{\mathrm{seg}} = J_g, \qquad J_g\,P_{\mathrm{seg}} = \mathbf{0}.
\end{equation}
Because the gates $\alpha,\beta$ are feature-wise and broadcast across tokens,
they commute with the token projectors. Applying $J_g$ to the
pre-normalization MV-Split merge with $J,P$ instantiated as $J_{\mathrm{seg}},P_{\mathrm{seg}}$ (Eq.~\ref{eq:mv_merge}) gives
\begin{equation}
  J_g Z_l
  \;=\;
  J_g X_l
  \;+\;
  \beta\odot J_g\,P_{\mathrm{seg}}F_l
  \;+\;
  \alpha\odot J_g\,J_{\mathrm{seg}}(F_l-X_l)
  \;=\;
  (1-\alpha)\odot J_g X_l + \alpha\odot J_g F_l.
\end{equation}
Thus the segment-wise implementation applies the same leaky control to the
global MMS mode that the main-text theory analyzes, while avoiding direct
averaging of image and text means in the residual control path.

\clearpage
\section{Step-Level Gradient Trace for Failure Attribution}
\label{sec:appendix_diagnosis}

The main text analyzes MMS through token-space and gradient
decompositions. This appendix describes the inline trace that audits the
representative gradient spike before those subspace diagnostics are applied:
it rules out data/loss-side artifacts, localizes where large gradients
appear, and identifies which internal quantities to measure next.

\begin{figure}[htbp]
    \centering
    \includegraphics[width=0.85\linewidth]{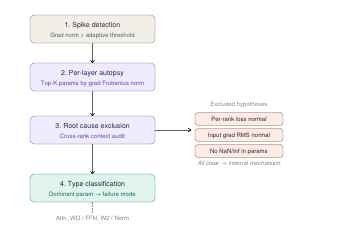}
    \caption{\textbf{Step-level gradient trace pipeline.} A global-norm threshold (1) triggers a per-family top-$K$ ranking of distributed gradient norms (2) and a cross-rank exclusion audit (3) that checks per-rank loss agreement, final-output-gradient RMS, and \texttt{NaN}/\texttt{Inf} in stored parameters. When all three exclusions pass, the dominant top-$K$ parameter family at the detected step (4) is recorded for the gradient-mode audit and subsequent paragraphs.}
    \label{fig:diagnostic_pipeline}
\end{figure}

\FloatBarrier
\paragraph{Trace protocol.}
\label{sec:diagnostic_protocol}
Figure~\ref{fig:diagnostic_pipeline} summarizes the pipeline. When the global
gradient norm crosses the threshold, the training loop records (i) per-rank
loss, loss weight, and output-gradient statistics, (ii) a
\texttt{NaN}/\texttt{Inf} scan over stored parameters, (iii) distributed
gradient norms grouped by layer and parameter family, and (iv) at instrumented
residual writers, the mean-coherent and centered components from
Eq.~\ref{eq:writer-gradient-decomposition}.

For a parameter family $\tau$ at layer $l$, the grouped norm aggregated across $R$ distributed ranks is
\begin{equation}
    G_{l,\tau}(t) = \left( \sum_{r=1}^{R} \sum_{\theta \in \Theta_{l,\tau}} \|\nabla_\theta \mathcal{L}_r(t)\|_F^2 \right)^{1/2}.
\end{equation}
This top-$K$ grouping localizes which parameter families receive large
gradients at the detected step; it does not identify the responsible
token-space mode.

\FloatBarrier
\paragraph{Data/loss-side checks.}
At the detected steps in the representative trace, per-rank losses remain clustered and the final-output-gradient statistics stay small at the printed precision. The maximum per-sample output-gradient norm is also nearly identical across ranks. This rules out two simple explanations: a single-rank data outlier and a global loss-weighting jump. The parameter scan finds no \texttt{NaN}/\texttt{Inf} values in stored parameters, so the event is not explained by persistent parameter corruption.

\begin{figure}[htbp]
    \centering
    \includegraphics[width=\linewidth]{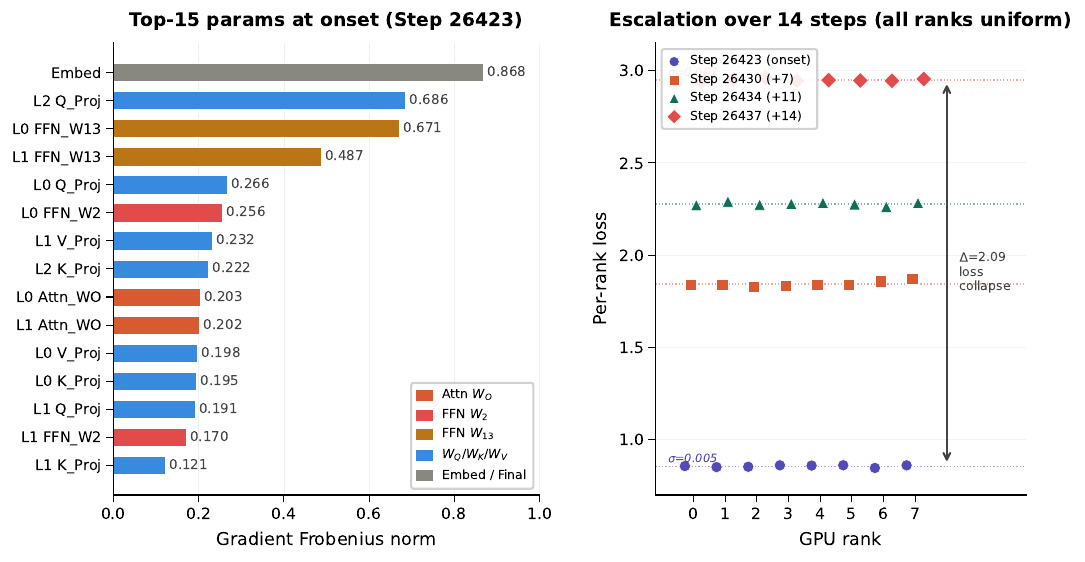}
    \caption{\textbf{Step-level gradient trace at a representative spike (400 layers, Base $\eta/2$).}
    \textbf{(Left)} Top parameter-family gradient norms $G_{l,\tau}$ at one detected step (Step~26423). The top-$K$ entries span embedding/final parameters, Q/K/V projections, FFN input weights, and residual output projections.
    \textbf{(Right)} Per-rank loss across four snapshots (Steps 26423, 26430, 26434, 26437). The eight ranks stay tightly clustered ($\sigma \in [0.005, 0.007]$) and the maximum per-sample output-gradient norm $M_{\mathrm{out}}^{(r)}\!\approx\!1{\times}10^{-4}$ matches across ranks (inset).}
    \label{fig:spike_baseline}
\end{figure}

\paragraph{Parameter-family localization.}
\label{sec:baseline_spike}
We examine the representative trace in the lower-learning-rate baseline (Base $\eta/2$, no MV-Split, no LayerScale), where divergence at Step~26423 offers a longer pre-spike window than the default-LR run in Section~\ref{sec:exp_scaling_law} (Base $\eta$, $t^\star{=}3400$). The first warning snapshots are mixed: their largest entries include embedding/final parameters, Q/K/V projections, FFN input weights, and residual output projections (Figure~\ref{fig:spike_baseline}, left). We therefore do not interpret these early warnings as a fixed shallow-layer mechanism. At the escalation step, the largest printed entries shift toward residual output interfaces, with \texttt{Attn\_WO} accounting for most of the top-$K$ squared-norm mass and \texttt{FFN\_W2} also appearing. This localization motivates auditing the residual writers directly rather than attributing the event to the output head, the batch, or a specific attention-logit pathology.

\FloatBarrier
\paragraph{Gradient-mode audit.}
For each instrumented writer, the trace caches the writer input $y_t$ during the forward pass and the output adjoint $\delta_t$ during the backward pass. It then computes
\begin{equation}
    \Delta W_\mu = T\bar\delta\bar y^\top, \qquad
    \Delta W_c = \sum_t \tilde\delta_t \tilde y_t^\top,
\end{equation}
and reports $G_{\mathrm{mean}}=\|\Delta W_\mu\|_F$ and $G_{\mathrm{ctr}}=\|\Delta W_c\|_F$. This is the measurement that links the top-$K$ localization to the mechanism in Section~\ref{sec:mechanism_trigger}: at the spike, the writer update is amplified in the mean-coherent component, while the centered component does not show a comparable increase.

\begin{figure}[htbp]
    \centering
    \includegraphics[width=\linewidth]{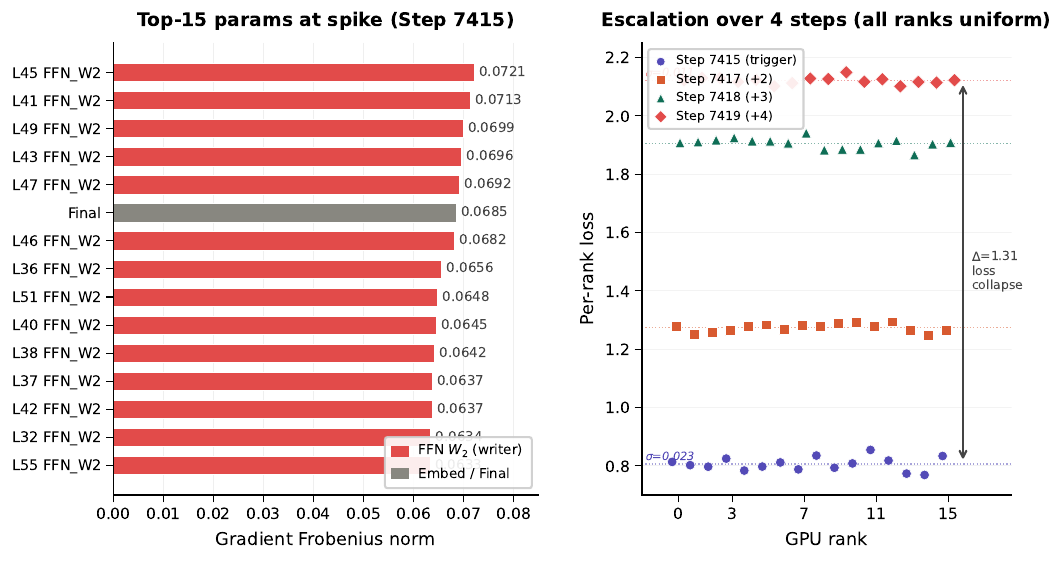}
    \caption{\textbf{Attention-branch-only MV-Split control (1000 layers).}
    \textbf{(Left)} Top parameter-family gradient norms at the detected step (Step~7415). With the attention output branch protected, no \texttt{Attn\_WO} entries appear in the top-$K$; the largest printed entries are \texttt{FFN\_W2}.
    \textbf{(Right)} Per-rank loss over four consecutive steps (7415--7419). Cross-rank losses stay tightly clustered ($\sigma \in [0.011, 0.023]$) while the loss rises uniformly from $\sim$0.81 to $\sim$2.12.}
    \label{fig:spike_attn_only}
\end{figure}

\FloatBarrier
\paragraph{Attention-branch-only control.}
\label{sec:partial_intervention}
To test whether protecting the attention branch alone is sufficient, we apply MV-Split only to the attention output residual branch and leave the FFN residual branch unchanged. We tested this on a 1000-layer configuration. The training still spikes at Step~7415 (Global Norm $\approx 0.665$), and the largest printed gradients move to the unprotected FFN branch: \texttt{Attn\_WO} disappears from the top-$K$ entirely, while \texttt{FFN\_W2} accounts for 14 of the top 15 contributors and about 93\% of the top-$K$ squared-norm mass at this step (Figure~\ref{fig:spike_attn_only}). Depending on the step, these entries can include both the FFN input transformation and the FFN output projection.

The subsequent escalation is rapid. Within four steps (7415$\to$7419), per-rank loss rises uniformly from $\sim$0.81 to $\sim$2.12 across all 16 ranks, with per-step standard deviation remaining low ($\sigma \in [0.011, 0.023]$), consistent with a global update event rather than a single-rank or single-batch fault.

The conclusion is therefore branch-level rather than weight-level: attention-only residual control is insufficient, and both attention and FFN residual branches require the mean/centered split.

\FloatBarrier

\clearpage
\section{Training Configuration Details}
\label{sec:appendix_training_config}

Table~\ref{tab:training_config} summarizes the architecture and optimization hyperparameters for the four DiT configurations reported in this paper: \textbf{DiT-400L-Baseline} (400-layer Post-Norm without residual gating, used to characterize the Mean-Mode Screaming failure), \textbf{DiT-400L-LayerScale} (the matched 400-layer LayerScale control), \textbf{DiT-400L-MVSplit} (the matched 400-layer model with MV-Split residuals), and \textbf{DiT-1000L-MVSplit} (the 1000-layer text-to-image scale-up demonstration). The base learning rate for each run is obtained from the target value via $\mu$P~\cite{yang2022tensor} width-scaling, $\text{base}=\text{target}/(0.2\sqrt{d_{\text{model}}})$ with target $=10^{-3}$ and $d_{\text{model}}=1024$, yielding $\text{base}=1.5625\!\times\!10^{-4}$.

\begin{table}[h]
    \centering
    \caption{\textbf{Architecture and training hyperparameters for the four DiT runs reported in this paper.} The three 400-layer controls share the same backbone, optimizer, batch size, and non-residual primitives, and vary in residual stabilization and initialization protocol. The 1000-layer run uses the same backbone family and MV-Split residual design, but differs in depth, hardware scale, and post-training pipeline. In MV-Split, $\alpha$ and $\beta$ are unconstrained learnable vectors; empirically, $\alpha$ remains in $[0,1]$ throughout training in all reported MV-Split runs. For the 1000-layer run, $\beta_{\mathrm{init}}$ is set to $0.03\approx 1/\sqrt{L}$ following standard depth-variance scaling; the MMS protection itself stems from the anisotropic split ($\alpha_{\mathrm{init}}{=}0 < \beta_{\mathrm{init}}$) rather than from isotropic shrinkage.}
    \label{tab:training_config}
    \scriptsize
    \renewcommand{\arraystretch}{1.3}
    \begin{tabularx}{\linewidth}{@{}l C C C C@{}}
        \toprule
        \textbf{Field} & \textbf{DiT-400L-Baseline} & \textbf{DiT-400L-LayerScale} & \textbf{DiT-400L-MVSplit} & \textbf{DiT-1000L-MVSplit} \\
        \midrule
        Pretraining Dataset        & \multicolumn{4}{c}{ImageNet-2012} \\
        Image Autoencoder          & \multicolumn{4}{c}{Frozen FLUX.2 VAE} \\
        Text Encoder               & \multicolumn{4}{c}{Frozen Qwen3-0.6B} \\
        Trainable Components       & \multicolumn{4}{c}{DiT backbone only} \\
        Training Hardware          & 8$\times$H100 & 8$\times$H100 & 8$\times$H100 & 16$\times$H100 \\
        Post-training Dataset      & --- & --- & --- & $\sim$50k curated images \\
        \midrule
        DiT Params                 & 5.45\,B & 5.45\,B & 5.45\,B & 13.64\,B \\
        Layers                     & 400     & 400     & 400     & 1000 \\
        Residual Mode              & None & LayerScale & Mean-Variance Split & Mean-Variance Split \\
        Residual Gates             & --- & learnable $\lambda$, $\lambda_{\mathrm{init}}\!\in\!\{10^{-2},\dots,10^{-5}\}$ & learnable $\alpha,\beta$; $\alpha_{\mathrm{init}}{=}0,\ \beta_{\mathrm{init}}{=}1$ & learnable $\alpha,\beta$; $\alpha_{\mathrm{init}}{=}0,\ \beta_{\mathrm{init}}{=}0.03$ \\
        Learning Rate              & $1.5625\!\times\!10^{-4}$, $7.8125\!\times\!10^{-5}$ & $1.5625\!\times\!10^{-4}$ & $1.5625\!\times\!10^{-4}$ & $1.5625\!\times\!10^{-4}$ \\
        Initialization Method      & zero init $W_O, W_2$;\newline standard init others & standard init, $\mathcal{N}(0, 0.02^2)$ & zero init $W_O, W_2$;\newline standard init others & zero init $W_O, W_2$;\newline standard init others \\
        \midrule
        Dimension ($d_{\text{model}}$) & \multicolumn{4}{c}{1024} \\
        FFN Dimension              & \multicolumn{4}{c}{3072} \\
        FFN Type                   & \multicolumn{4}{c}{SwiGLU} \\
        Attention Heads            & \multicolumn{4}{c}{8} \\
        Attention Head Dim         & \multicolumn{4}{c}{128} \\
        KV Heads                   & \multicolumn{4}{c}{8} \\
        Attention Type             & \multicolumn{4}{c}{MHA} \\
        Position Embedding         & \multicolumn{4}{c}{2D RoPE} \\
        RoPE $\theta$              & \multicolumn{4}{c}{10000} \\
        Layer Norm Type            & \multicolumn{4}{c}{RMSNorm, non-affine} \\
        RMSNorm Affine Gain        & \multicolumn{4}{c}{disabled} \\
        RMSNorm $\epsilon$         & \multicolumn{4}{c}{$10^{-6}$} \\
        QK-Norm                    & \multicolumn{4}{c}{\checkmark, non-affine} \\
        \midrule
        LR Scheduler               & \multicolumn{4}{c}{warmup $\rightarrow$ constant} \\
        Warmup Steps               & \multicolumn{4}{c}{1000} \\
        Global Batch Size          & \multicolumn{4}{c}{1024} \\
        Optimizer                  & \multicolumn{4}{c}{AdamW~\cite{loshchilov2017decoupled}} \\
        AdamW Betas                & \multicolumn{4}{c}{$(0.9,\,0.999)$} \\
        AdamW $\epsilon$           & \multicolumn{4}{c}{$10^{-8}$} \\
        Weight Decay               & \multicolumn{4}{c}{0.1 (2D Weights Only)} \\
        Training Steps             & crashed & 100\,k & 100\,k & 100\,k \\
        Gradient Clipping          & \multicolumn{4}{c}{1.0} \\
        \bottomrule
    \end{tabularx}
\end{table}

\FloatBarrier

\clearpage
\section{The Token Mean as an Implicit Timestep Carrier}
\label{sec:appendix_timestep_probe}

Our backbone has no AdaLN and no explicit timestep embedding, so the model
must infer the continuous rectified-flow time $t$ from the noisy latent itself.
In this appendix, we use ``timestep'' to refer to this continuous interpolation
coordinate; equivalently, it is the noise-level coordinate controlling
$z_t=(1-t)x_0+t x_1$ from data latent to Gaussian noise.

We run a post-hoc linear probe on ImageNet-2012 validation images. Each image
is encoded into the same VAE latent space used during training. We sample
$x_1\sim\mathcal{N}(0,I)$ and $t\sim U[0,1]$, form
$z_t=(1-t)x_0+t x_1$, and record hidden states from the trained 400-layer
MV-Split checkpoint. For each probed layer, we fit ridge regressors to predict
$t$ from the image-token mean $m_l^{\mathrm{img}}$, a centered image-token RMS
summary $\mathrm{rms}(c_l^{\mathrm{img}})$, and the text-token mean
$m_l^{\mathrm{txt}}$. Train/test splits are grouped by image id; scalar
input-statistic, shuffled-label, and untrained-model controls are included.
For panel~(b), we report the fraction of residual squared error left by scalar
input statistics that is removed by adding a hidden-state feature,
$1-\mathrm{SSE}_{\mathrm{input}+h}/\mathrm{SSE}_{\mathrm{input}}$.

\begin{figure}[H]
    \centering
    \includegraphics[width=\linewidth]{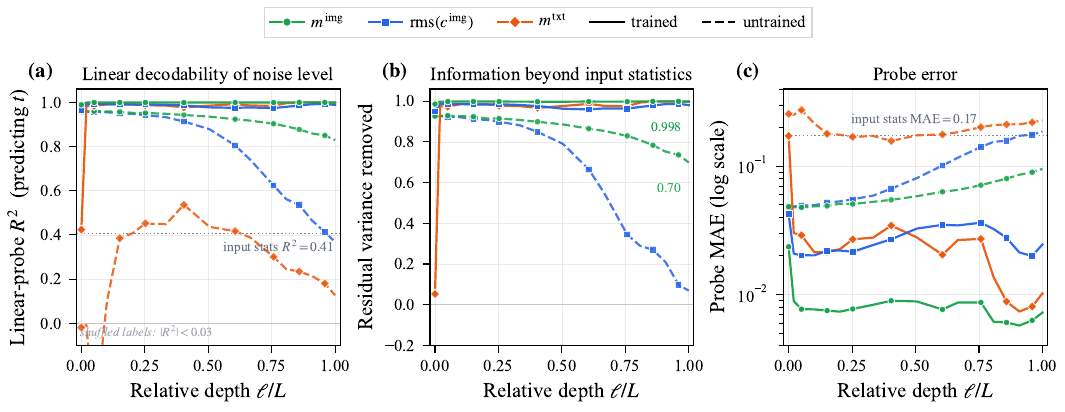}
    \caption{
    \textbf{Real-image timestep linear probe.}
    The backbone has no explicit timestep embedding or AdaLN modulation.
    \textbf{(a)} The trained image-token mean $m^{\mathrm{img}}$ predicts $t$ with
    near-perfect linear $R^2$ across depth. The text-token mean $m^{\mathrm{txt}}$
    becomes predictive after a few single-stream layers, indicating that the
    trained model routes image-derived timestep information into the text-token
    side.
    \textbf{(b)} Adding hidden-state summaries removes nearly all residual error
    left by scalar input statistics.
    \textbf{(c)} Probe MAE (mean absolute error) shows the same pattern on a log scale.
    The result shows that the token mean is not merely a collapse-prone direction:
    it is also a useful global timestep carrier. The same coordinate is also
    decodable from centered-energy summaries, so the claim is not uniqueness of
    the mean subspace but its usefulness and stability as a global-state path.
    }
    \label{fig:timestep_probe}
\end{figure}

Figure~\ref{fig:timestep_probe} shows that the trained image-token mean
predicts $t$ with near-perfect linear $R^2$ across depth. The trained text-token
mean becomes predictive after only a few single-stream layers, indicating that
image-derived timestep information is routed into the shared multimodal
sequence. A randomly initialized network already exposes substantial
decodability in early image-token states, showing that the timestep is
structurally available from the input latent; training preserves this signal
through depth and routes it to the text-token side.

The token mean is therefore not only a collapse-prone direction but also a
useful global-state carrier for the timestep. MV-Split preserves the trunk
mean while gain-limiting new mean-path residual writes, controlling the
dangerous mean-coherent writer channel without erasing this useful global
state.

\FloatBarrier
\clearpage
\section{System Implementation: Triton Fusion of RoPE, QK-Norm, SwiGLU, and MV-Split+RMSNorm}
\label{sec:infra}

At the 8--16 H100 scale used in this work, training the ultra-deep DiT requires
activation checkpointing~\cite{chen2016checkpointing} to fit in GPU memory.
Checkpointing reduces activation memory, but it also replays checkpointed
blocks during the backward pass. As a result, lightweight per-block operators
such as RoPE, QK-Norm~\cite{henry2020qknorm,dehghani2023vit22b},
SwiGLU~\cite{shazeer2020glu}, and MV-Split+RMSNorm are executed far more
frequently than in a forward-only view of the model, making their memory-bound
overhead non-negligible in ultra-deep training~\cite{dao2022flashattention}.

\paragraph{Fused operators.}
We implement these operators in Triton~\cite{tillet2019triton}.
For MV-Split+RMSNorm, an eager implementation materializes the pre-normalized
residual state and launches separate kernels for segment-wise correction,
residual merging, and RMS normalization. Given precomputed segment means, the
fused kernel applies the segment-wise MV-Split update and the subsequent
non-affine RMS normalization without materializing the pre-normalized
intermediate residual. Its backward uses pointwise recomputation together with
compact segment-wise sufficient statistics, rather than caching the full
pre-normalized state across the checkpoint boundary.

\paragraph{Two-pass backward recomputation.}
Let $F_l=f_l(X_l)$ denote the residual branch output and let $Z_l$ be the
pre-normalized MV-Split merge:
\begin{equation}
  Z_l
  =
  X_l
  + \beta\odot P_{\mathrm{seg}}F_l
  + \alpha\odot J_{\mathrm{seg}}(F_l-X_l),
  \qquad
  X_{l+1}=\mathrm{RMSNorm}(Z_l).
\end{equation}
The backward does not require caching $Z_l$ across the checkpoint boundary.
Pass~A recomputes $Z_l$ on chip, evaluates the RMSNorm adjoint, and accumulates
the segment-wise sufficient statistics needed for the input and gate gradients;
Pass~B then applies the closed-form gradients to $X_l$ and $F_l$ using the
aggregated statistics. The full derivation is provided in
Section~\ref{sec:mvsplit_backward} below.

\paragraph{In-situ profiling.}
We evaluate the fused backend inside the distributed training loop rather than
with isolated microbenchmarks. In our 8-GPU, 400-block profiling setup with
activation checkpointing applied to three out of every four blocks, 300 blocks
are replayed during backward.
Consequently, RoPE, QK-Norm, and SwiGLU are each executed 700 times per active
optimizer step, while MV-Split+RMSNorm is executed 1400 times.

Relative to a matched eager PyTorch baseline, the fused Triton backend reduces
the aggregated self-CUDA time of these operators from
1697.4\,ms to 614.0\,ms per active optimizer step ($2.76\times$). The
individual reductions are from 359.7\,ms to 105.6\,ms for RoPE ($3.41\times$),
255.2\,ms to 101.2\,ms for QK-Norm ($2.52\times$), 118.3\,ms to 27.9\,ms for
SwiGLU ($4.24\times$), and 964.2\,ms to 379.3\,ms for MV-Split+RMSNorm
($2.54\times$). The explicit DiT forward range decreases from 1889.8\,ms to
1553.9\,ms, and the in-loop optimizer-step wall-clock decreases by 22.0\%, from
5.87\,s to 4.58\,s, excluding dataloader wait. QKV projection and SDPA remain
within a few percent under the same instrumentation, localizing the speedup
to repeated normalization, activation, and residual-merge paths rather than to
the main attention kernels.

\subsection{Closed-form Backward of MV-Split+RMSNorm}
\label{sec:mvsplit_backward}
For token $i$ in segment $s(i)$, the pre-normalized merge can be written as
\begin{equation}
  Z_{l,i}
  =
  X_{l,i}
  + \beta\odot (F_{l,i}-\bar F_l^{(s)})
  + \alpha\odot (\bar F_l^{(s)}-\bar X_l^{(s)}).
\end{equation}

Let $G_i=\partial\mathcal{L}/\partial X_{l+1,i}$ be the incoming gradient after
RMSNorm, and let
\[
  r_i=\left(\tfrac{1}{D}\|Z_{l,i}\|_2^2+\epsilon\right)^{-1/2}
\]
be the inverse-RMS factor. The pre-normalization adjoint
$\Delta_i=\partial\mathcal{L}/\partial Z_{l,i}$ is
\begin{equation}
    \Delta_i
    =
    r_i G_i
    -
    Z_{l,i}
    \left(
      \dfrac{r_i^3}{D}\langle G_i,Z_{l,i}\rangle
    \right).
\end{equation}
Define the segment-wise mean adjoint
\[
  \bar\Delta^{(s)}=\frac{1}{|s|}\sum_{i\in s}\Delta_i .
\]
Then the merge gradients are
\begin{align}
    \frac{\partial \mathcal{L}}{\partial X_{l,i}}
    &= \Delta_i - \alpha\odot\bar\Delta^{(s(i))}, \\
    \frac{\partial \mathcal{L}}{\partial F_{l,i}}
    &= \beta\odot\Delta_i + (\alpha-\beta)\odot\bar\Delta^{(s(i))}.
\end{align}
The gate gradients are
\begin{align}
    \frac{\partial \mathcal{L}}{\partial \alpha}
    &=
    \sum_s\sum_{i\in s}
    \Delta_i\odot(\bar F_l^{(s)}-\bar X_l^{(s)}), \\
    \frac{\partial \mathcal{L}}{\partial \beta}
    &=
    \sum_s\sum_{i\in s}
    \Delta_i\odot(F_{l,i}-\bar F_l^{(s)}).
\end{align}
These expressions require only pointwise recomputation and segment-wise
sums, which is what the two-pass kernel above evaluates.

\clearpage
\section{Methods we try but failed to prevent MMS}
\label{sec:appendix_negative_results}

We tested several interventions that target related objects: token means,
attention mixing, attention-output gating, scalar gradient-norm control, and
optimizer-side update geometry. None of these controls removed the
mean-dominated failure in this backbone; some additionally degraded
optimization. Their common limitation is that they do not combine local
mean/centered branch-gradient control with the forward leaky mean replacement
used by MV-Split.

\paragraph{Hard centering and attention reparameterizations.}
Explicit centering, $X\leftarrow PX$, removes the token mean rather than
gain-limiting new mean writes. This degraded optimization in our runs and
also removes useful global information, including image-level context and
the implicit timestep signal discussed in
Appendix~\ref{sec:appendix_timestep_probe}. Attention-matrix modifications
such as $A-I$, $I-A$, or $(1-\lambda)I+\lambda A$ change the attention
branch but do not protect the FFN branch or the residual merge. Moreover,
row-stochastic interpolations still preserve pure-mean states, and in
multimodal sequences global centering does not remove segment-wise mean
modes (image and text groups may each become internally homogeneous while
their global average remains zero). None of these implement the local branch-gradient split
\[
    G \mapsto \alpha J G + \beta P G, \qquad \alpha \ll \beta,
\]
nor the forward leaky mean replacement $JX_l \mapsto (1-\alpha)JX_l + \alpha JF_l$ that defines MV-Split.

\paragraph{Gated attention.}
We also tested attention-output gates of the form
\[
    Y_i = g_i(X_i)\odot \mathrm{SDPA}(Q,K,V)_i ,
\]
where $g_i$ is computed token-locally and acts along the \textbf{head} or \textbf{feature} dimension. It is not a sequence-level \textbf{token-space} projector and does not form $JY$ or $PY$. Such gates~\cite{qiu2025gatedattention}
can reduce attention-output magnitude and have been reported to mitigate
attention-sink behavior~\cite{xiao2023streamingllm}, but they do not compute
$JY$ and $PY$ or apply different gains to them. In the mean-dominated regime,
tokens are already aligned, so $g_i \approx \bar g$ tends to be similar across
tokens; the gate then scales the mean and centered components together rather
than separating them. More generally, attention-only controls leave the FFN $W_2$ uncontrolled.
Our attention-only MV-Split trace shows that, once the attention branch is
protected, the spike can relocate to the remaining ungated FFN branch
(Appendix~\ref{sec:partial_intervention}).

\paragraph{The gradient-clipping paradox.}
All runs in the main comparison use global gradient clipping with threshold
$1.0$ (Appendix~\ref{sec:appendix_training_config}), yet this does not remove
MMS. The reason is that MMS is not only a large-norm event; it is a
directional collapse of the writer update into the token-mean subspace. For a
writer gradient decomposed as $G = G_\mu + G_c$, global norm clipping applies
a scalar multiplier $\operatorname{clip}_\tau(G)=sG$ with
$s=\min(1,\tau/\|G\|)$, so
\[
    \operatorname{clip}_\tau(G) = sG_\mu+sG_c,
    \qquad
    \frac{\|sG_\mu\|}{\|sG_c\|}
    =
    \frac{\|G_\mu\|}{\|G_c\|}.
\]
Clipping can reduce the step length, but it cannot rotate a mean-coherent
writer update back into the centered subspace. When $G_\mu$ dominates, the
same scalar shrinkage also suppresses the already-small centered update,
leaving the feature-learning path starved. A global-norm safety check is also
blind to subspace structure: it can remain quiet while the residual writer
direction has already become structurally unsafe. MV-Split addresses this
failure mode at the residual interface by applying different gains to $JG$
and $PG$, rather than by thresholding the scalar norm of $G$.

\paragraph{Muon optimizer.}
Muon~\cite{jordan2024muon,liu2025muonscalable} orthogonalizes the
momentum/update matrix using Newton--Schulz iterations. This can remove
singular-value scale from a matrix update, but it acts after token gradients
have already been summed into $G=\sum_{t}\delta_t y_t^\top = G_\mu + G_c$.
If the momentum is dominated by an isolated mean-coherent term
$G_\mu=T\bar\delta\bar y^\top=\sigma u v^\top$, the orthogonalized update
removes $\sigma$ but keeps the direction $uv^\top$. For homogeneous token
inputs, this direction still produces the same update for every token and
therefore remains mean-coherent. More generally, Muon reshapes singular
values in parameter space; it does not implement the token-space split
$G \mapsto \alpha J G+\beta P G$ or the forward leaky mean replacement
$JX_l\mapsto(1-\alpha)JX_l+\alpha JF_l$ used by MV-Split. In our runs, Muon
could reduce update magnitude but did not remove the mean-dominated
trajectory, consistent with MMS being a residual-subspace failure rather
than only an optimizer-magnitude artifact.

\clearpage
\section{Additional Results on the MV-Split Runs}
\label{sec:appendix_mvsplit_extra}

\subsection{Text-Conditioned Evaluation of the 1000-Layer Checkpoint}
\label{sec:appendix_t2i_eval}

We report GenEval and DPG-Bench measurements for the post-trained 1000-layer
MV-Split checkpoint as a calibration of text-conditioned generation ability.
These numbers are not intended as a controlled comparison to large public
text-to-image systems: our model is trained on substantially smaller and
differently sourced data (ImageNet-2012 pretraining followed by SFT and DPO~\cite{rafailov2023dpo,wallace2023diffusiondpo} on
$\sim$50k curated images), uses a shorter training schedule, and uses a
simpler post-training pipeline. The purpose is only to confirm that the
1000-layer scale-validation run remains usable as a text-conditioned
generator, not to claim state-of-the-art text-to-image performance.

\begin{table}[h]
\centering
\caption{\small\textbf{Text-conditioned evaluation of the 1000-layer MV-Split
checkpoint.} Reported for calibration only and not used for state-of-the-art
comparison.}
\label{tab:t2i_calibration}
\begin{tabular}{lc}
\toprule
Metric & Score \\
\midrule
GenEval overall (avg.\ over tasks) & 0.534 \\
GenEval correct images             & 52.44\% \\
GenEval correct prompts            & 67.63\% \\
DPG-Bench overall                  & 74.91 \\
\bottomrule
\end{tabular}
\end{table}

\begin{table}[h]
\centering
\caption{\small\textbf{GenEval task breakdown (1000-layer MV-Split checkpoint).}}
\label{tab:geneval_breakdown}
\begin{tabular}{lc}
\toprule
Task & Accuracy \\
\midrule
single\_object & 92.81\% \\
two\_object    & 63.64\% \\
counting       & 33.75\% \\
colors         & 72.61\% \\
position       & 25.75\% \\
color\_attr    & 31.75\% \\
\bottomrule
\end{tabular}
\end{table}

\FloatBarrier

\subsection{Full-Horizon Training Loss Curve for the MV-Split Runs}
\label{sec:appendix_mvsplit_loss}

\begin{figure}[h]
    \centering
    \includegraphics[width=0.85\linewidth]{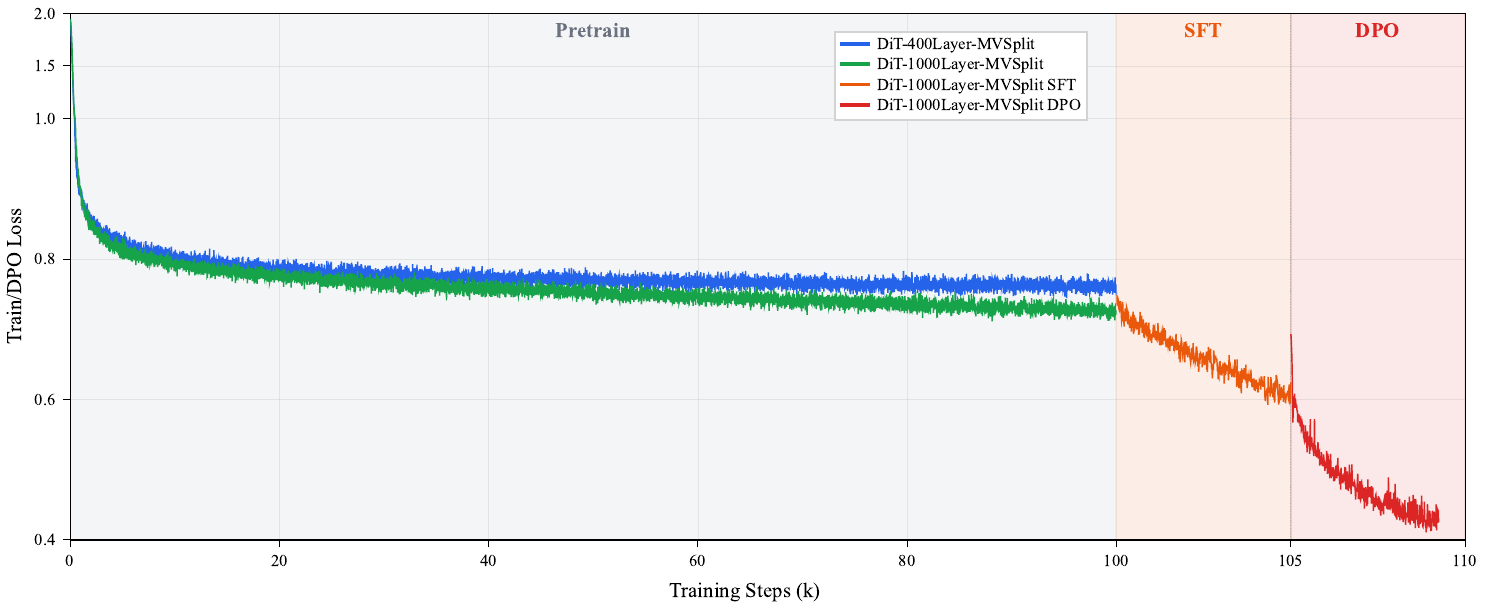}
    \caption{\textbf{Full-horizon training loss for the MV-Split 400-layer
    and 1000-layer runs.} Note that the SFT and DPO stages use a separately
    curated $\sim$50k image set rather than the ImageNet-2012 pre-training
    distribution; since loss values are data-dependent, the curves are shown
    for reference only.}
    \label{fig:mvsplit_loss_full}
\end{figure}

\FloatBarrier

\clearpage
\section{Limitations and Future Work}
\label{sec:appendix_limitations}

Our analysis identifies a residual-subspace failure pathway in ultra-deep
Diffusion Transformers and shows that MV-Split stabilizes this pathway in the
studied setting. The following boundary conditions define natural extensions
rather than contradictions of the mechanism.

\paragraph{Predicting the exact onset time of MMS.}
The alignment-amplification law in Eq.~\ref{eq:alignment_amp_exact}
characterizes when token-wise writer gradients stop canceling and enter a
coherent accumulation regime. This provides a mechanistic diagnostic for the
MMS transition, but it does not by itself predict the exact training step
$t^\star$ at which an un-stabilized run will cross the critical regime before
the run is observed. The onset time depends on the coupled evolution of token
representations, backward adjoints, optimizer momentum, data ordering, and
mini-batch statistics. We therefore view exact onset prediction as a separate
problem from architectural stabilization: MV-Split removes the unstable
residual interface by controlling the mean and centered writer-gradient
components directly, while deriving a closed-form scaling law for $t^\star$
remains an interesting direction for predictive theories of deep-network
training dynamics.

\paragraph{Architectures beyond Softmax attention.}
Several parts of our analysis use Transformer-specific structure. In
particular, row-stochastic attention preserves pure-mean token states
(Proposition~\ref{prop:mean-fixed}), and value homogenization suppresses Q/K
logit gradients through the null space of the Softmax Jacobian
(Lemma~\ref{lem:softmax-nullspace}). These arguments do not directly transfer
to attention-free sequence mixers such as convolutional diffusers or
state-space models such as Mamba~\cite{gu2023mamba}. At the same time, the
writer-gradient decomposition in Eq.~\ref{eq:writer-gradient-decomposition}
only assumes a token-wise residual writer and is not specific to Softmax
attention. This suggests a broader question: which components of the
mean-dominated collapse mechanism are consequences of attention, and which are
more general consequences of ultra-deep residual streams with token-wise
writers? Testing this distinction in convolutional, hybrid, and state-space
diffusion backbones is a natural next step.

\paragraph{Extreme-context spatiotemporal generation.}
Our scale validation focuses on image and text-to-image diffusion. Video,
3D, and other spatiotemporal generators often operate with substantially
longer token sequences and additional structure across time, views, or
modalities. In the coherent-alignment regime, the mean-coherent writer
component can scale with sequence length as in
Eq.~\ref{eq:writer-gradient-decomposition}, so these settings may place even
stronger pressure on the residual interface. MV-Split is designed to decouple
this mean-mode accumulation from the centered feature-learning path, but
validating and possibly adapting the mechanism for ultra-long-context
spatiotemporal DiTs remains an important direction for future large-scale
generative modeling.

\FloatBarrier

\clearpage
\section{More Visual Results}
\label{sec:appendix_more_samples}

We present additional uncurated samples from our 1000-layer MV-Split DiT to demonstrate the breadth and fidelity of the model across diverse semantic categories. All images are generated at $256 \times 256$ resolution using a Euler sampler~\cite{karras2022elucidating} with 35 NFE steps and classifier-free guidance~\cite{ho2022classifier} scale $w = 2.0$.

\paragraph{Text-Conditioned Generation.}
Unlike class-conditional DiTs that condition on a one-hot class label, our model is a \textbf{text-to-image} generator. Each sample is conditioned on a natural-language caption drawn from the ImageNet-2012 validation set, where captions were generated by a modern large language model and describe the scene content in 10--25 words (e.g., \textit{``A colorful green jacamar on a branch with an insect in its beak, set against a blurred natural background''} or \textit{``Snow-covered mountains under a dramatic cloudy sky, with sunlit huts and long shadows across the landscape''}). The captions vary in viewpoint, lighting, composition, and context, providing a diverse conditioning signal that goes beyond categorical labels. Within each grid below, the 12 images correspond to 12 \emph{distinct} captions from the same ImageNet class, showcasing the model's ability to faithfully render varied scene descriptions. In each grid, the top row displays 4 images at $2\times$ magnification for detail inspection, and the bottom row shows 8 additional samples at $1\times$ scale.


\begin{figure}[H]
    \centering
    \includegraphics[width=\linewidth]{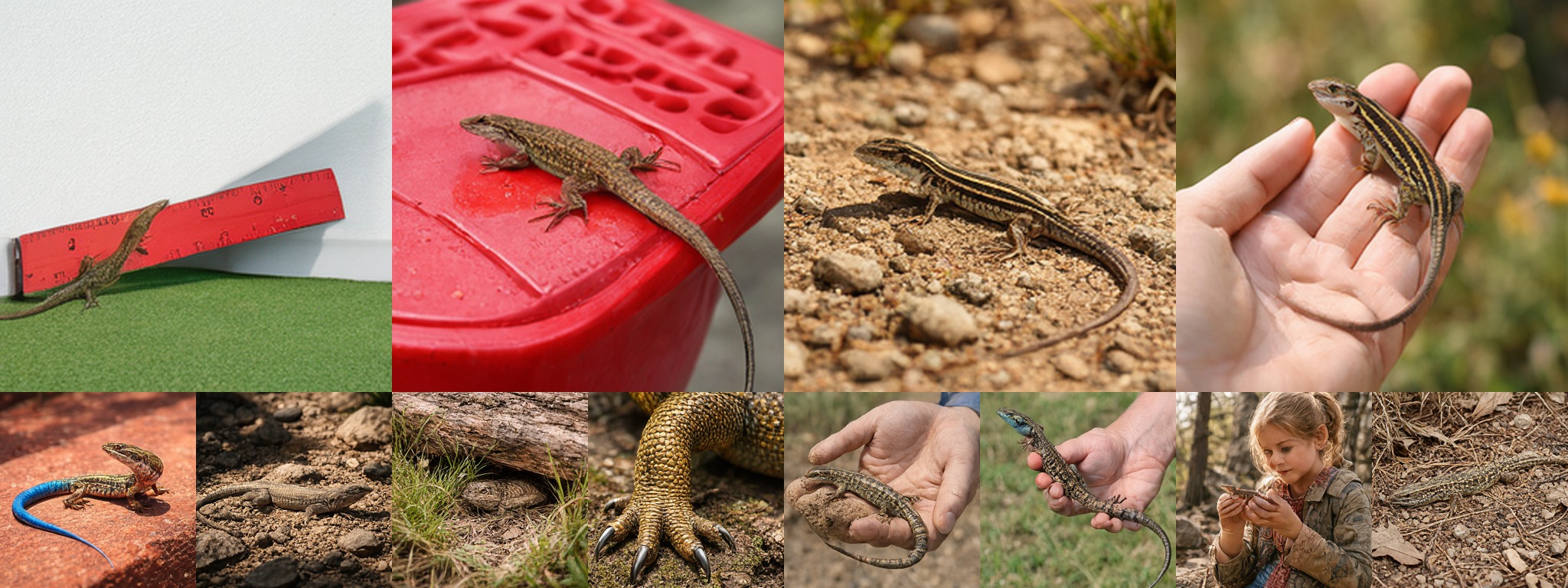}
    \caption{Class ``Alligator lizard'' (044). Euler sampler, 35 NFE, CFG $w = 2.0$.}
    \label{fig:app_grid_044}
\end{figure}

\begin{figure}[H]
    \centering
    \includegraphics[width=\linewidth]{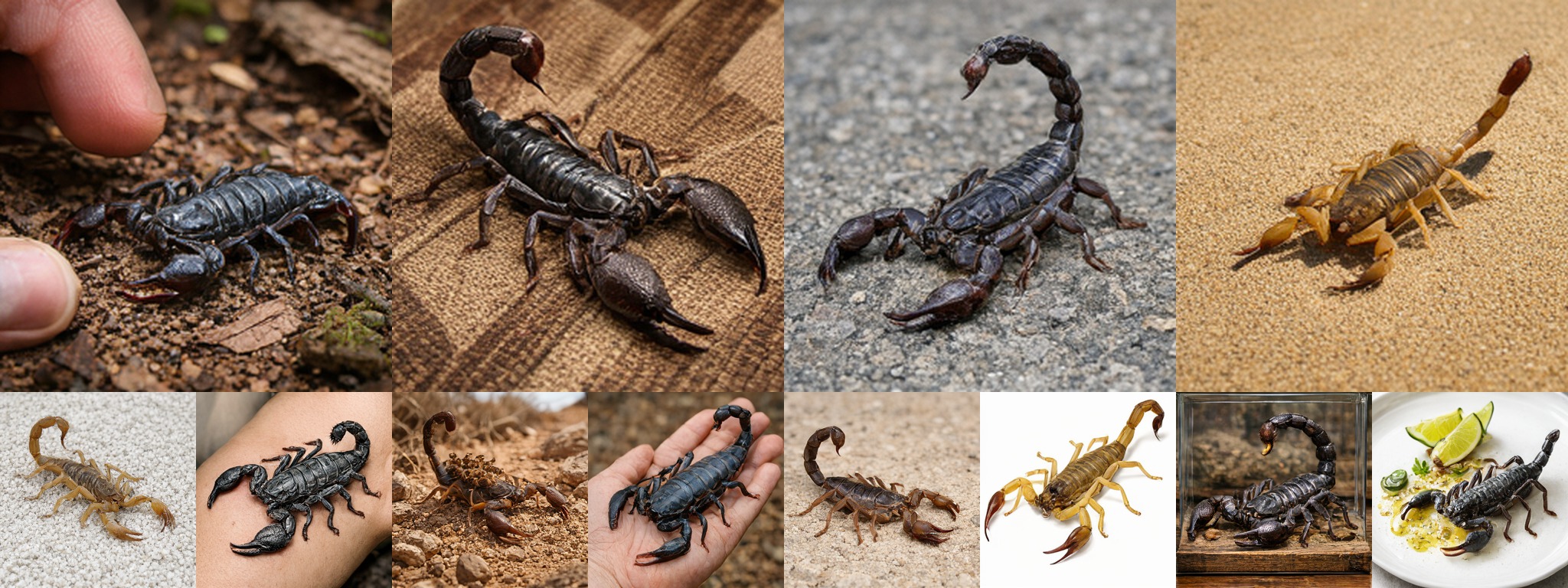}
    \caption{Class ``Scorpion'' (071). Euler sampler, 35 NFE, CFG $w = 2.0$.}
    \label{fig:app_grid_071}
\end{figure}

\begin{figure}[H]
    \centering
    \includegraphics[width=\linewidth]{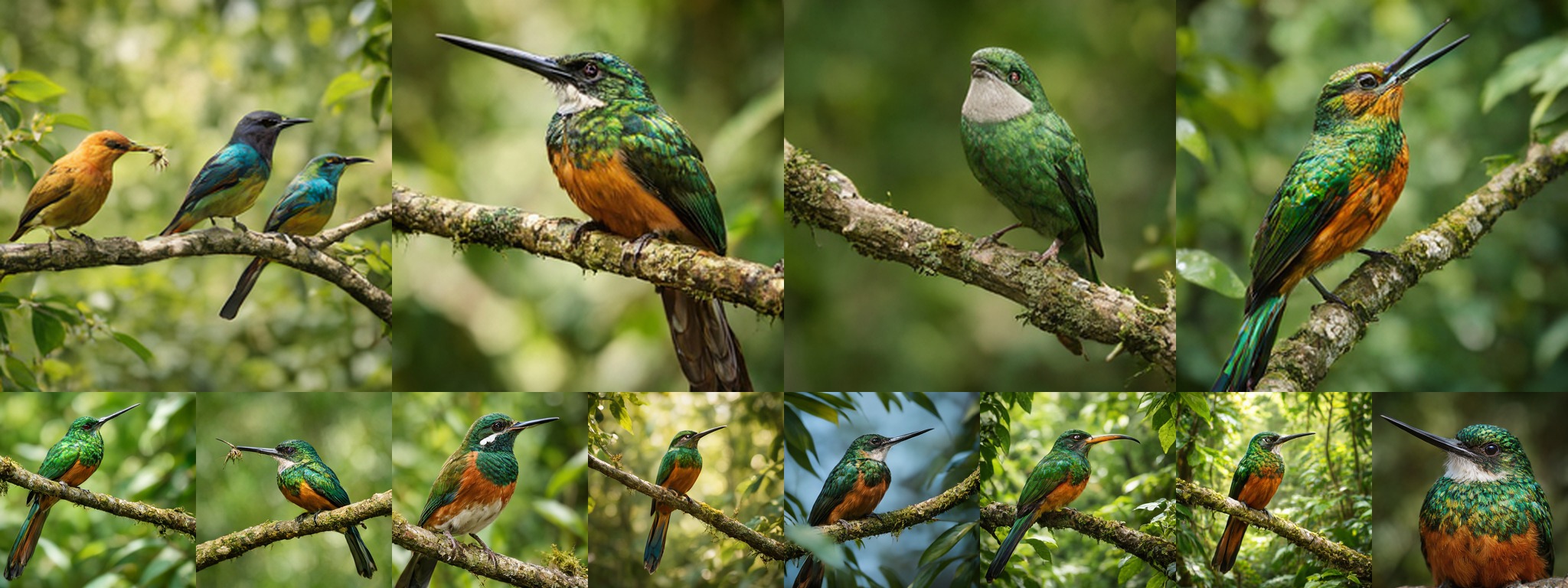}
    \caption{Class ``Jacamar'' (095). Euler sampler, 35 NFE, CFG $w = 2.0$.}
    \label{fig:app_grid_095}
\end{figure}

\begin{figure}[H]
    \centering
    \includegraphics[width=\linewidth]{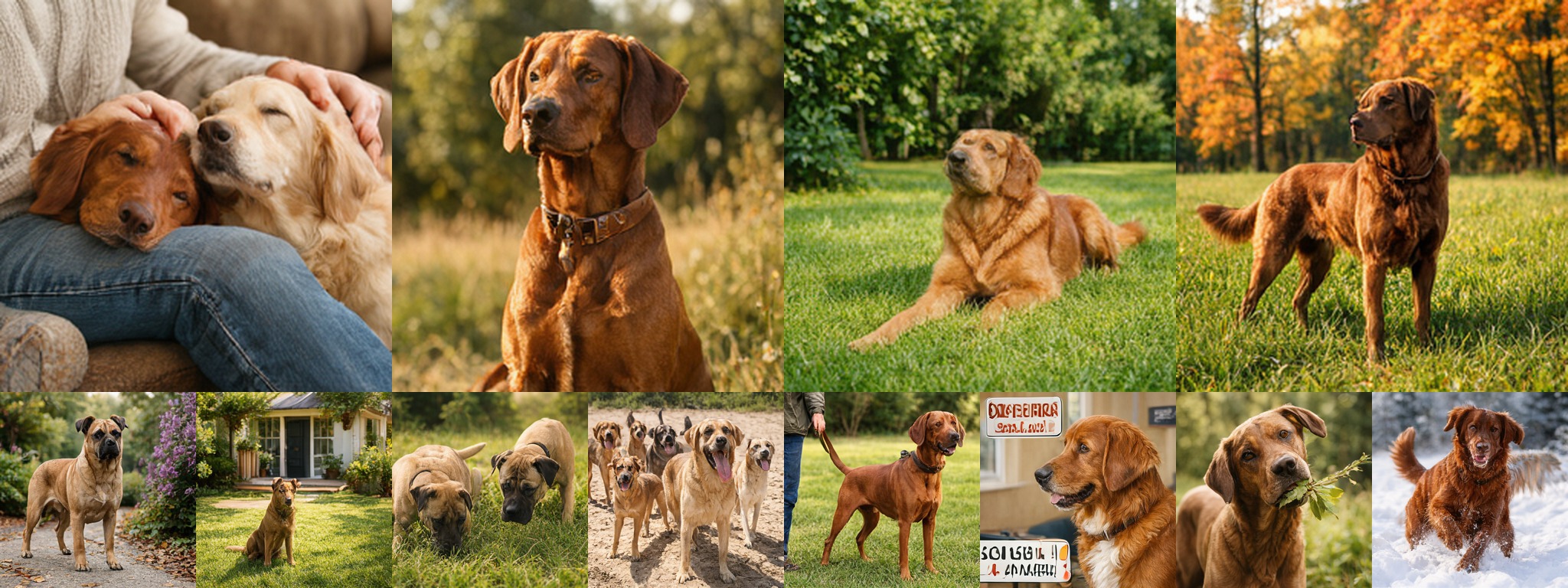}
    \caption{Class ``Rhodesian ridgeback'' (159). Euler sampler, 35 NFE, CFG $w = 2.0$.}
    \label{fig:app_grid_159}
\end{figure}

\begin{figure}[H]
    \centering
    \includegraphics[width=\linewidth]{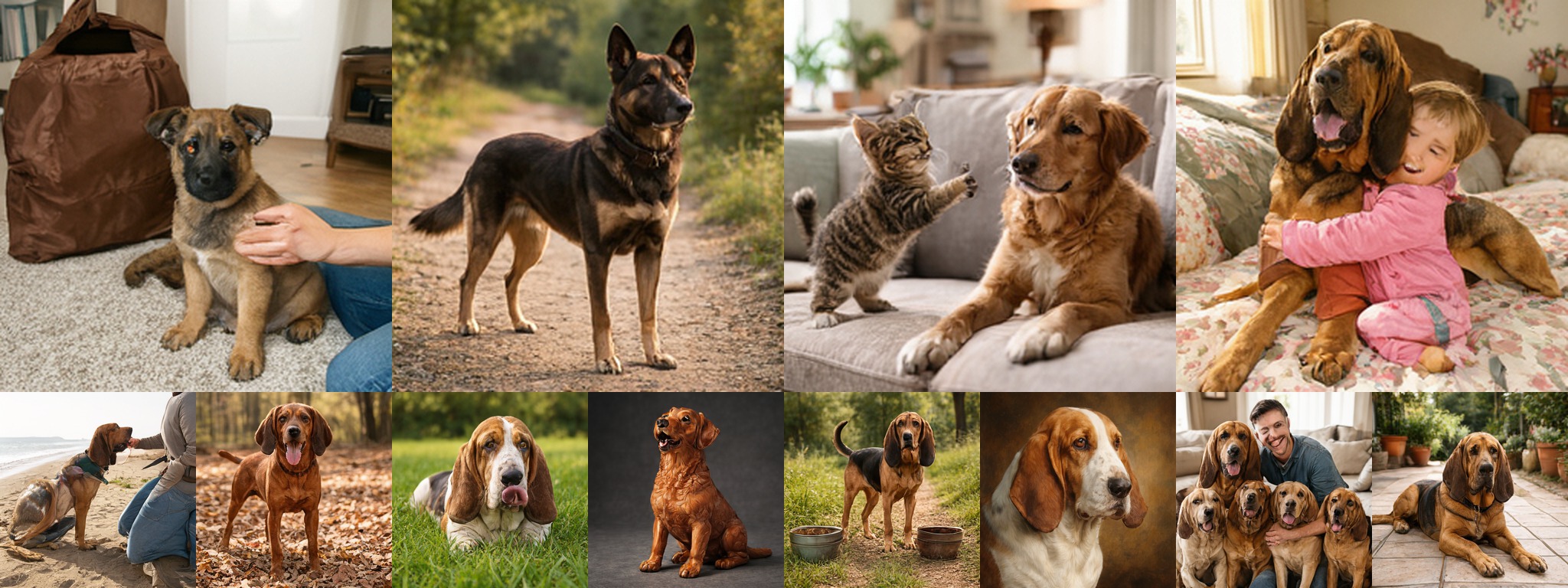}
    \caption{Class ``Bloodhound'' (163). Euler sampler, 35 NFE, CFG $w = 2.0$.}
    \label{fig:app_grid_163}
\end{figure}

\begin{figure}[H]
    \centering
    \includegraphics[width=\linewidth]{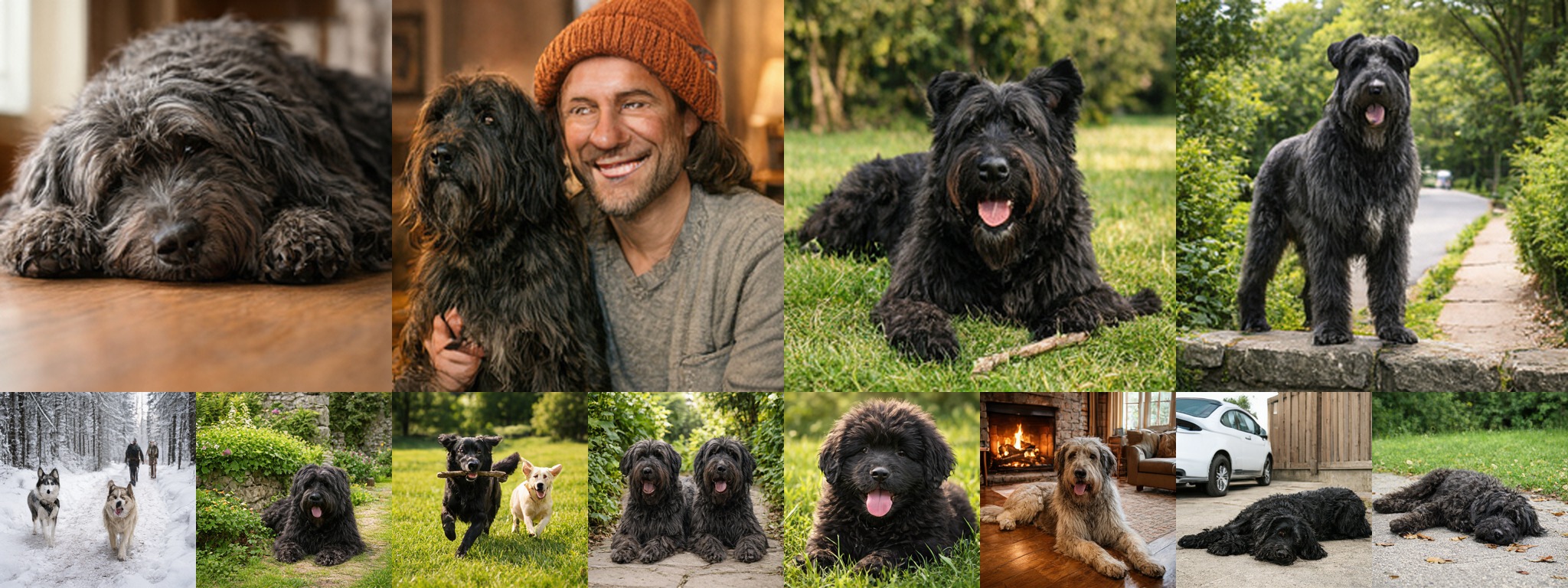}
    \caption{Class ``Bouvier des Flandres'' (233). Euler sampler, 35 NFE, CFG $w = 2.0$.}
    \label{fig:app_grid_233}
\end{figure}

\begin{figure}[H]
    \centering
    \includegraphics[width=\linewidth]{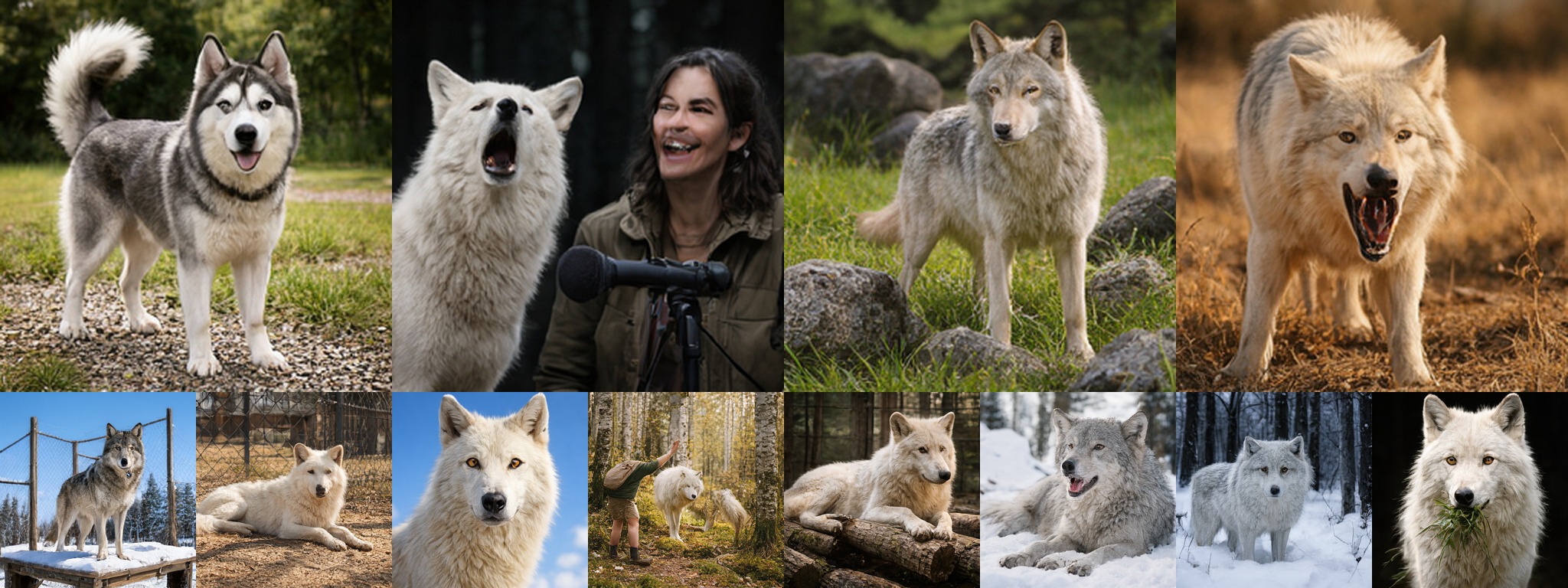}
    \caption{Class ``White wolf'' (270). Euler sampler, 35 NFE, CFG $w = 2.0$.}
    \label{fig:app_grid_270}
\end{figure}

\begin{figure}[H]
    \centering
    \includegraphics[width=\linewidth]{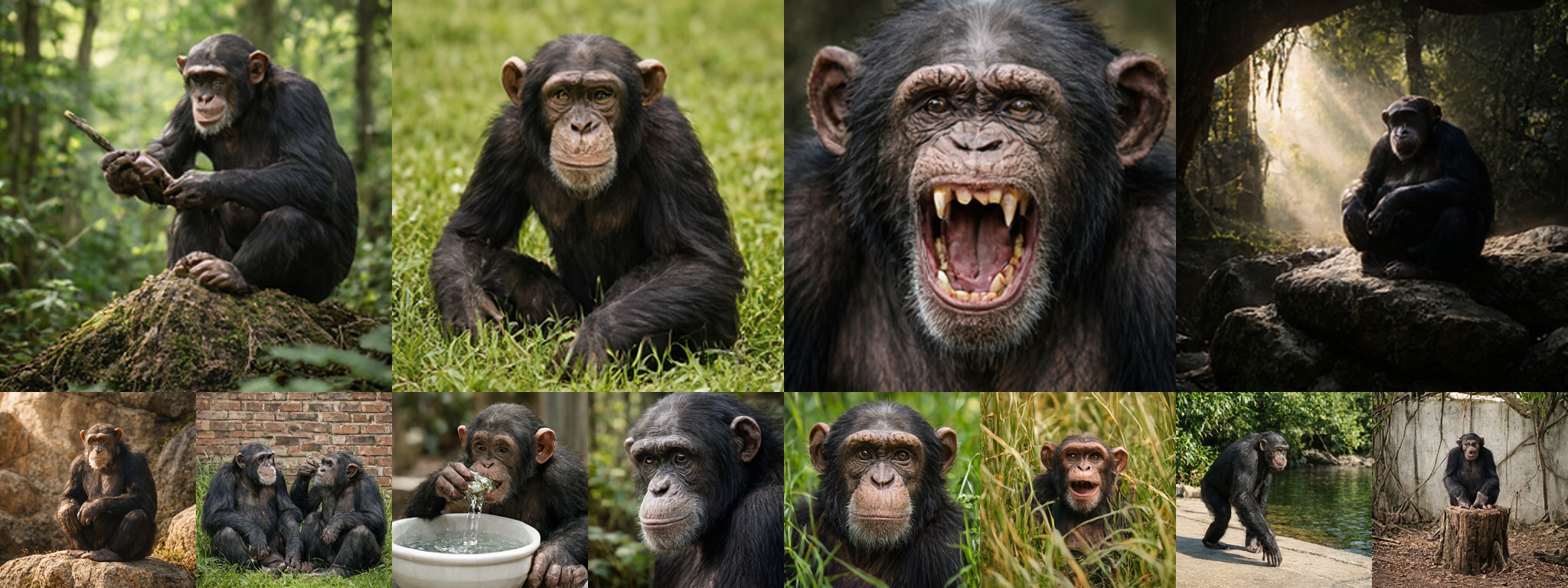}
    \caption{Class ``Chimpanzee'' (367). Euler sampler, 35 NFE, CFG $w = 2.0$.}
    \label{fig:app_grid_367}
\end{figure}

\begin{figure}[H]
    \centering
    \includegraphics[width=\linewidth]{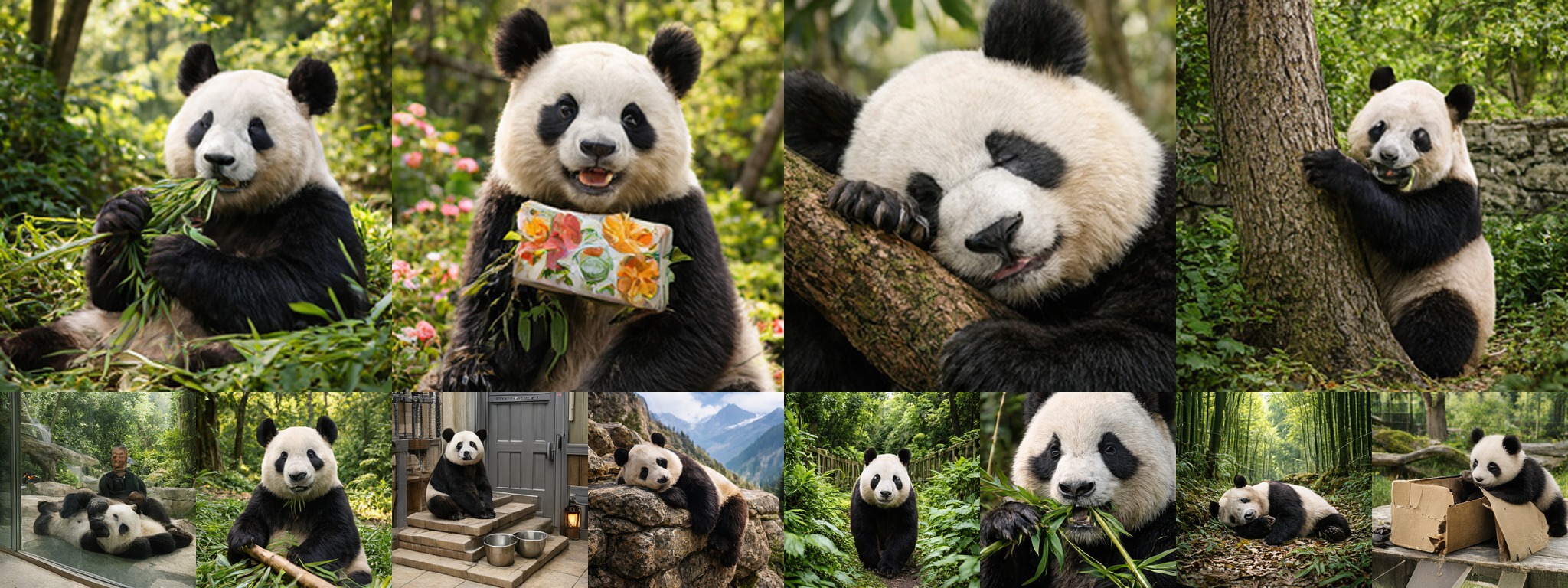}
    \caption{Class ``Giant panda'' (388). Euler sampler, 35 NFE, CFG $w = 2.0$.}
    \label{fig:app_grid_388}
\end{figure}


\begin{figure}[H]
    \centering
    \includegraphics[width=\linewidth]{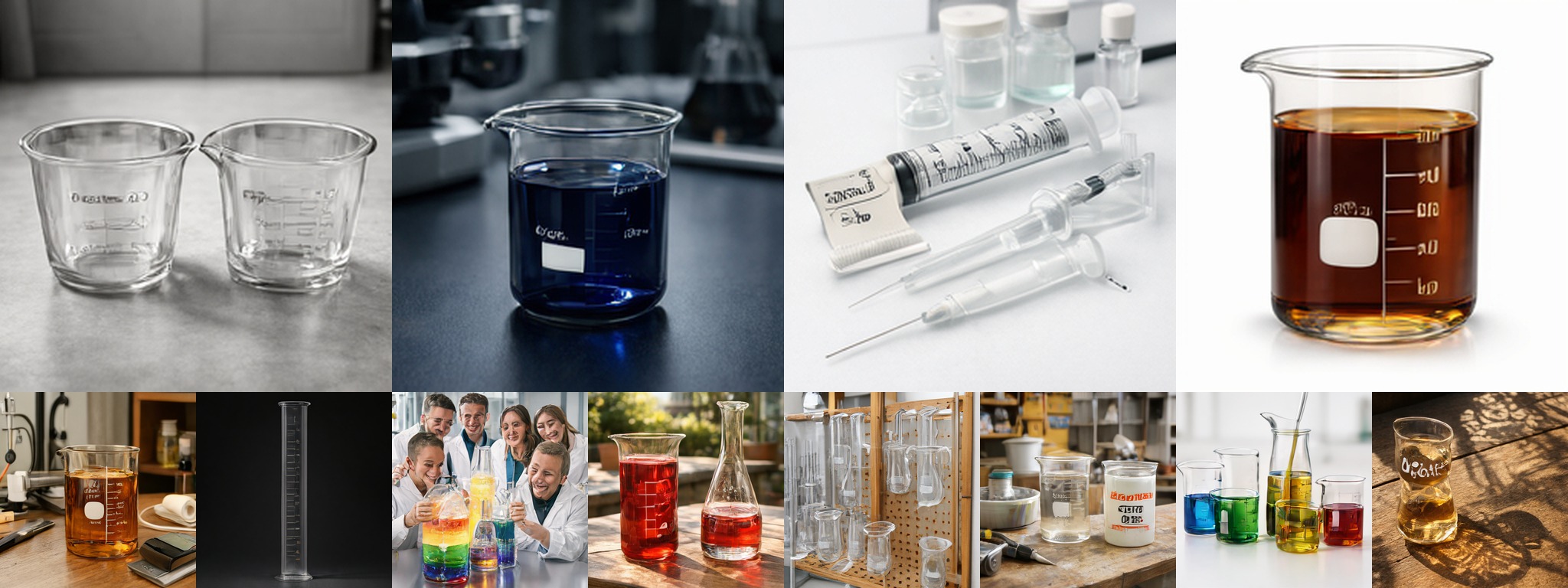}
    \caption{Class ``Beaker'' (438). Euler sampler, 35 NFE, CFG $w = 2.0$.}
    \label{fig:app_grid_438}
\end{figure}

\begin{figure}[H]
    \centering
    \includegraphics[width=\linewidth]{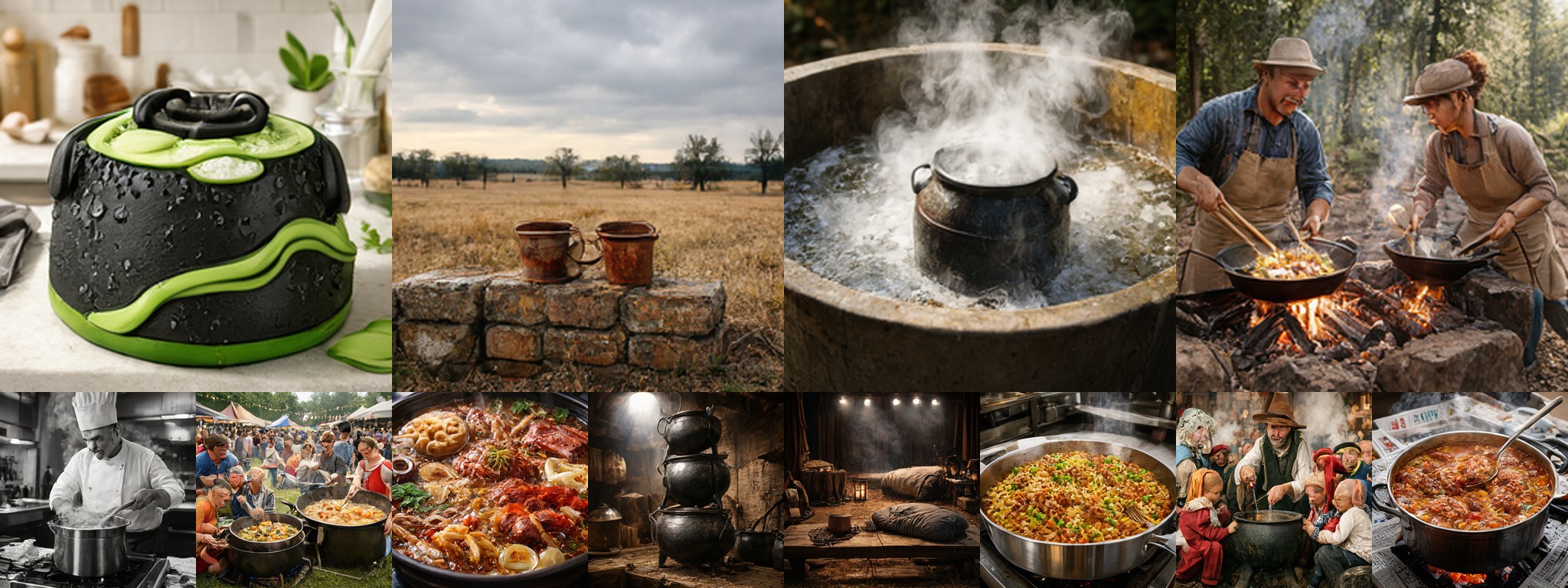}
    \caption{Class ``Caldron'' (469). Euler sampler, 35 NFE, CFG $w = 2.0$.}
    \label{fig:app_grid_469}
\end{figure}

\begin{figure}[H]
    \centering
    \includegraphics[width=\linewidth]{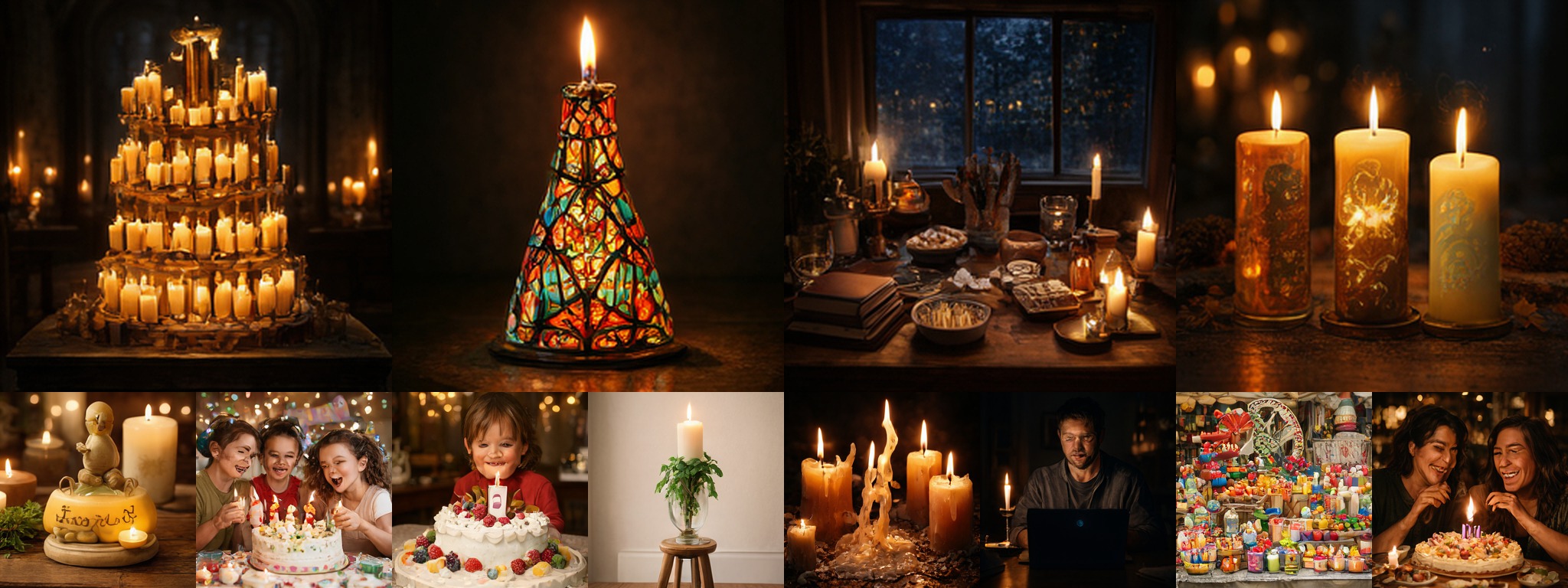}
    \caption{Class ``Candle'' (470). Euler sampler, 35 NFE, CFG $w = 2.0$.}
    \label{fig:app_grid_470}
\end{figure}

\begin{figure}[H]
    \centering
    \includegraphics[width=\linewidth]{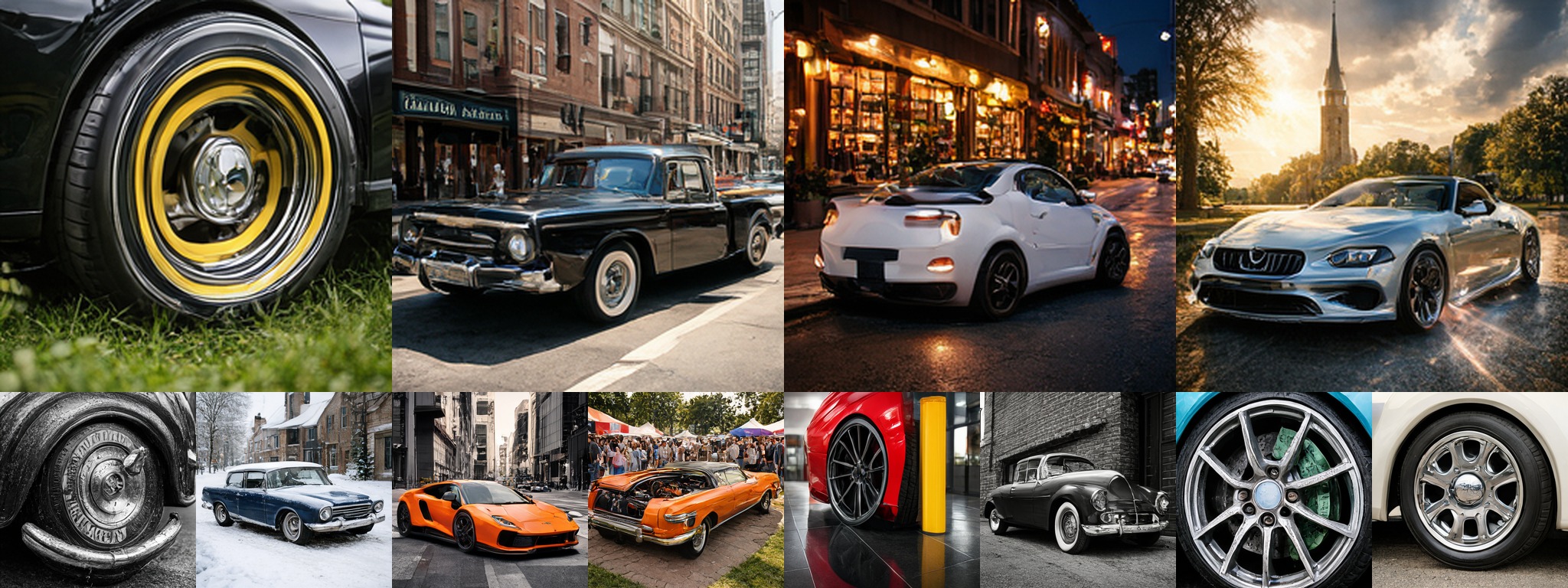}
    \caption{Class ``Car wheel'' (479). Euler sampler, 35 NFE, CFG $w = 2.0$.}
    \label{fig:app_grid_479}
\end{figure}

\begin{figure}[H]
    \centering
    \includegraphics[width=\linewidth]{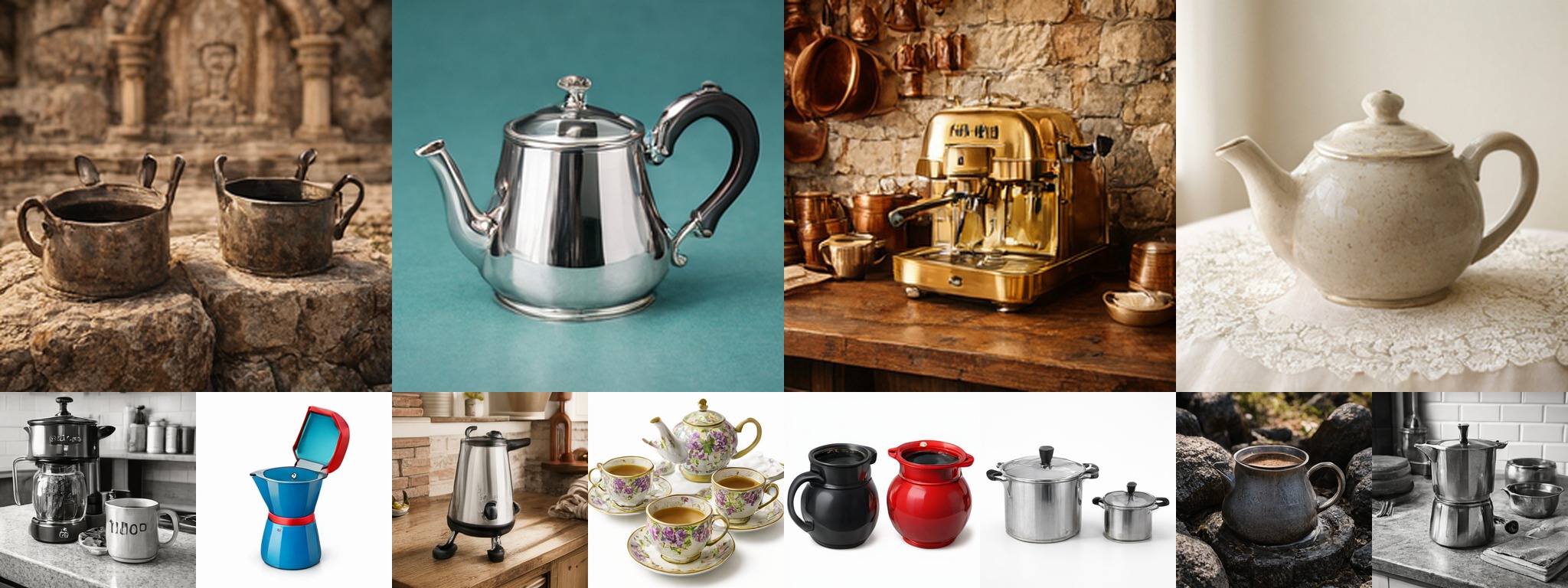}
    \caption{Class ``Coffeepot'' (505). Euler sampler, 35 NFE, CFG $w = 2.0$.}
    \label{fig:app_grid_505}
\end{figure}

\begin{figure}[H]
    \centering
    \includegraphics[width=\linewidth]{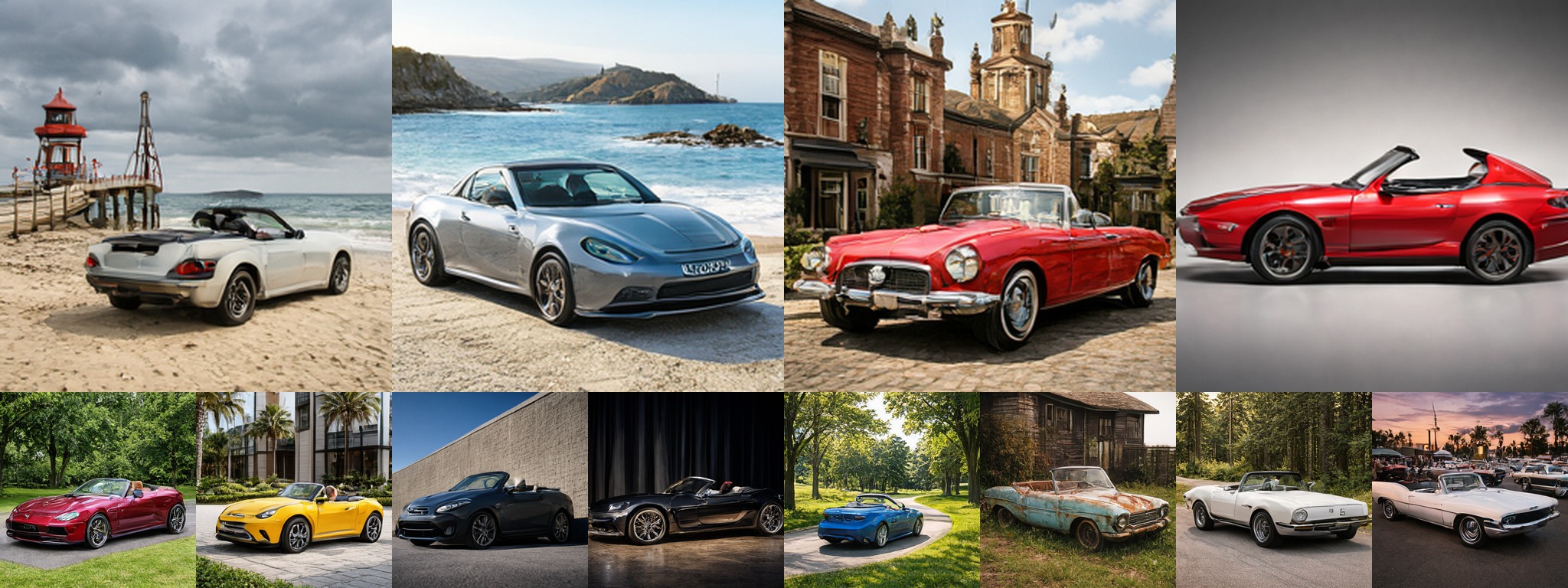}
    \caption{Class ``Convertible'' (511). Euler sampler, 35 NFE, CFG $w = 2.0$.}
    \label{fig:app_grid_511}
\end{figure}

\begin{figure}[H]
    \centering
    \includegraphics[width=\linewidth]{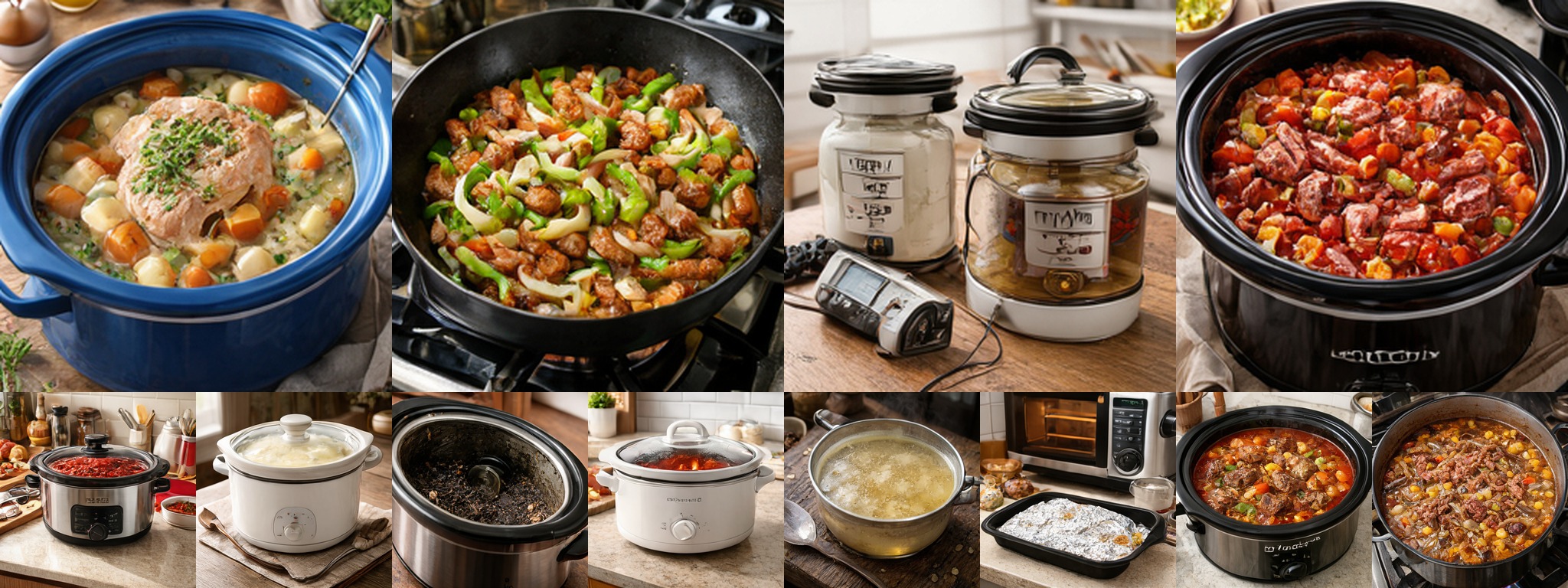}
    \caption{Class ``Crock Pot'' (521). Euler sampler, 35 NFE, CFG $w = 2.0$.}
    \label{fig:app_grid_521}
\end{figure}

\begin{figure}[H]
    \centering
    \includegraphics[width=\linewidth]{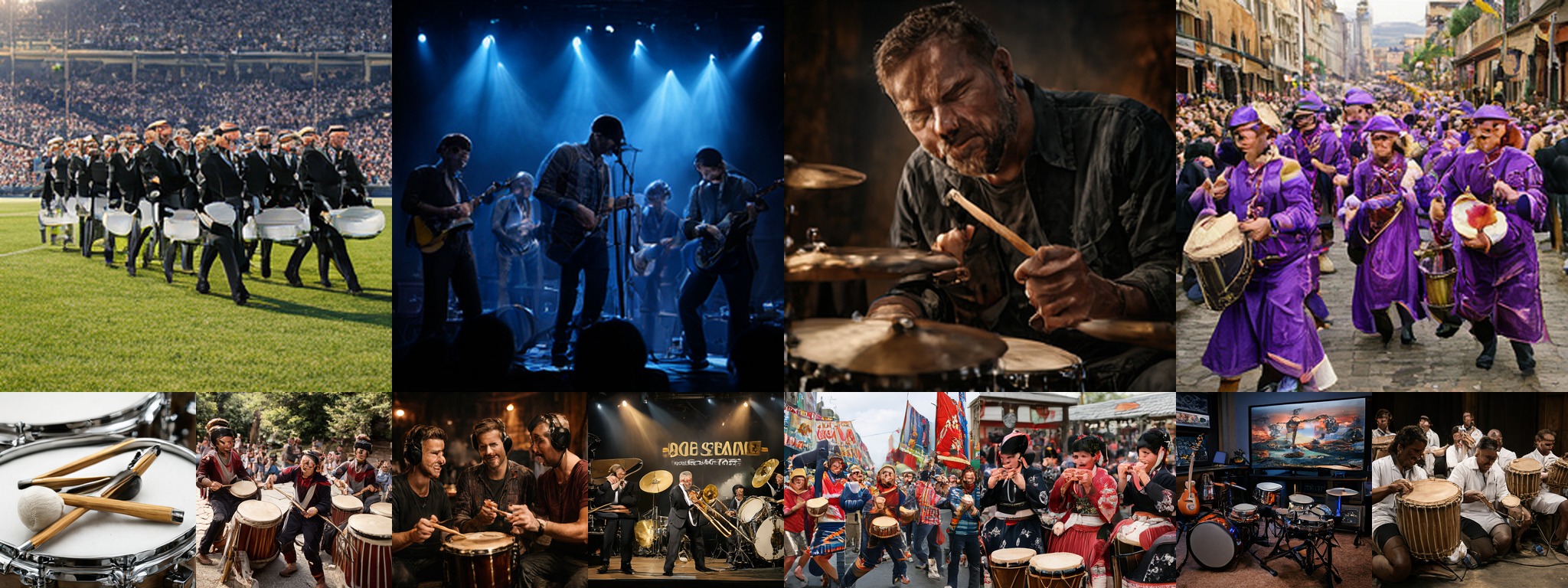}
    \caption{Class ``Drum'' (541). Euler sampler, 35 NFE, CFG $w = 2.0$.}
    \label{fig:app_grid_541}
\end{figure}

\begin{figure}[H]
    \centering
    \includegraphics[width=\linewidth]{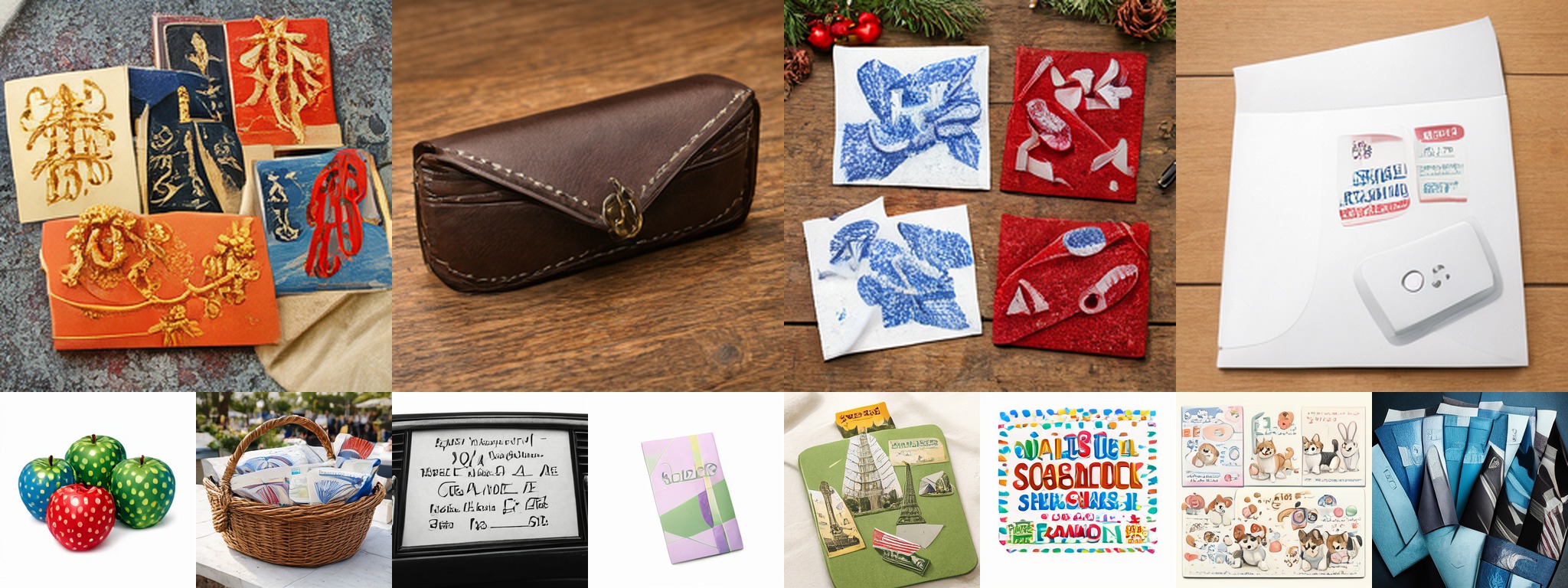}
    \caption{Class ``Envelope'' (549). Euler sampler, 35 NFE, CFG $w = 2.0$.}
    \label{fig:app_grid_549}
\end{figure}

\begin{figure}[H]
    \centering
    \includegraphics[width=\linewidth]{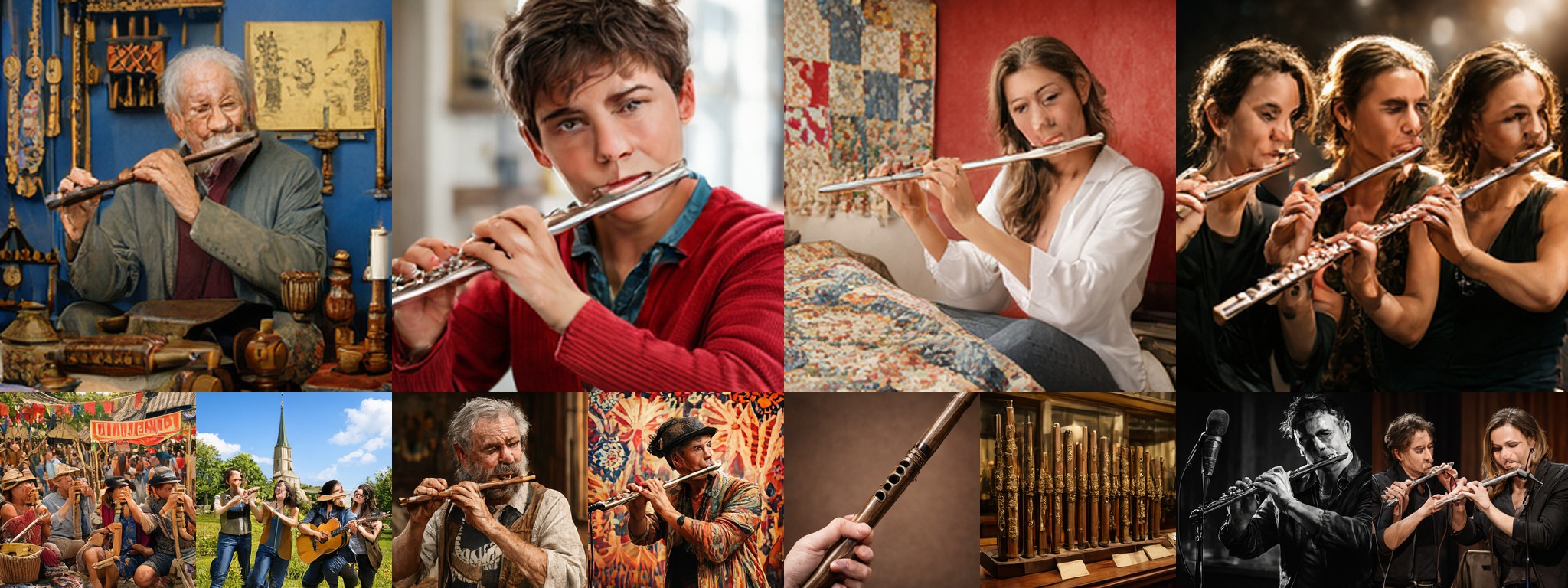}
    \caption{Class ``Flute'' (558). Euler sampler, 35 NFE, CFG $w = 2.0$.}
    \label{fig:app_grid_558}
\end{figure}

\begin{figure}[H]
    \centering
    \includegraphics[width=\linewidth]{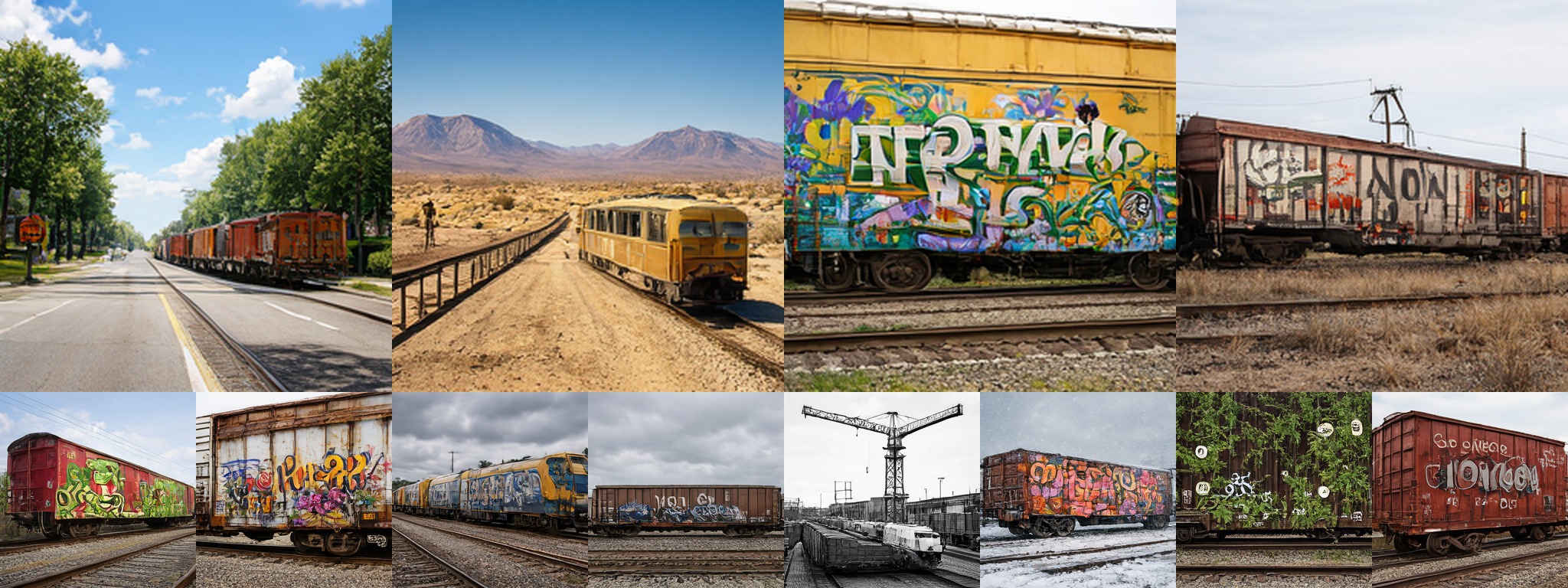}
    \caption{Class ``Freight car'' (565). Euler sampler, 35 NFE, CFG $w = 2.0$.}
    \label{fig:app_grid_565}
\end{figure}

\begin{figure}[H]
    \centering
    \includegraphics[width=\linewidth]{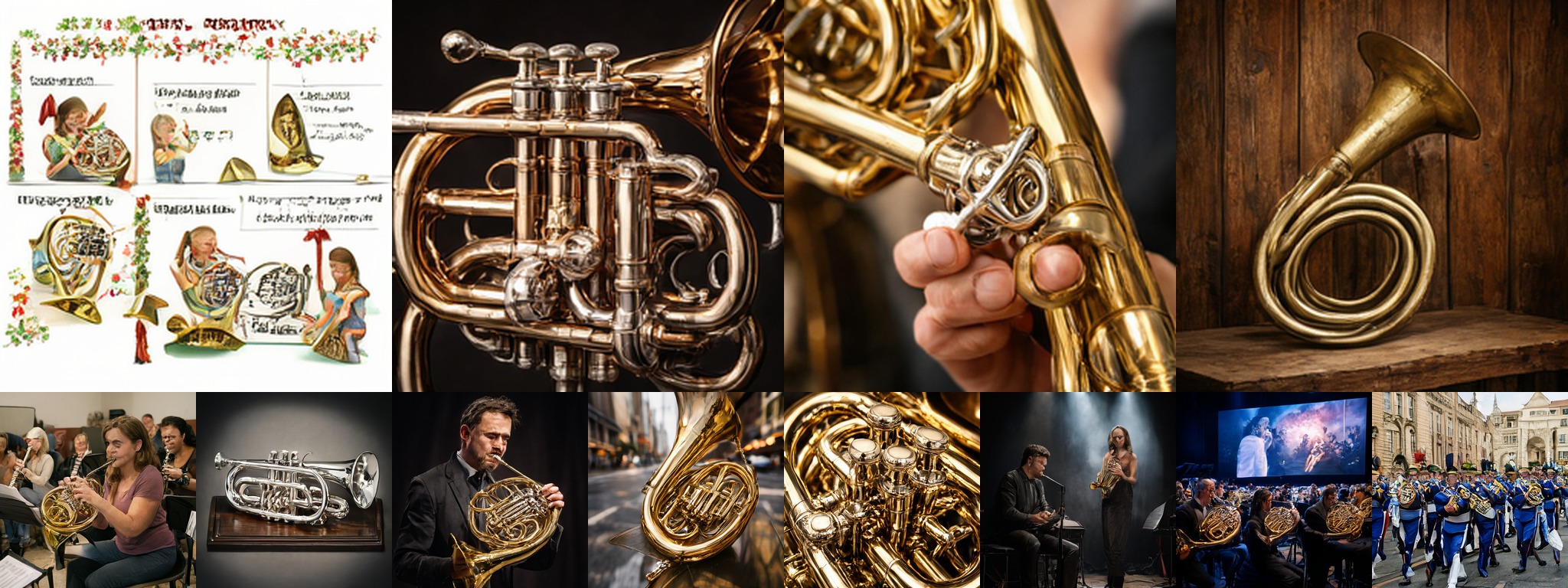}
    \caption{Class ``French horn'' (566). Euler sampler, 35 NFE, CFG $w = 2.0$.}
    \label{fig:app_grid_566}
\end{figure}

\begin{figure}[H]
    \centering
    \includegraphics[width=\linewidth]{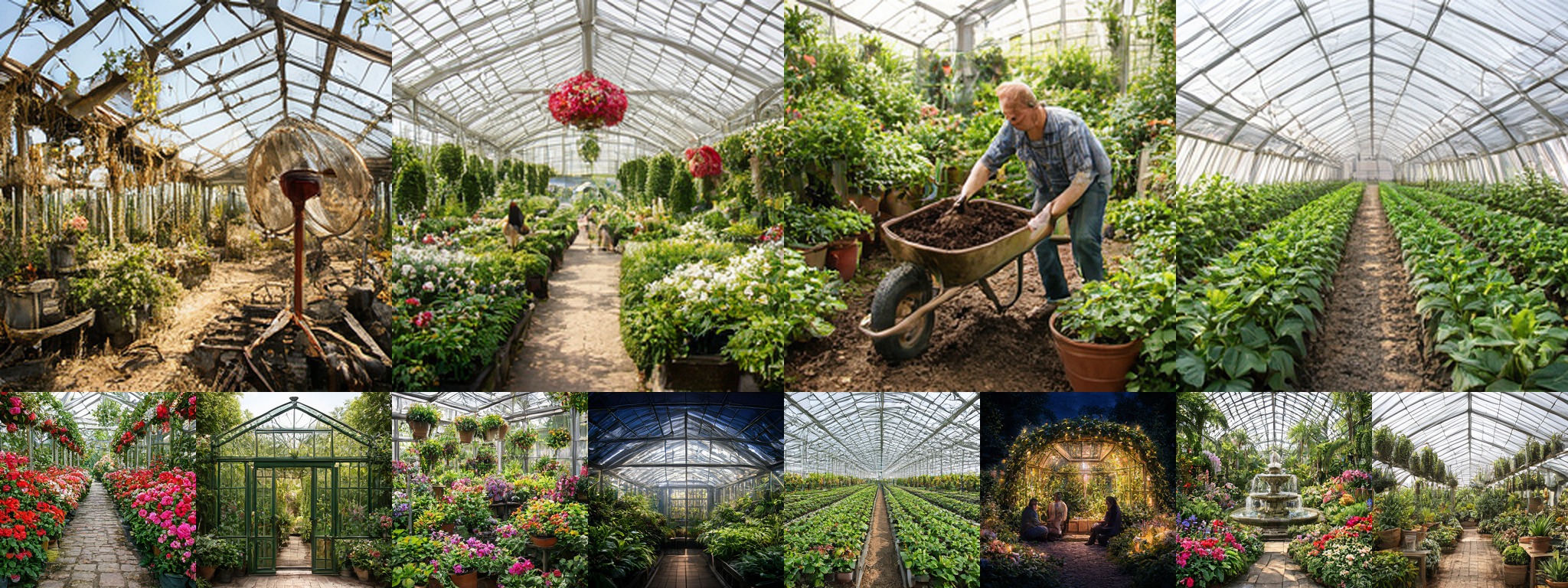}
    \caption{Class ``Greenhouse'' (580). Euler sampler, 35 NFE, CFG $w = 2.0$.}
    \label{fig:app_grid_580}
\end{figure}

\begin{figure}[H]
    \centering
    \includegraphics[width=\linewidth]{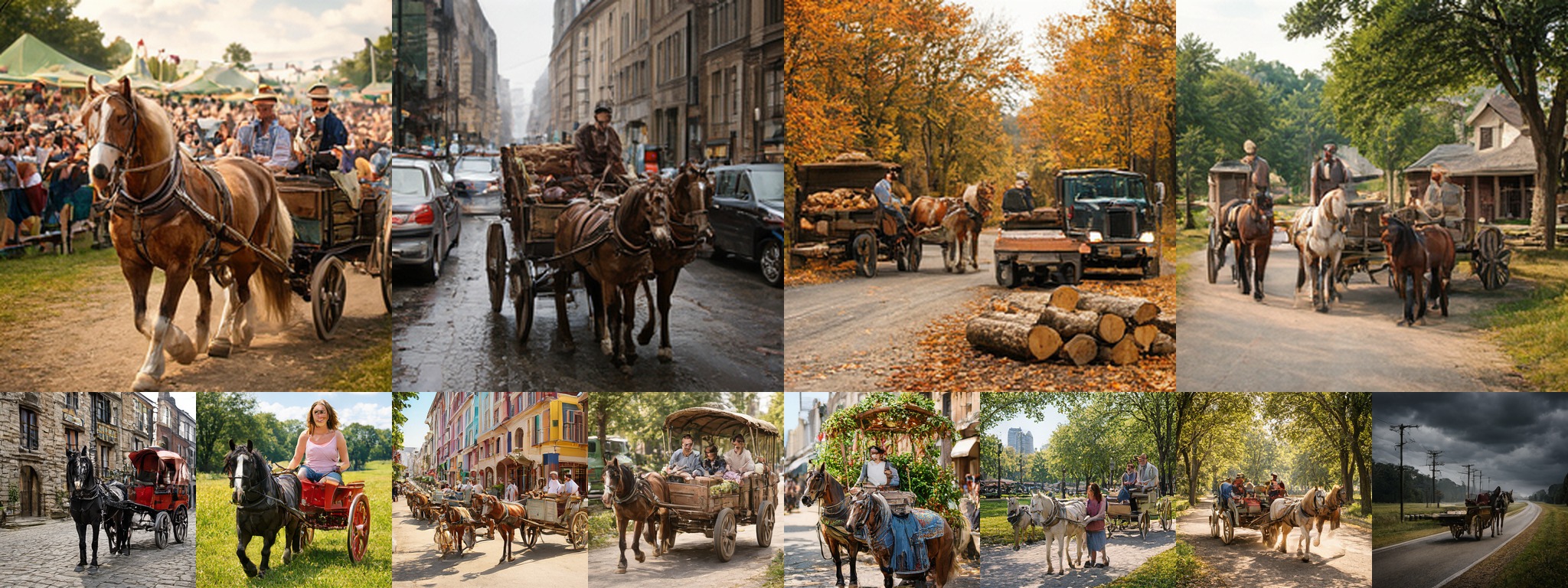}
    \caption{Class ``Horse cart'' (603). Euler sampler, 35 NFE, CFG $w = 2.0$.}
    \label{fig:app_grid_603}
\end{figure}

\begin{figure}[H]
    \centering
    \includegraphics[width=\linewidth]{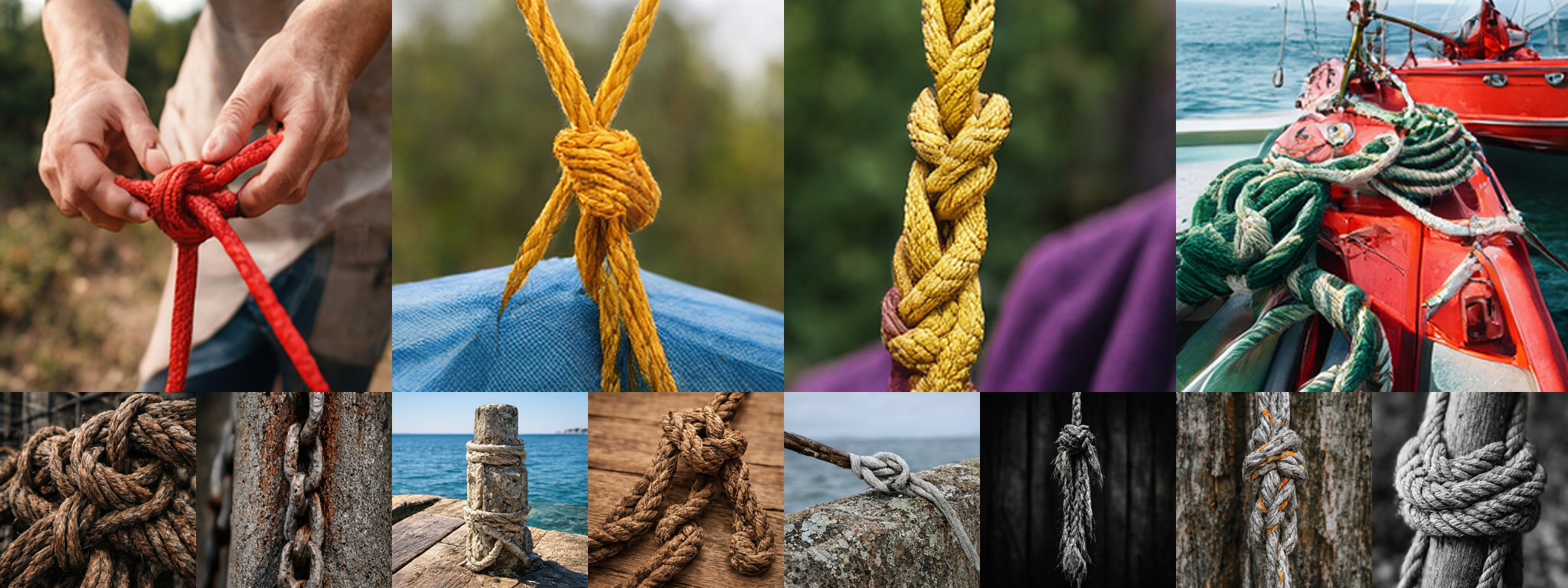}
    \caption{Class ``Knot'' (616). Euler sampler, 35 NFE, CFG $w = 2.0$.}
    \label{fig:app_grid_616}
\end{figure}

\begin{figure}[H]
    \centering
    \includegraphics[width=\linewidth]{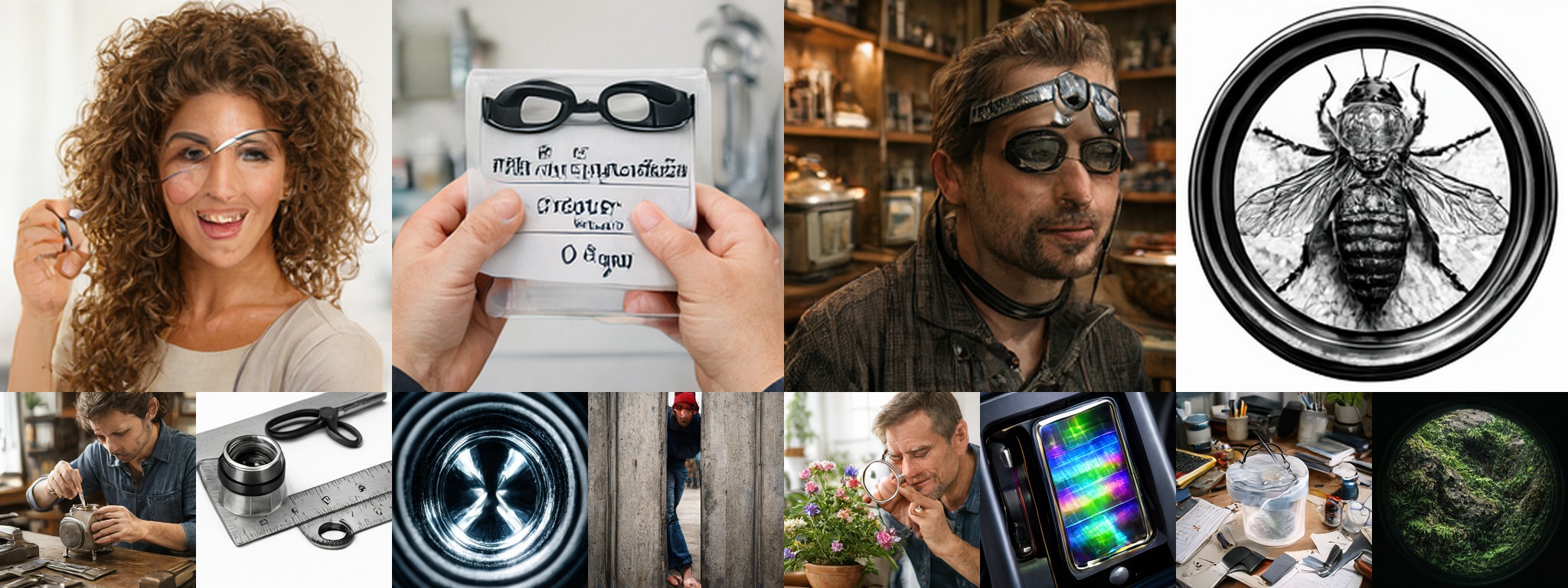}
    \caption{Class ``Loupe'' (633). Euler sampler, 35 NFE, CFG $w = 2.0$.}
    \label{fig:app_grid_633}
\end{figure}

\begin{figure}[H]
    \centering
    \includegraphics[width=\linewidth]{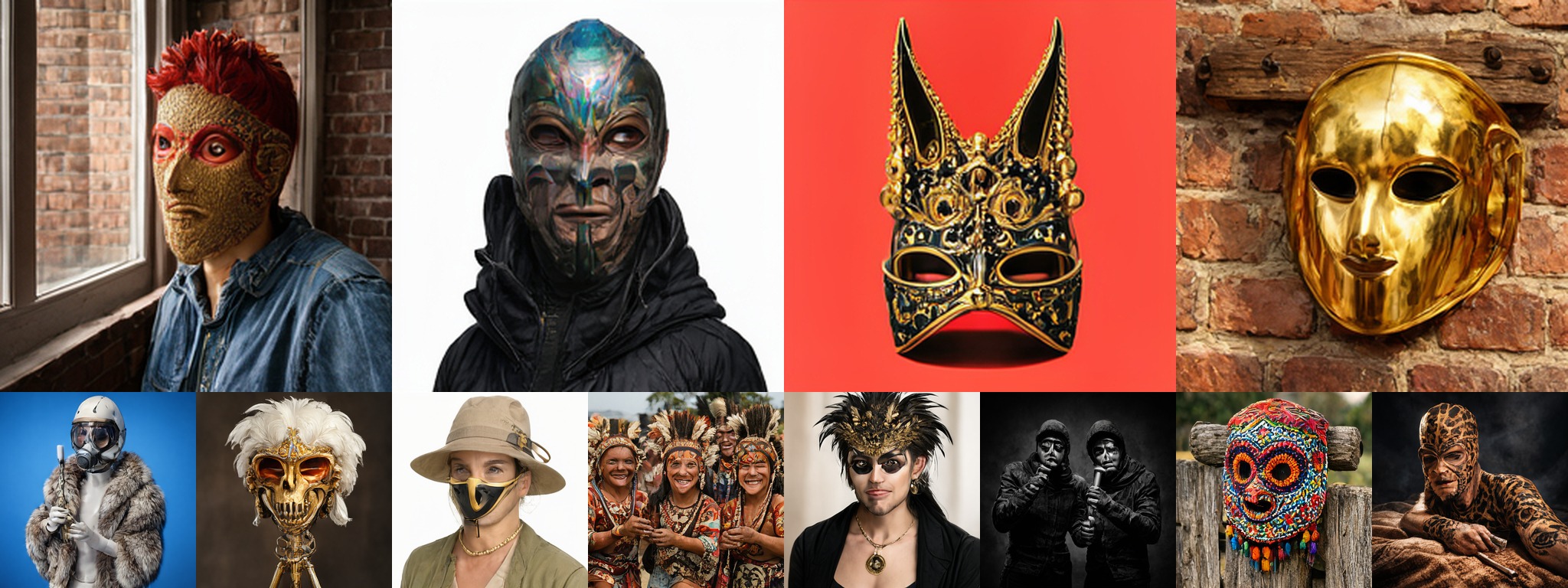}
    \caption{Class ``Mask'' (643). Euler sampler, 35 NFE, CFG $w = 2.0$.}
    \label{fig:app_grid_643}
\end{figure}


\begin{figure}[H]
    \centering
    \includegraphics[width=\linewidth]{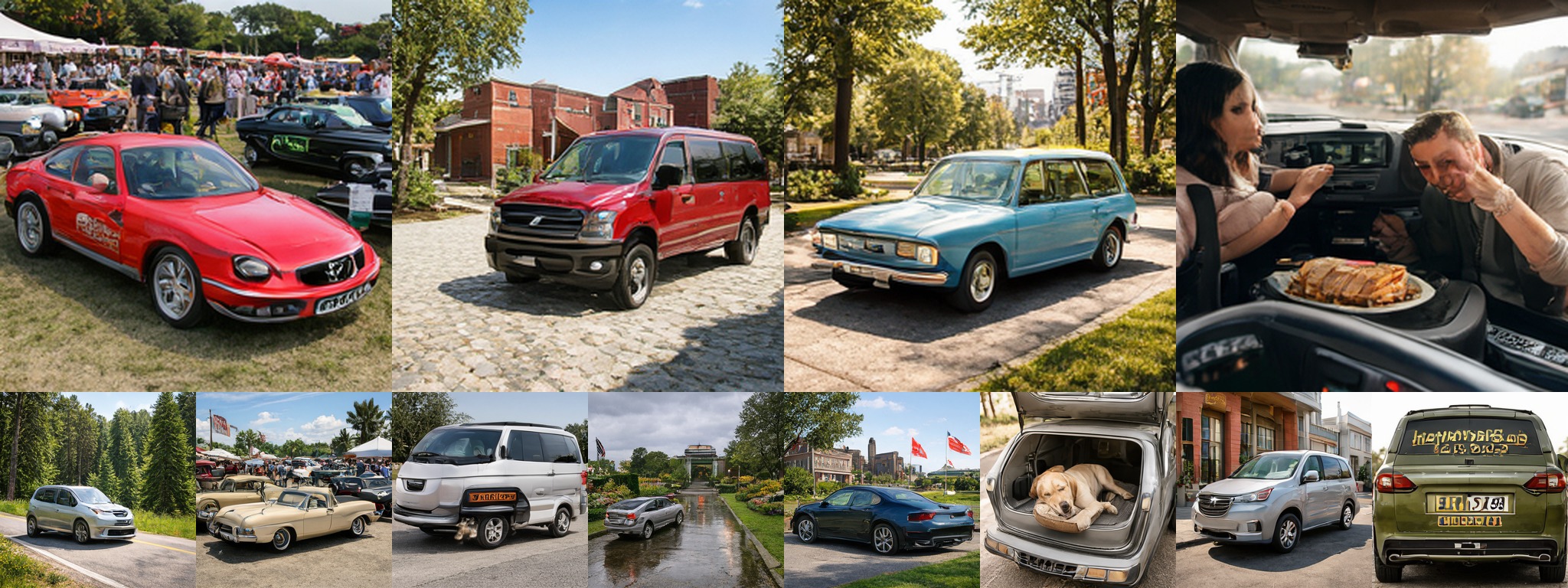}
    \caption{Class ``Minivan'' (656). Euler sampler, 35 NFE, CFG $w = 2.0$.}
    \label{fig:app_grid_656}
\end{figure}

\begin{figure}[H]
    \centering
    \includegraphics[width=\linewidth]{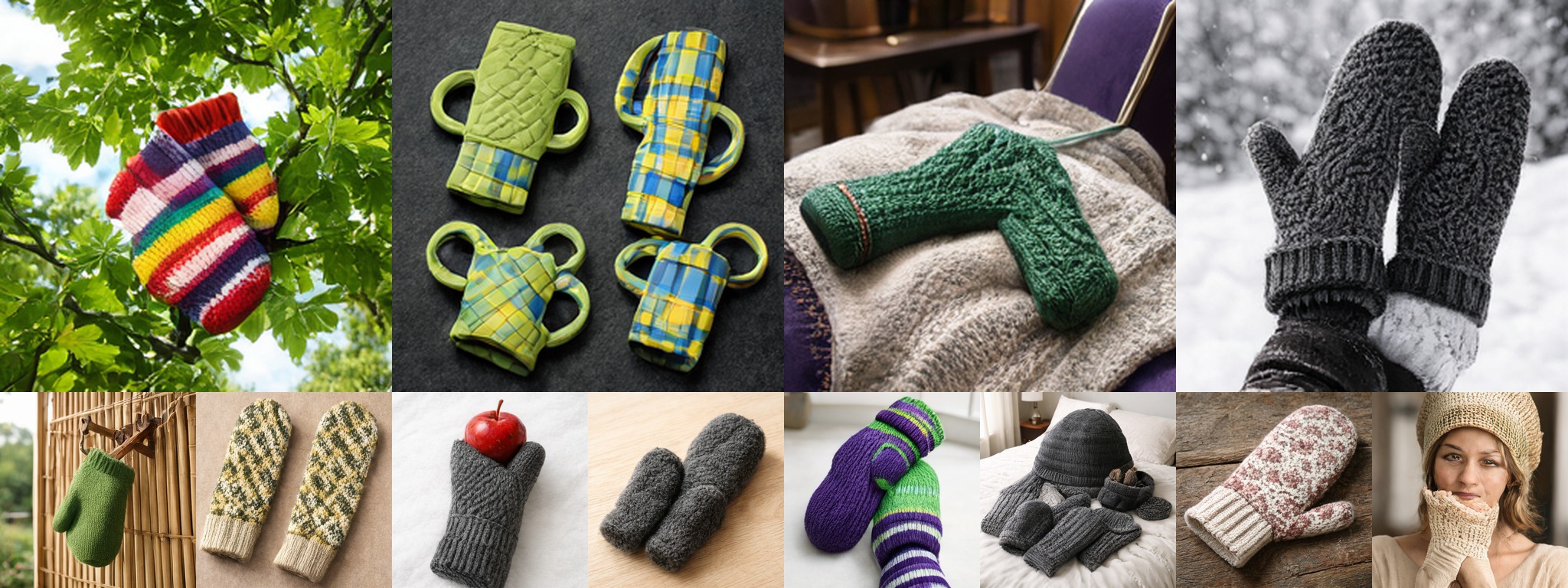}
    \caption{Class ``Mitten'' (658). Euler sampler, 35 NFE, CFG $w = 2.0$.}
    \label{fig:app_grid_658}
\end{figure}

\begin{figure}[H]
    \centering
    \includegraphics[width=\linewidth]{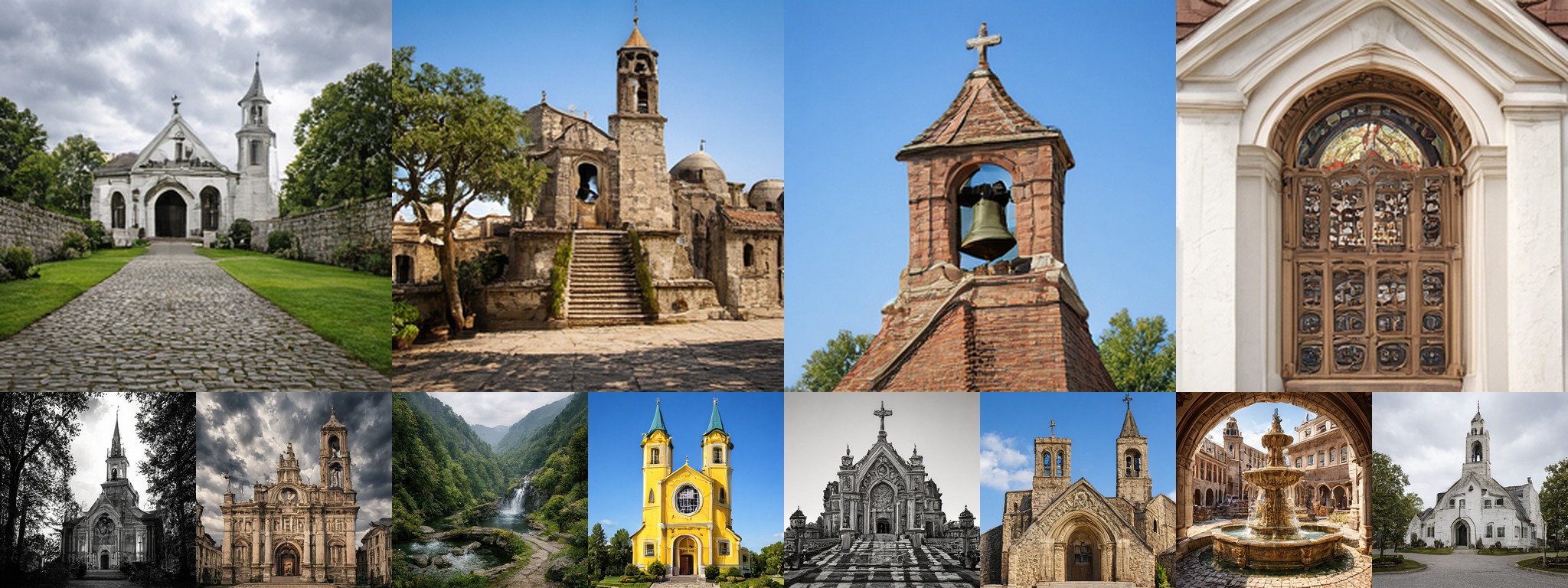}
    \caption{Class ``Monastery'' (663). Euler sampler, 35 NFE, CFG $w = 2.0$.}
    \label{fig:app_grid_663}
\end{figure}

\begin{figure}[H]
    \centering
    \includegraphics[width=\linewidth]{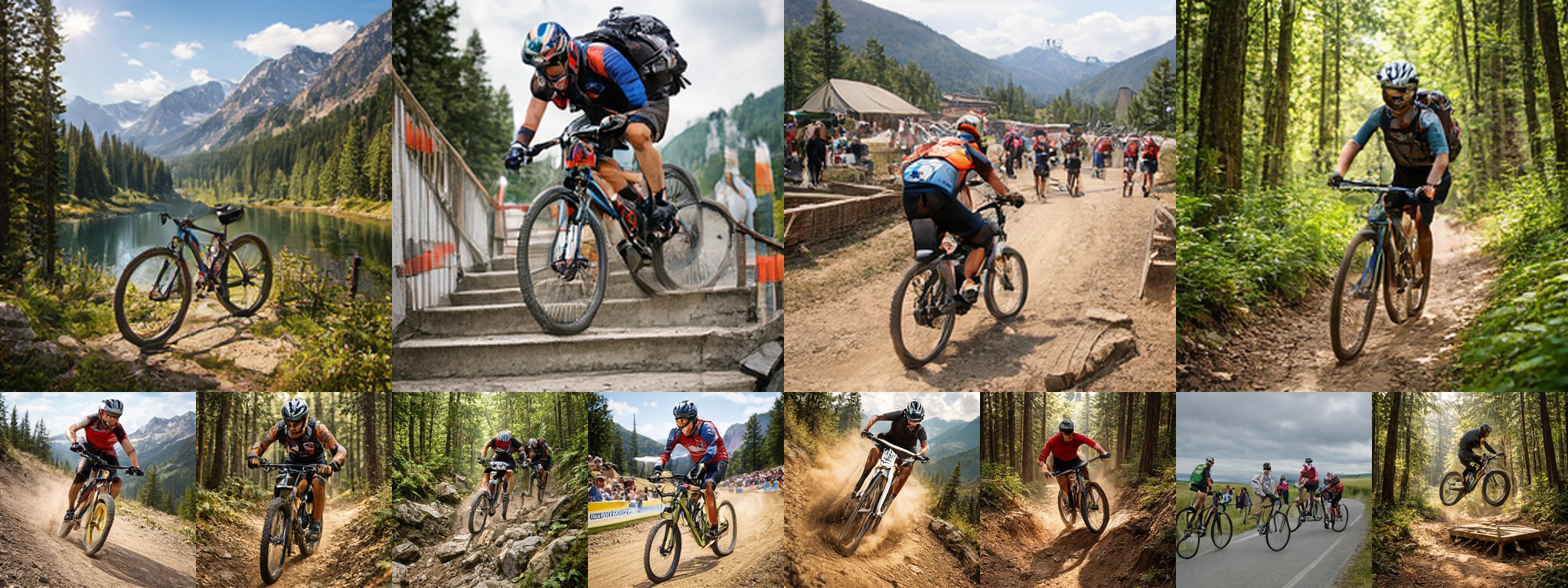}
    \caption{Class ``Mountain bike'' (671). Euler sampler, 35 NFE, CFG $w = 2.0$.}
    \label{fig:app_grid_671}
\end{figure}

\begin{figure}[H]
    \centering
    \includegraphics[width=\linewidth]{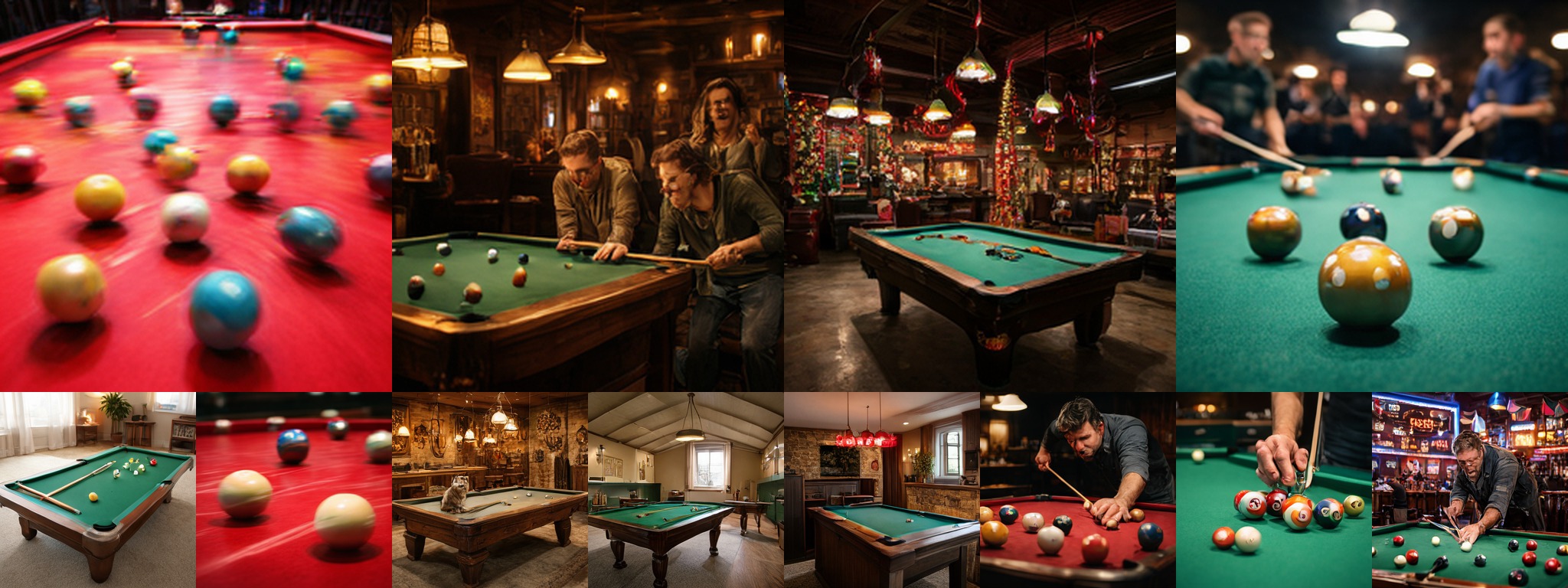}
    \caption{Class ``Pool table'' (736). Euler sampler, 35 NFE, CFG $w = 2.0$.}
    \label{fig:app_grid_736}
\end{figure}

\begin{figure}[H]
    \centering
    \includegraphics[width=\linewidth]{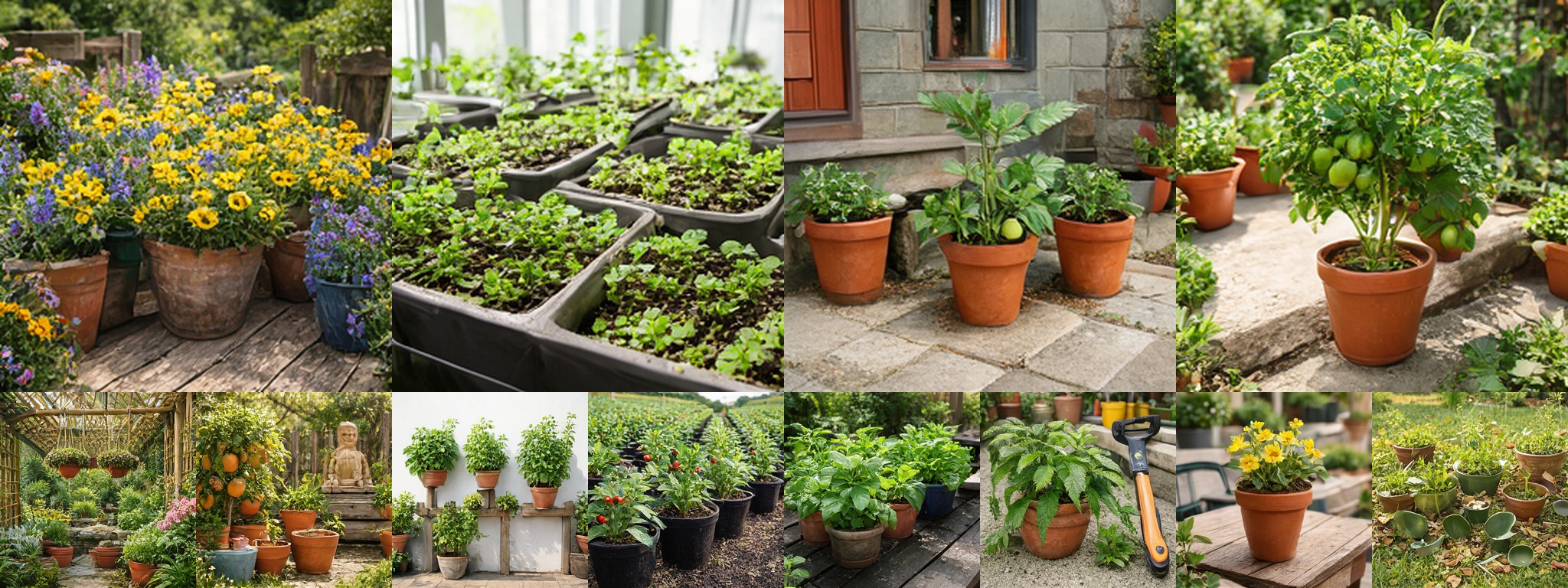}
    \caption{Class ``Pot'' (738). Euler sampler, 35 NFE, CFG $w = 2.0$.}
    \label{fig:app_grid_738}
\end{figure}

\begin{figure}[H]
    \centering
    \includegraphics[width=\linewidth]{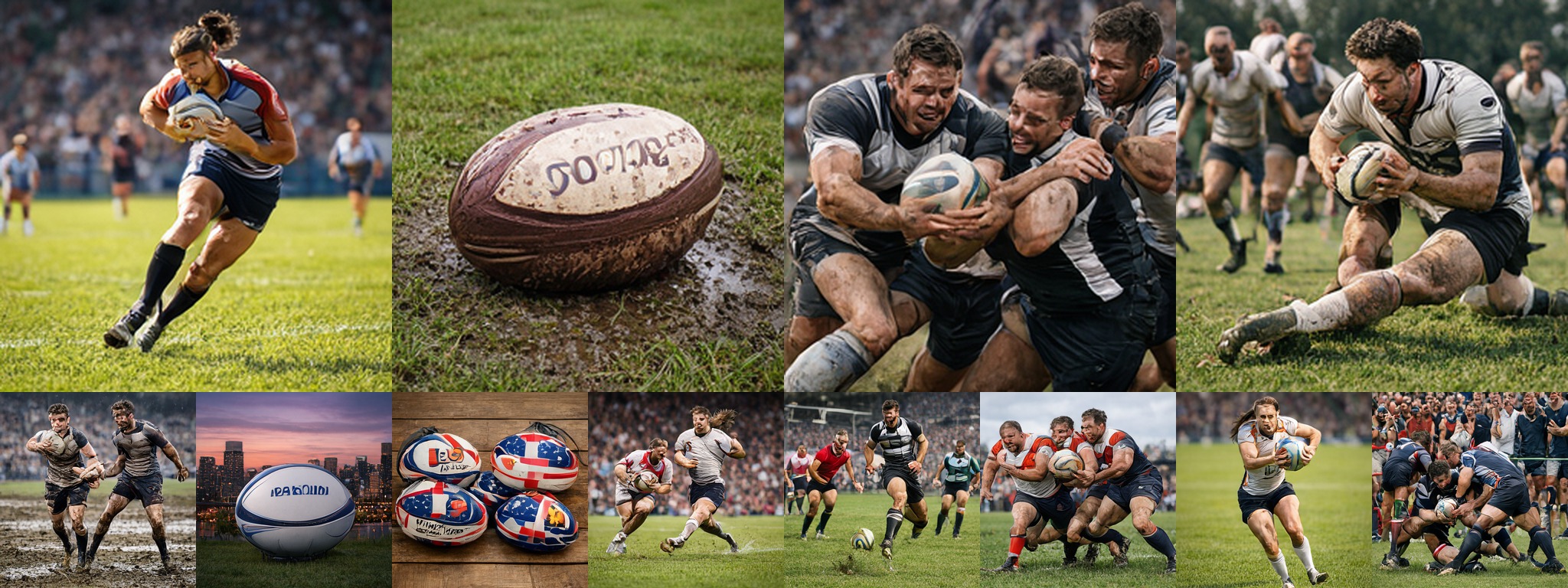}
    \caption{Class ``Rugby ball'' (768). Euler sampler, 35 NFE, CFG $w = 2.0$.}
    \label{fig:app_grid_768}
\end{figure}

\begin{figure}[H]
    \centering
    \includegraphics[width=\linewidth]{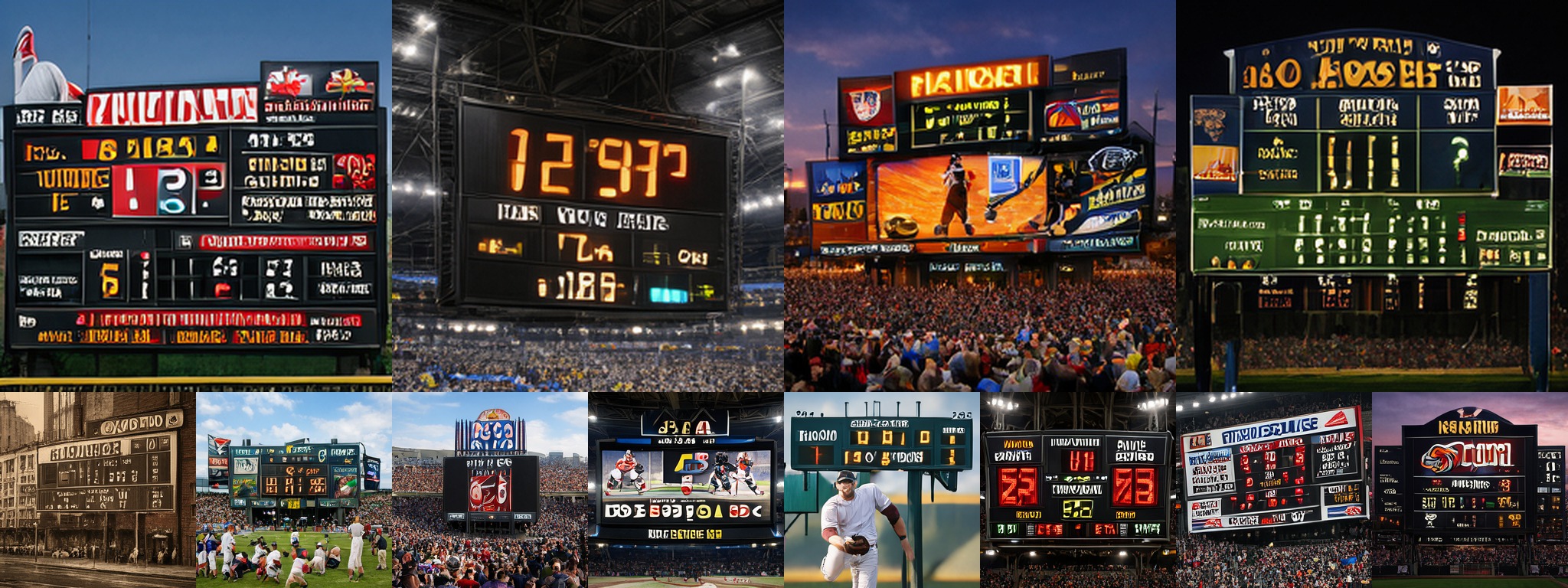}
    \caption{Class ``Scoreboard'' (781). Euler sampler, 35 NFE, CFG $w = 2.0$.}
    \label{fig:app_grid_781}
\end{figure}


\begin{figure}[H]
    \centering
    \includegraphics[width=\linewidth]{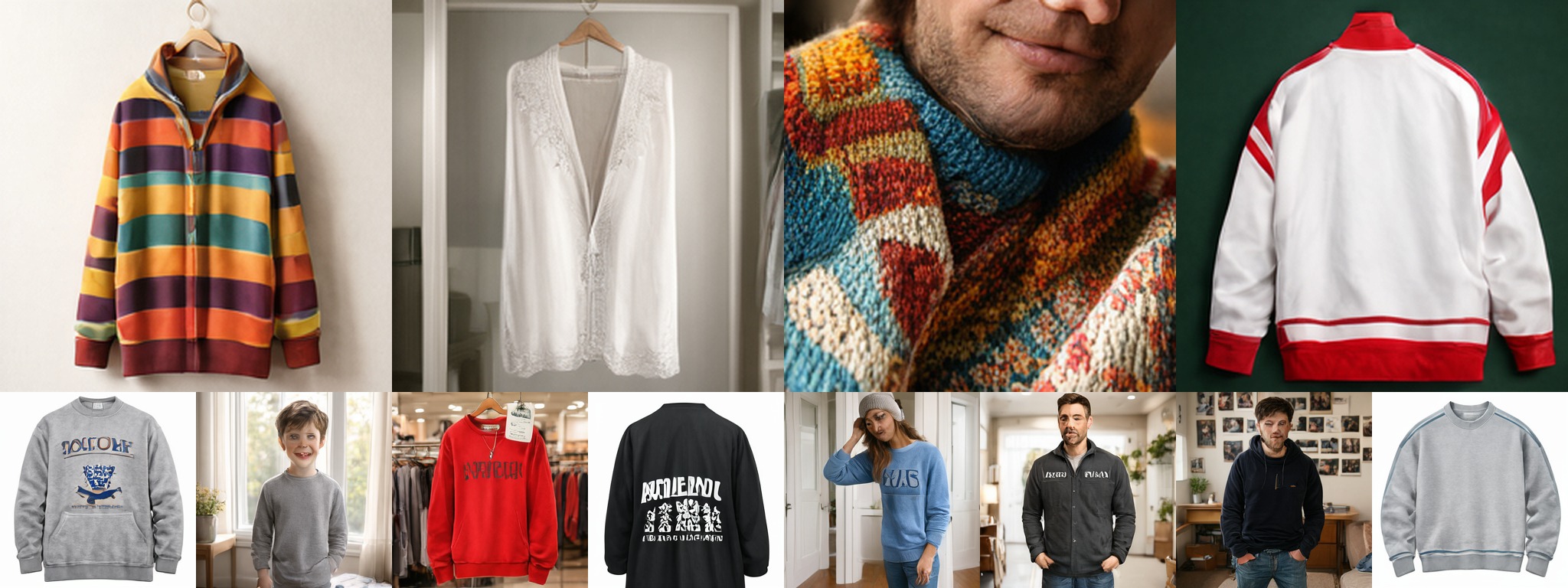}
    \caption{Class ``Sweatshirt'' (841). Euler sampler, 35 NFE, CFG $w = 2.0$.}
    \label{fig:app_grid_841}
\end{figure}

\begin{figure}[H]
    \centering
    \includegraphics[width=\linewidth]{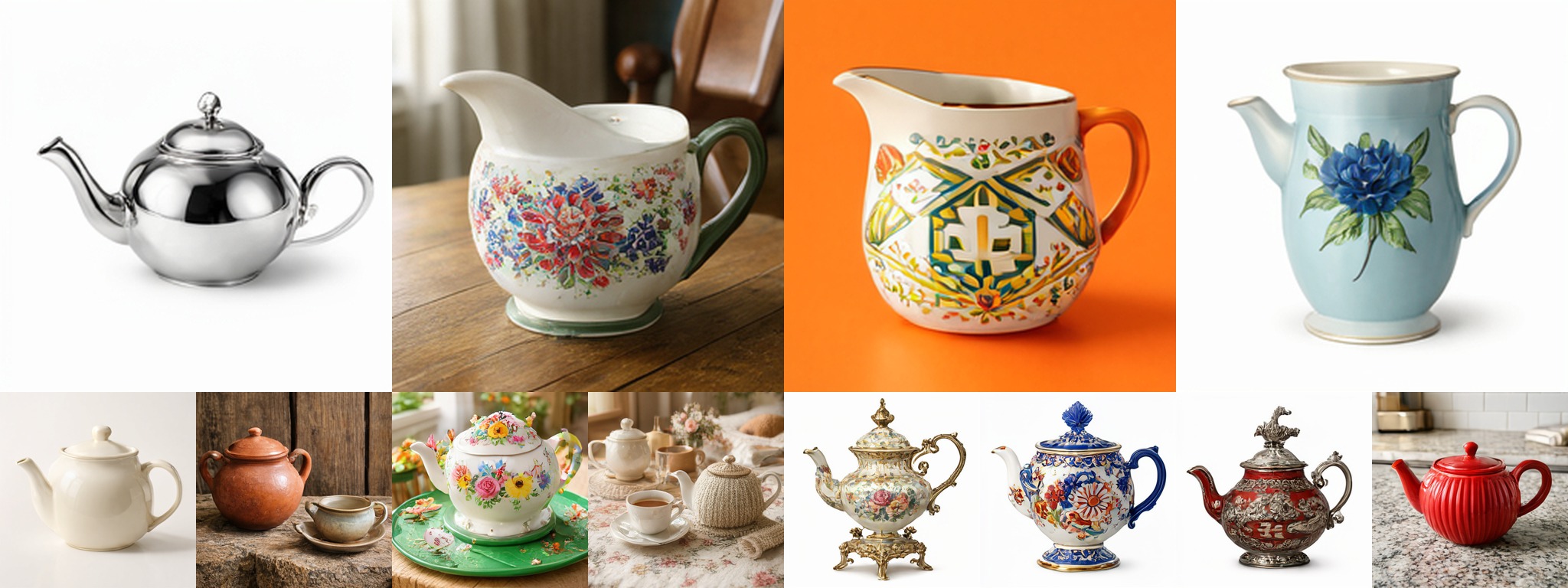}
    \caption{Class ``Teapot'' (849). Euler sampler, 35 NFE, CFG $w = 2.0$.}
    \label{fig:app_grid_849}
\end{figure}

\begin{figure}[H]
    \centering
    \includegraphics[width=\linewidth]{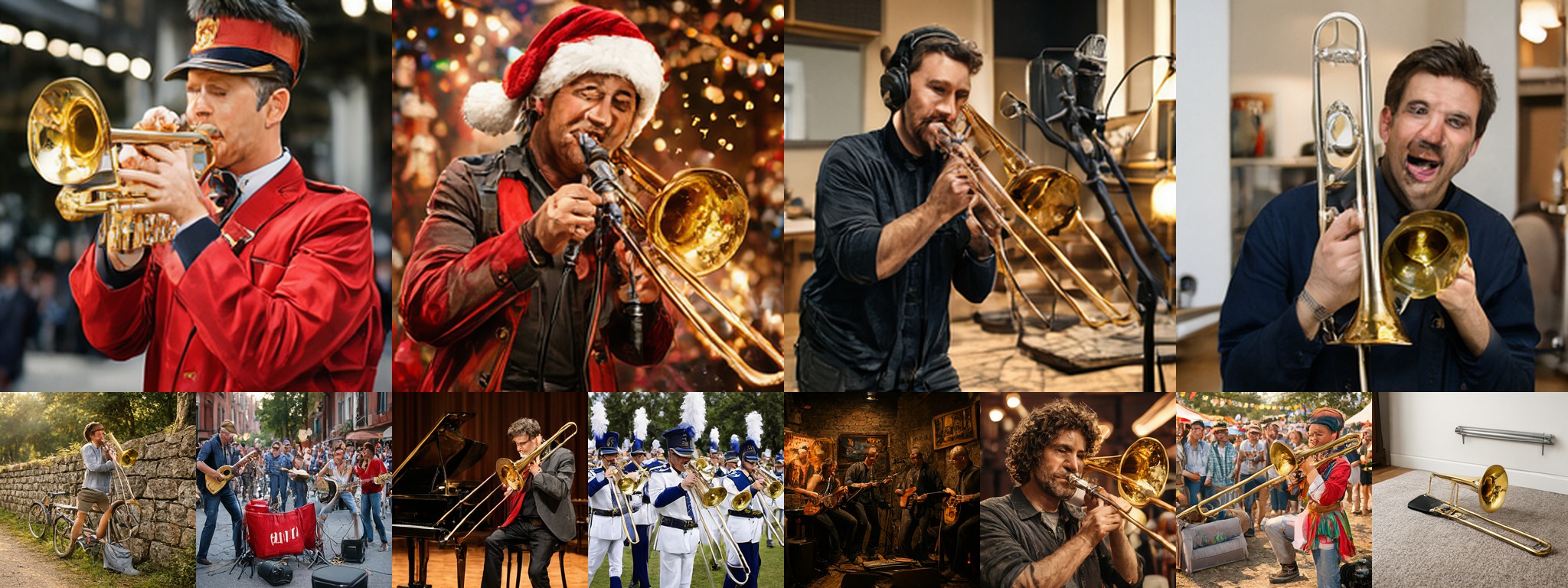}
    \caption{Class ``Trombone'' (875). Euler sampler, 35 NFE, CFG $w = 2.0$.}
    \label{fig:app_grid_875}
\end{figure}

\begin{figure}[H]
    \centering
    \includegraphics[width=\linewidth]{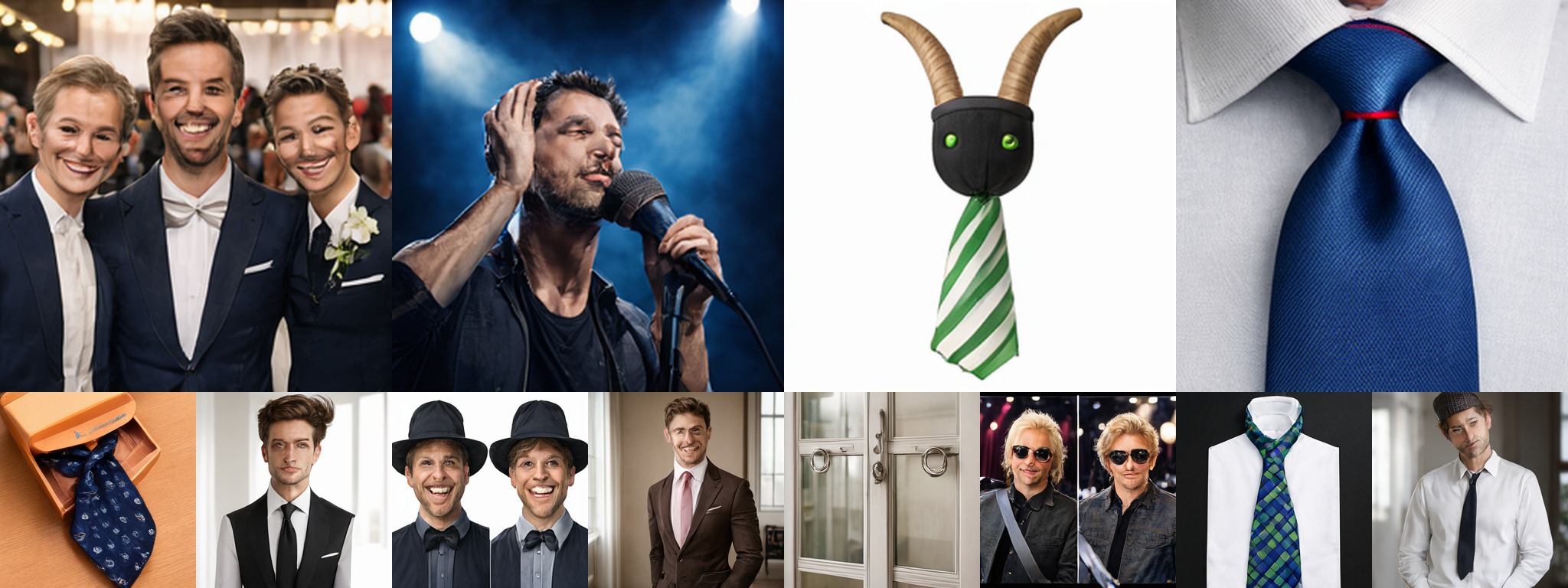}
    \caption{Class ``Windsor tie'' (906). Euler sampler, 35 NFE, CFG $w = 2.0$.}
    \label{fig:app_grid_906}
\end{figure}


\begin{figure}[H]
    \centering
    \includegraphics[width=\linewidth]{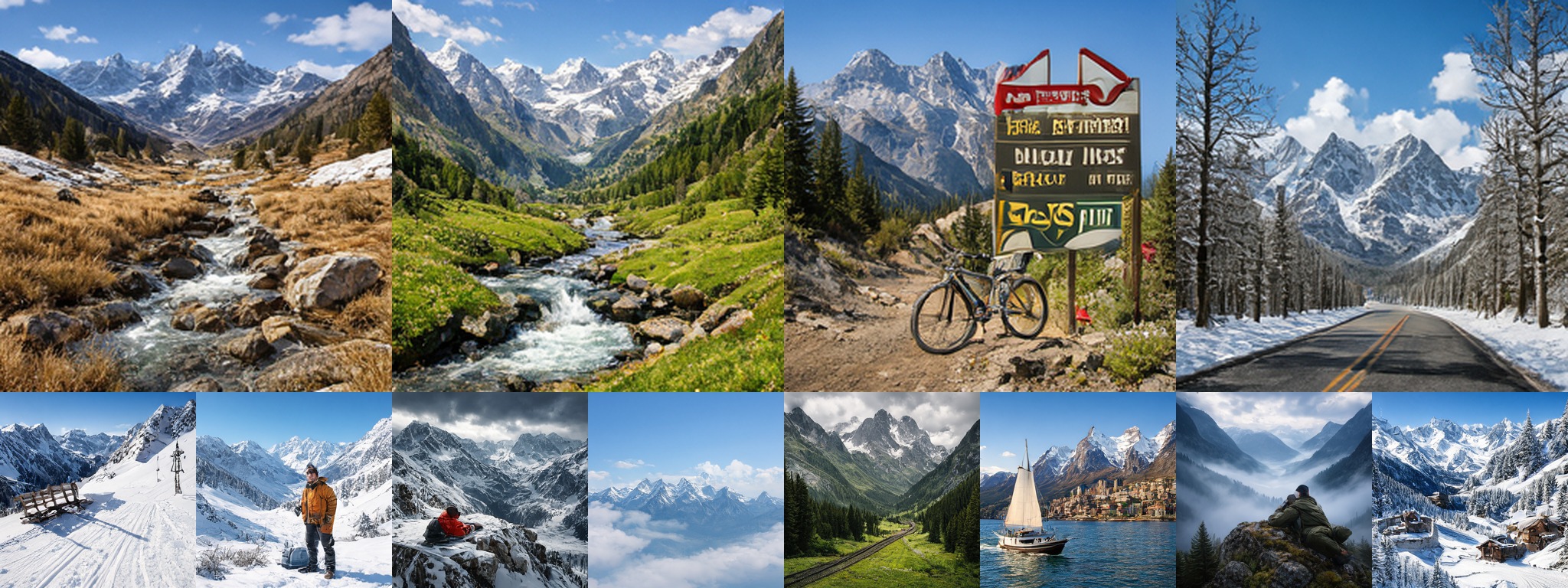}
    \caption{Class ``Alp'' (970). Euler sampler, 35 NFE, CFG $w = 2.0$.}
    \label{fig:app_grid_970}
\end{figure}


\begin{figure}[H]
    \centering
    \includegraphics[width=\linewidth]{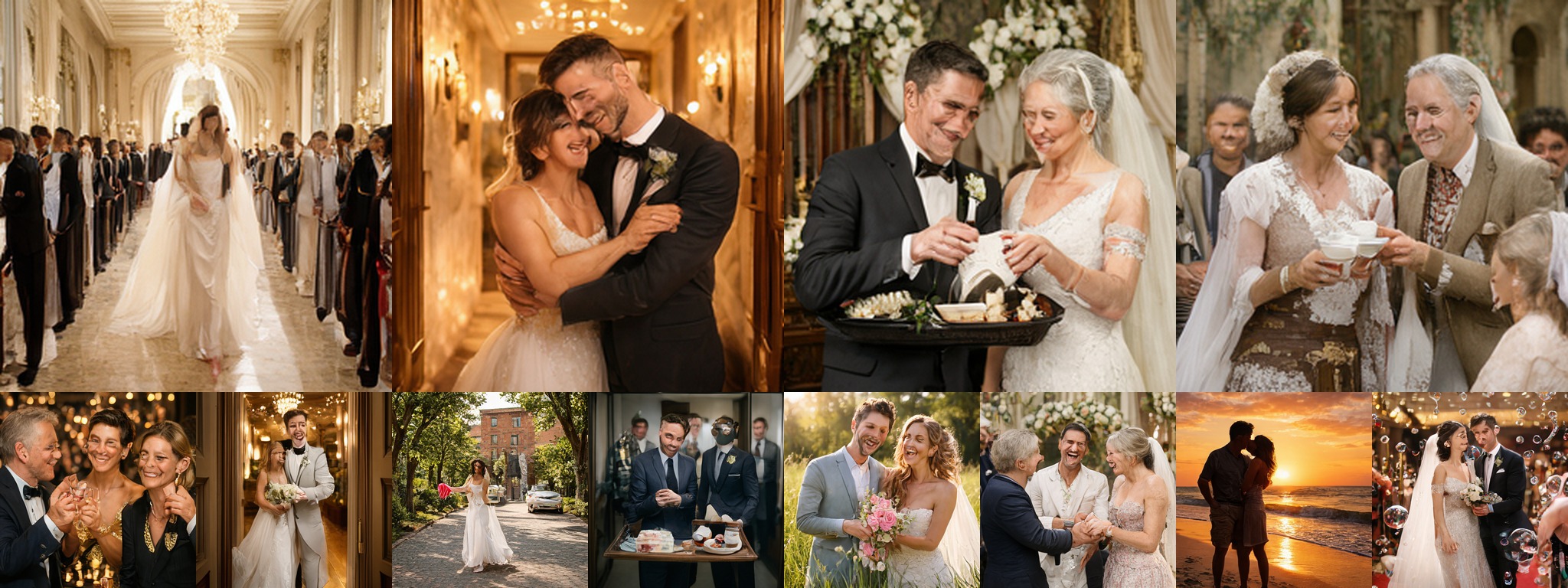}
    \caption{Class ``Groom'' (982). Euler sampler, 35 NFE, CFG $w = 2.0$.}
    \label{fig:app_grid_982}
\end{figure}

\end{document}